%% file: arxiv/arxiv_version.tex
\documentclass{article}

\usepackage{arxiv}

\usepackage[utf8]{inputenc} % allow utf-8 input
\usepackage[T1]{fontenc}    % use 8-bit T1 fonts
\usepackage{hyperref}       % hyperlinks
\usepackage{url}            % simple URL typesetting
\usepackage{booktabs}       % professional-quality tables
\usepackage{amsfonts}       % blackboard math symbols
\usepackage{nicefrac}       % compact symbols for 1/2, etc.
\usepackage{microtype}      % microtypography
\usepackage{lipsum}
\usepackage{graphicx}
\graphicspath{ {./images/} }

\PassOptionsToPackage{numbers}{natbib}

\usepackage{natbib} 
\usepackage[utf8]{inputenc} % allow utf-8 input
\usepackage[T1]{fontenc}    % use 8-bit T1 fonts
\usepackage{hyperref}       % hyperlinks
\usepackage{url}            % simple URL typesetting
\usepackage{booktabs}       % professional-quality tables
\usepackage{amsfonts}       % blackboard math symbols
\usepackage{nicefrac}       % compact symbols for 1/2, etc.
\usepackage{microtype}      % microtypography
\usepackage{xcolor}         % colors

\author{
Willa Potosnak$^{1}$, 
Cristian Challu*$^{1,2}$, 
Mononito Goswami*$^{1}$, 
Kin G. Olivares$^{1,3}$, 
Michał Wiliński$^{1}$, \\
\textbf{Nina Żukowska}$^{1}$, 
\textbf{Artur Dubrawski}$^{1}$%
\thanks{*Equal contribution. 
$^{1}$Auton Lab, School of Computer Science, Carnegie Mellon University. 
$^{2}$Nixtla. 
$^{3}$Amazon. 
Correspondence to: Willa Potosnak <wpotosna@andrew.cmu.edu>.}
}

\newcommand{\model}[1]{\texttt{#1}\xspace}
\newcommand{\ARIMA}{\model{ARIMA}}
\newcommand{\ETS}{\model{AutoETS}}

\newcommand{\MLP}{\model{MLP}}
\newcommand{\DLinear}{\model{DLinear}}
\newcommand{\NLinear}{\model{NLinear}}
\newcommand{\TCN}{\model{TCN}}
\newcommand{\LSTM}{\model{LSTM}}
\newcommand{\RNN}{\model{RNN}}
\newcommand{\TimesNet}{\model{TimesNet}}
\newcommand{\TSMixer}{\model{TSMixer}}

\newcommand{\NHITS}{\model{NHITS}}
\newcommand{\NBEATS}{\model{NBEATS}}

\newcommand{\VanillaTransformer}{\model{VanillaTransformer}}
\newcommand{\TFT}{\model{TFT}}
\newcommand{\PatchTST}{\model{PatchTST}}

\newcommand{\Autoformer}{\model{Autoformer}}
\newcommand{\Informer}{\model{Informer}}
\newcommand{\iTransformer}{\model{iTransformer}}
\newcommand{\Tfive}{\model{T5 Model}}
\newcommand{\BasisFormer}{\model{BasisFormer}}

\newcommand{\MOMENT}{\model{MOMENT}}

\usepackage{caption}
\usepackage{subcaption}
\usepackage{multirow}
\usepackage{graphicx} 
\usepackage{amsmath}
\usepackage{xcolor}
\usepackage{wrapfig}

\usepackage{xspace}
\usepackage{enumerate}
\usepackage{amssymb}
\usepackage{mathtools}
\usepackage{amsthm}
\begin{document}

\title{Investigating Compositional Reasoning in Time Series Foundation Models}

\maketitle

\begin{abstract} 
Large pre-trained time series foundation models (TSFMs) have demonstrated promising zero-shot performance across a wide range of domains. However, a question remains: Do TSFMs succeed by memorizing patterns in training data, or do they possess the ability to reason about such patterns? While reasoning is a topic of great interest in the study of Large Language Models (LLMs), it is undefined and largely unexplored in the context of TSFMs. In this work, inspired by language modeling literature, we formally define compositional reasoning in forecasting and distinguish it from in-distribution generalization. We evaluate the reasoning and generalization capabilities of 16 popular deep learning forecasting models on multiple synthetic and real-world datasets. Additionally, through controlled studies, we systematically examine which design choices in 7 popular open-source TSFMs contribute to improved reasoning capabilities. Our study yields key insights into the impact of TSFM architecture design on compositional reasoning and generalization. We find that patch-based Transformers have the best reasoning performance, closely followed by residualized MLP-based architectures, which are 97\% less computationally complex in terms of FLOPs and 86\% smaller in terms of the number of trainable parameters. Interestingly, in some zero-shot out-of-distribution scenarios, these models can outperform moving average and exponential smoothing statistical baselines trained on in-distribution data. Only a few design choices, such as the tokenization method, had a significant (negative) impact on Transformer model performance. Our code is available at \url{https://github.com/PotosnakW/neuralforecast/tree/tsfm_reasoning}.
\end{abstract}

\section{Introduction}
\label{section:introduction}
\input{sections/01_introduction.tex}
\section{Related Work}
\label{section:related_work}
\input{sections/02_related_work}

\section{Methods}
\label{section:methods}
\input{sections/03_methods}

\section{Results}
\label{section:results}
\input{sections/04_results}
\section{Discussion}
\label{section:discussion}
\input{sections/05_discussion}`

\section*{Acknowledgements}
This work was partially supported by the NSF (awards
2406231 and 2427948), NIH (awards R01NS124642 and
R01DK131586), DARPA (HR00112420329), and the US
Army (W911NF-20-D0002).

\bibliographystyle{plainnat}
\bibliography{references}

\newpage
\section{Supplemental Information}
\input{sections/appendixA}\label{section:appendixA}

\section{Supplemental Results}
\input{sections/appendixB} \label{section:appendixB}

\end{document}

%% file: sections/01_introduction.tex
Foundation models have demonstrated an exceptional ability to generalize in zero-shot prediction tasks. Inspired by the success of such models in Natural Language Processing, recent work has adapted Transformers to build time series foundation models (TSFM). Zero-shot inference is particularly important for time series models, which must handle complex patterns, seasonal variations, and emerging trends where little to no reference data may be available.

To achieve zero-shot generalization, TSFMs are trained on increasingly large and diverse datasets, aiming to expand coverage and minimize unseen patterns. This raises a critical question: \textbf{Do TSFMs succeed by memorizing patterns in training data, or do they possess the ability to reason about such patterns?} If TSFMs rely primarily on memorization, their ability to generalize to unseen data may be limited. In such cases, these memorization-dependent models would be overly reliant on training data and suffer from inefficient knowledge storage, requiring progressively larger models to generalize effectively. We argue that a good TSFM should be capable of \emph{implicit reasoning}, enabling it to go beyond memorization and extrapolate unseen patterns. Such models would require fewer data points to generalize, utilize fewer parameters, and demonstrate greater robustness.

Although reasoning has been extensively studied in language models, their time series counterparts remain largely unexplored in these contexts and require investigation. 

In this work, we take an initial step toward assessing one form reasoning abilities of neural forecasting models, with a focus on compositional reasoning--the ability to leverage and compose learned patterns from simpler forecasting tasks to generalize to more complex, unseen patterns. This definition refines out-of-distribution (OOD) generalization by emphasizing how learning fundamental patterns equips models to handle logically derived, more intricate scenarios.

The main contributions of this paper are:
\begin{enumerate}[(i)]
\item \textbf{Reasoning Framework for Time Series.} We formally define compositional reasoning in time series and introduce a framework to evaluate neural models' reasoning capabilities in forecasting. Using spectral analysis, we assess various methods' capacity to \emph{logically} generalize to unseen periodic patterns through basis function extrapolation. In addition, we propose a new evaluation metric to help identify and rule out models that do not show evidence of compositional reasoning.

\item \textbf{Large-Scale Study of Model Architectures and Design Components.} We evaluate the performance of 16 popular deep learning forecasting models on both synthetic and real-world datasets, identifying architectures that consistently demonstrate generalization and reasoning abilities. Additionally, through controlled studies, we systematically examine which design choices in 7 popular open-source TSFMs contribute to improved reasoning capabilities. We provide open-source code for the modular Transformer model, allowing for easy integration of new TSFM components for further experiments.

\item \textbf{Synthetic Datasets for Scalable Compositional Reasoning Benchmarks.}
We present a scalable synthetic benchmark designed to evaluate compositional reasoning in time series forecasting. The dataset enables controlled experimentation by allowing for varying levels of complexity, including multiple concept compositions and non-stationary signal components. 

\end{enumerate}

% The rest of the paper is structured as follows. Section~\ref{section:related_work} reviews relevant literature. Section~\ref{section:methods} introduces generalization and compositional-reasoning tasks. Section~\ref{section:results} contains our empirical findings. Finally, Section~\ref{section:discussion} discusses future research directions
% and concludes. Complementary material may be found in the Appendices.

%% file: sections/02_related_work.tex
\vspace*{-3pt}
\paragraph{Reasoning in LLMs.} Reasoning is a topic of great interest in the study of LLMs where prior work has developed experiments to elucidate various forms of reasoning \citep{allenshu2023physics3.2, allenzhu2023physics3.1, wang2024grokked, soheeyang2024multihopreasoning, zhong2023mquake}. One of the most widely explored reasoning tasks is ``composition'', which has also been referred to as ``multi-hop'' or chain-of-thought'' reasoning. Compositional reasoning generally requires that a model combine multiple learned concepts observed during training to address a new question or prompt seen only during inference \citep{wang2024grokked, soheeyang2024multihopreasoning, zhong2023mquake}. For example, a composition task for an LLM might include training the model on specific sentences or facts represented as (subject, relation, object) triplets: ``John's friend is Jane" and ``Jane is from NYC." Given a new prompt at inference: ``Where is John's friend from?" the model is evaluated on whether it can generate the correct prediction of `NYC', which requires that the model combine concepts from two facts seen due training. Mathematical reasoning via `addition'-like prompts has also been studied and similarly requires that a model can compose concepts observed during training. For example, given a new prompt, ``Put together 5 and 40", prior work has shown that pre-trained LLMs implicitly compute addition with dimensions in the hidden state that represent numbers through sparse frequency-domain features \citep{zhou2024llmfourieraddition}. Through curated experiments, such as this, language model reasoning is evaluated based on prediction matches, which can only be achieved if the model takes advantage of a specific reasoning strategy being tested. Similarly, for time series reasoning studies, it is crucial to design tailored experiments that effectively uncover reasoning capabilities—similar to those established in prior language model studies—with clear definitions of concepts, compositions, and evaluation strategies.

\vspace*{-3pt}
\paragraph{Reasoning in Time Series Foundation Models.} 
Many foundation models have been proposed for time series forecasting \citep{wolff2024spade, ansari2024chronos, gao2024units, rasul2024lagllama, moirai2024, goswami2024moment, das2024TimesFM, garza2023timegpt1, liutimer, ekambaram2024ttms}. Recent work has primarily focused on extending the capabilities of these models using new architectures~\citep{moiraimoe, timemoe} or by accounting for long and multivariate context~\citep{infinichannelmixer, timerxl}. Fewer studies have focused on what these models are learning and what contributes to their success. These studies have largely explored the concepts that TSFMs learn \citep{wilinski2024exploring, goswami2024moment}, the impact of scaling the model~\citep{yao2024scaling, edwards2024scaling}, and their failure modes in controlled settings~\citep{ansari2024chronos}. However, to the best of our knowledge, no studies have explored whether time series forecasting models and TSFMs exhibit evidence of inherent reasoning abilities. Reasoning in time series foundation models (TSFMs), much like in LLMs, can take a variety of forms. In the LLM literature, distinct types of reasoning—such as chain-of-thought, causal inference, or mathematical reasoning—have each warranted dedicated investigations with tailored datasets and evaluation methods. Similarly, TSFMs may require targeted study across different reasoning capabilities, including compositional reasoning, comparison, inverse search, and causal reasoning involving exogenous variables or contextual information. In this work, we focus specifically on compositional reasoning, motivated by its potential to support robust zero-shot generalization to complex, out-of-distribution (OOD) time series scenarios through through the addition of concepts. By taking the first step to isolate and investigate this type of reasoning ability in regression-based forecasting tasks, we aim to encourage more systematic exploration of this and other reasoning abilities in TSFMs, similar to how the LLM field has advanced through focused reasoning benchmarks.

\begin{figure*}[t!]
    \centering
    % First subfigure
    \begin{subfigure}[b]{0.495\textwidth}
        \centering
        \includegraphics[width=\textwidth, trim=15 15 0 5, clip]{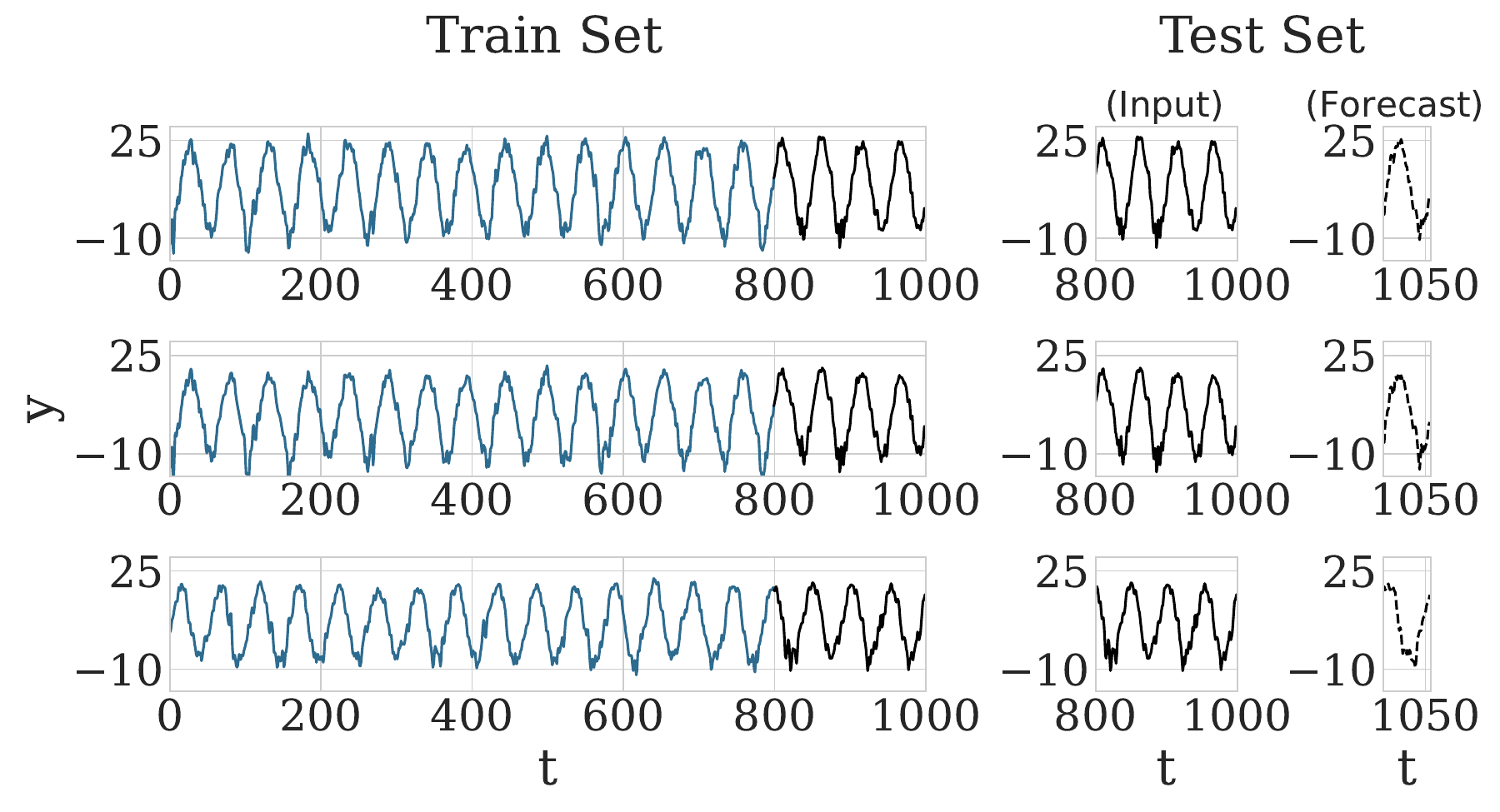}
        \caption{Traditional forecasting paradigm
        }
        \label{fig:methods_general}
    \end{subfigure}
    \hfill
    % Second subfigure
    \begin{subfigure}[b]{0.495\textwidth}
        \centering
        \includegraphics[width=\textwidth, trim=15 15 0 5, clip]{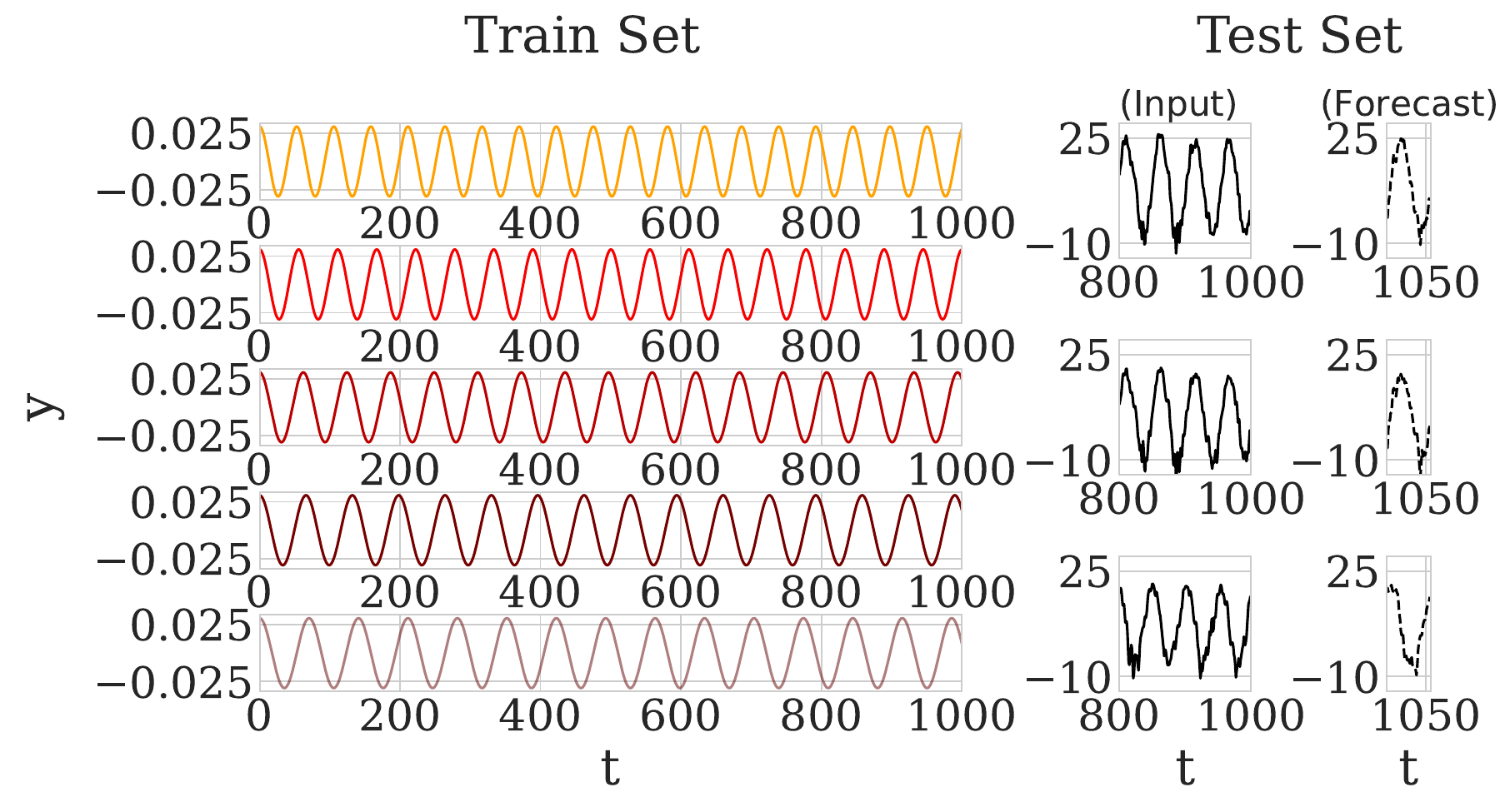}
        \caption{Compositional reasoning forecasting paradigm}
        \label{fig:comparison}
    \end{subfigure}
    \caption{\textbf{(a)} Traditional forecasting paradigm: models are trained directly on the training set of forecast target signals, with forecasts for the temporally subsequent test set generated using the preceding context window. \textbf{(b)} Compositional reasoning forecasting paradigm: models are trained on basis function series that compose the ground truth signal, with forecasts for the temporally subsequent test set generated using the preceding context window.}
    \label{fig:method_compositional}
\end{figure*}

%% file: sections/03_methods.tex
\subsection{Compositional Reasoning Forecasting Framework}
\label{section:compositional_reasoning}

In this section, we introduce our compositional reasoning forecasting framework and discuss its distinctions from the traditional forecasting paradigm.

\vspace*{-3pt}
\paragraph{Forecasting Task.} 
Let a time series be denoted by $\mathbf{y}(t)$, with time $t \in \{1,...,T+H\}$. We consider a special regression problem that aims to learn a function $f: \mathcal{X} \mapsto \mathcal{Y}$, where the domain $\mathcal{X}=\{\textbf{y}_{t-l:t}\}$ is composed of auto-regressive context of length $l$, and codomain is the $h$-step future of a series $\mathcal{Y}=\{\textbf{y}_{t:t+h} \}$. We denote the predictions:
\begin{align}
    \hat{\textbf{y}}_{t:t+h} = f(\textbf{y}_{t-l:t}).
\end{align}

\vspace*{-3pt}
\paragraph{Traditional Forecasting Paradigm.} 
In the traditional forecasting paradigm, depicted in Fig.~\ref{fig:methods_general}, the train and test sets are defined by the non-overlapping sections of a time series, before and after a time $T$.
\begin{align}
    \mathcal{D}^{(train)} &= \{(\textbf{y}_{t-l:t}, \textbf{y}_{t:t+h})\}
    \quad \mathrm{with} \quad t \leq T \\
    \mathcal{D}^{(test)} &= \{(\textbf{y}_{t-l:t}, \textbf{y}_{t:t+h})\}
    \quad \mathrm{with} \quad t > T    
\end{align}
Under reasonable stationarity assumptions the train and test datasets will have a stable distribution, also known as an in-distribution (ID) forecast scenario, 
\begin{equation}
\mathbb{P}(\mathcal{D}^{(train)})= \mathbb{P}(\mathcal{D}^{(test)}).
\end{equation}
The traditional forecasting paradigm optimizes a parameterized model $f_\theta$, which is used to generate forecasts for the test set. The model then is evaluated on the test set $\mathcal{D}^{(test)}$ using temporal cross-validation.

\vspace*{-3pt}
\paragraph{Compositional Reasoning Forecasting Paradigm.} 
Compositional reasoning cannot be evaluated within a traditional forecasting framework due to the unclear delineation between time series components and the composition—or target signal—we aim to forecast. To test the compositional reasoning capabilities of forecasting models, as illustrated in Fig.~\ref{fig:comparison}, we design a novel forecasting task where, instead of the original series $\mathbf{y}(t)$, the train dataset $\mathcal{D}^{(train)}$ consists of the top-$k$ largest spectral components of $\mathbf{y}(t)$, obtained using Discrete Fourier Transform: 
\begin{align}
    \text{If} \quad \mathbf{y}(t) &= \sum_{w=0}^n c_w e^{i \frac{2\pi w}{n} t}, \quad \text{then} \\
    \mathcal{D}^{(train)} &= 
    \left\{ 
    \left( e^{i \frac{2\pi w}{n} [t-l:t]},  e^{i \frac{2\pi w}{n} [t:t+h]} \right) 
    \right\} \;  \\
    &\text{with} \; t \leq T  \; \text{ and } |c_w| \in \text{top-}k \left( \{ |c_0|, |c_1|, \dots, |c_T| \} \right) \quad \notag \\
    \mathcal{D}^{(test)} &= \left\{ (\mathbf{y}_{t-l:t}, \mathbf{y}_{t:t+h}) \right\} \; \text{with} \; t > T,
\end{align}
where $c_w$ represents the Fourier coefficients associated with the frequency component $w$ in the Fourier series expansion of the time series $\textbf{y}(t)$ and $n$ is the number of individual timestamps. As discussed in Section~\ref{section:related_work}, compositional reasoning refers to a model's ability to synthesize well-defined concepts to make accurate predictions. In our proposed compositional reasoning paradigm, we train models on the spectral decomposition of the time series and evaluate them on the original series. This approach allows us to assess a model's ability to logically generalize to unseen periodic patterns through the synthesis of fundamental building blocks—specifically, sine and cosine basis functions.

By adopting this framework, we move beyond traditional ID generalization and instead create carefully crafted OOD scenarios, where the model can still succeed through its reasoning capabilities.
\begin{equation}
\mathbb{P}(\mathcal{D}^{(train)}) \neq \mathbb{P}(\mathcal{D}^{(test)})
\end{equation}
As models are evaluated in a zero-shot forecasting setting, generalization can be achieved only through the addition of time series basis functions seen during training, or \textit{composition-type reasoning}.

\subsection{Data}\label{section:methods_data}
The datasets used in this study span both synthetic and real-world time series. Together, these datasets offer a broad spectrum of temporal frequencies, patterns, and domains for evaluating model compositional reasoning capabilities.

\vspace*{-3pt}
\paragraph{Synthetic Datasets.}
We construct synthetic datasets for scalable compositional reasoning benchmarks. The \textbf{Synthetic Sinusoid} dataset, comprises $N = 100$ time series, each composed of $m = 2$ stationary signals drawn from $\mathcal{F}_{\text{stationary}} = \{a \sin(2\pi b t),\ a \cos(2\pi b t)\}$, where $a \in [1, 32]$ and $b \in [3, 32]$. Compositions of time series form the test set $D^{(\text{test})}$, while their basis function time series form the train set $D^{(\text{train})}$, as described in Section~\ref{section:compositional_reasoning}. While this benchmark can scale to more complex compositions, we leave such extensions to future work, as even minimal cases with $m=2$ compositions pose challenges for current models and warrant further study and architectural advances. In addition to the \textbf{Synthetic Sinusoid} dataset results presented in the main paper, we introduce two additional benchmarks with nonstationary time series, combining sinusoidal signals with trend signals from $\mathcal{F}_{\text{nonstationary}} = \{mt\}$, where $m \in [-32, 32]$, with experiments included in the appendix. Additional dataset details are included in Section~\ref{apd:synthetic_data_parameters}.

\vspace*{-3pt}
\paragraph{Real-World Datasets.} We use widely studied time series datasets available in open-source repositories, including GIFT-Eval. \citep{aksu2024giftevalbenchmark}. The \textbf{Electricity Transformer Temperature (ETTm2)} dataset records measurements from an electricity transformer in a region of a province in China. It includes data on oil temperature and various load types, such as high useful load and high useless load, collected at 15-minute intervals between July 2016 and July 2018. The \textbf{Electricity (ECL)} dataset contains the hourly electricity consumption of 321 customers from 2012 to 2014. The \textbf{Solar} dataset contains hourly data on solar power generation in the US in 2006 \citep{aksu2024giftevalbenchmark, ansari2024chronos}. The \textbf{Subseasonal} dataset contains climate time series data at the daily level \citep{mouatadid2024subseasonalsubseasonaldataset, aksu2024giftevalbenchmark}. Modeling seasonal patterns at the daily resolution requires long context lengths, so we downsample to weekly frequency to capture seasonal information within a manageable context length. The \textbf{Loop Seattle} dataset contains spatio-temporal speed data collected every 5 minutes from inductive loop detectors deployed on Seattle freeways \citep{cui2018deep, cui2019traffic}. For our study, we use the downsampled 1-hour version of the dataset \citep{jiang2024libcityunifiedlibraryefficient, aksu2024giftevalbenchmark}. More information on the datasets can be found in section \ref{apd:realworld_data_parameters}.

\vspace*{-3pt}
\paragraph{Preprocessing.}
We preprocess the real-world datasets by segmenting them into patches of 1056 time steps with a stride of 528, ensuring a consistent training set size of 1008. The forecast horizon is 48, consistent with the short-term prediction benchmarks in GIFT-Eval, such as ETTm2, Solar, and Loop Seattle \citep{aksu2024giftevalbenchmark}. We focus on stationary periodic subseries aligned with the synthetic dataset, as this alignment enables controlled experiments while avoiding the additional complexities of nonstationarity, which we leave for future work. To evaluate the stationarity of each series segment, we apply the Augmented Dickey-Fuller test ($\alpha=0.001$) to identify and exclude series with extreme nonstationary characteristics from the final subset. From the refined set of stationary segments, we select the top 100 with the highest mean autocorrelation function (ACF) values for inclusion in the final dataset, ensuring consistency in the number of series across datasets.

\begin{wrapfigure}[31]{r}{0.5\textwidth}
    \centering
    \includegraphics[
        width=0.98\textwidth,
        trim=0 36 400 0,
        clip
    ]{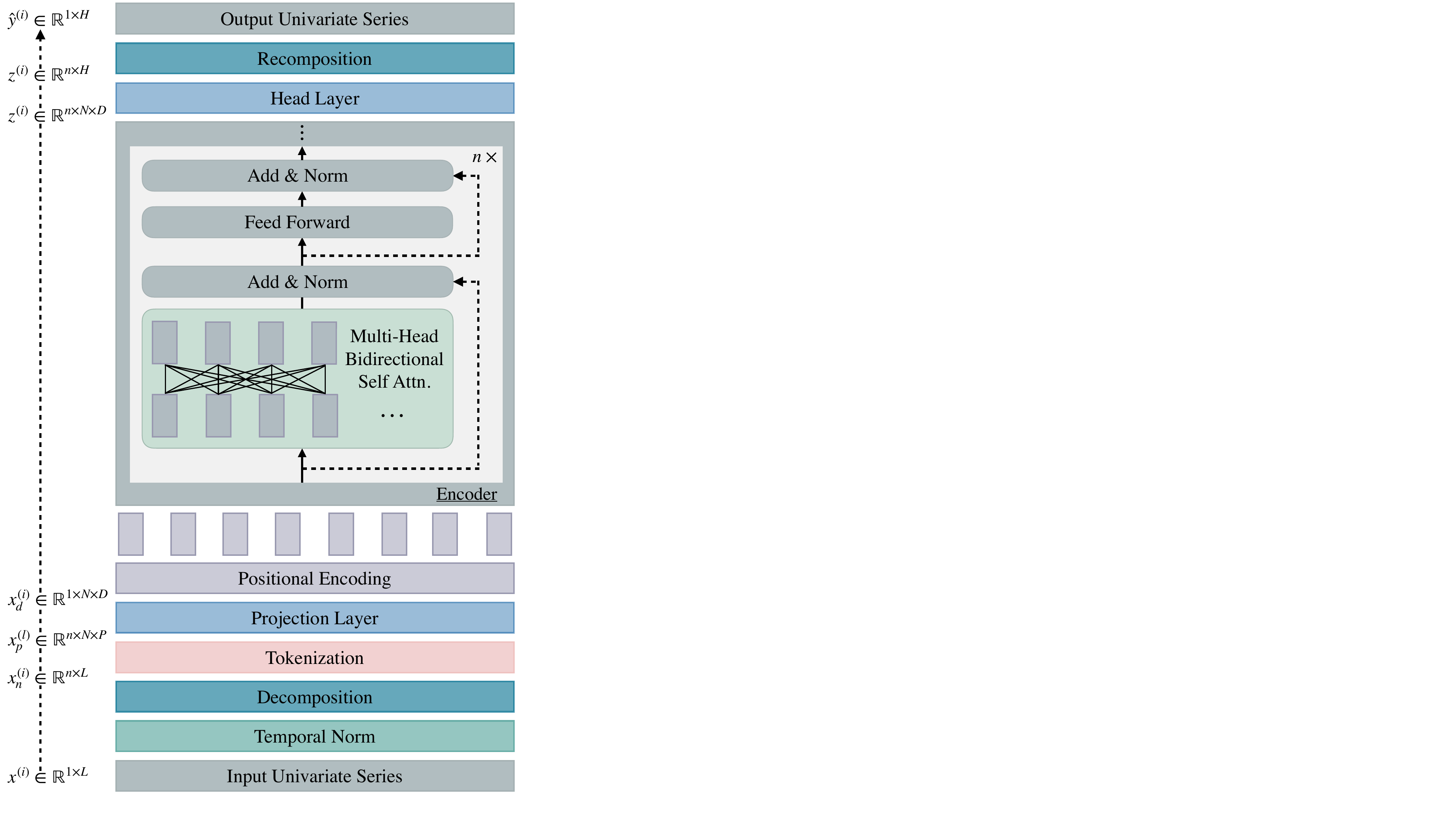}
    \caption{
        Transformer architecture based on the open-source T5 encoder backbone used in our controlled studies 
        to evaluate various architecture components used across TSFMs. Components for which no ablation studies 
        are conducted are shown in gray.
    }
\label{fig:t5_model_architecture}
\end{wrapfigure}

\subsection{Models}

\vspace*{-3pt}
\paragraph{Baselines.} We utilized \ARIMA \citep{hyndman2008_arima,hyndman2024forecasting} and \ETS \citep{brown_1956_ets} as our statistical baseline models.

\vspace*{-3pt}
\paragraph{Deep Learning Models.} We trained models using 16 algorithms that cover a range of architectures and are widely used in practice. Trained models include: \textbf{Linear models}: \DLinear~\citep{zeng_2023_dlinear}, \NLinear~\citep{zeng_2023_dlinear}; \textbf{Multi-Layer Perceptron-based}: (\MLP)~\citep{rosenblatt1958_mlp}, \NHITS~\cite{challu_olivares2022_nhits}, 
\NBEATS~\cite{oreshkin2020nbeats, OlivaresChallu2022_nbeats}, and \TSMixer~\citep{chen2023tsmixer}; \textbf{Recurrent Neural Network-based}: Long-Short Term Memory (\LSTM)~\citep{sak2014_lstm}; \textbf{Convolutional Neural Network-Based}: Temporal Convolution Network (\TCN)~\citep{bai2018_tcn, oord2016_tcn} and \TimesNet~\citep{wu2023timesnettemporal2dvariationmodeling}; and \textbf{Transformer-Based}: \VanillaTransformer~\citep{vaswani_2021_attentionisallyouneed, zhou2021informerefficienttransformerlong}, inverted Transformer (\iTransformer)~\citep{liu2024itransformerinvertedtransformerseffective}, \Autoformer~\citep{wu_2021_autoformer}, \Informer~\citep{zhou2021informerefficienttransformerlong}, Temporal Fusion Transformer (\TFT)~\citep{lim2021_tft}, and patch time series Transformer (\PatchTST)~\citep{nie2023patchtst}. More information on these models, as well as model training and hyperparameters, is provided in Appendix~\ref{apd:first_models} and \ref{apd:first_hyperparameters}.

\vspace*{-3pt}
\paragraph{T5-based Time Series Foundation Models.} 
In addition to the aforementioned models, we aim to evaluate reasoning abilities of TSFMs. TSFMs are combinations of many architectural component design decisions, as summarized in Table~\ref{tab:tsfm_design_decisions_comparison}. Comparing the performance of open-source TSFM models on compositional reasoning tasks alone does not allow us to isolate performance improvements attributable to specific components. To address this, we implement a deep learning Transformer architecture using the open-source \Tfive backbone, designed with modular components to enable flexible experimentation with various design decisions commonly employed in TSFMs, allowing for controlled testing. We pick T5 as the Transformer backbone as these are an open source family of LLMs of varying sizes, with efficient implementations, both encoders and decoders, and used by existing TSFMs \citep{infinichannelmixer, goswami2024moment}. The basic Transformer architecture, illustrated in Fig.~\ref{fig:t5_model_architecture}, incorporates diverse design choices, including input series tokenization (e.g., patching, binning), model size (e.g., tiny, mini), projection layer types (e.g., linear layers, residual networks), scalers (e.g., standard, robust), loss functions (e.g., mean squared error, quantile loss), attention mechanisms (e.g., bidirectional, causal), and positional encodings (e.g., sincos, relative). Additionally, we conduct ablation studies on context length, token (patch) length, and input series decomposition, the latter being inspired by models such as \DLinear and \Autoformer. An outline of the various component types and parameters for each design decision ablation is included in Table~\ref{tab:tsfm_design_decisions_experiments}. This controlled setup is crucial for directly comparing architectures under identical conditions. By evaluating Transformer models with key components aligned with various TSFMs, our work provides valuable insights into which TSFM design decisions are better suited for reasoning in OOD scenarios. 

\subsection{Evaluation}\label{section:evaluation}
Model forecasts were evaluated by computing the mean absolute error (MAE) across all dataset series, and the mean and standard deviation of the results over three random seeds were reported. The formulation for MAE is provided in Appendix~\ref{apd:eval_metrics}.

To assess compositional reasoning performance in the output space, we propose a new evaluation metric based on spectral analysis: \textit{Top-$k$ Basis Win}. We define a Top-$k$ Basis Win as a case where the model forecast $\hat{y}$ achieves equal to or lower error $\mathcal{L}$ (MAE) compared to the top-$k$ basis function signal compositions:
\begin{align} \text{Top-}k \text{ Basis Win} \iff \mathcal{L} (y, \hat{y}) \leq \mathcal{L} \left(y, \sum_{w=0}^{k} c_w e^{i \frac{2\pi w}{n} t} \right) 
\end{align}
We report the Top-$k_{max}$ Basis Win defined as $\max ( \{k : \text{Top-}k$ Basis Wins for $k \in \{1, 2, \dots, n\} \})$. This metric enables clear thresholds to distinguish between models with and without compositional reasoning, which MAE alone cannot. Specifically, we use a threshold of Top-$k\!=\!2$ Basis Wins. Models that fail to outperform this simple benchmark do not demonstrate evidence of compositional reasoning and may simply recall individual frequency components learned during training.

To compare different models, we use statistical tests used by prior work \citep{IsmailFawaz2018deep, goswami2024aqua}. Critical difference (CD) diagrams proposed in \citep{demsar2006cddiagrams} were used to visualize model average rank over the datasets based on average MAE computed over three random seeds for each dataset. In addition to showing model rank, CD diagrams were used to highlight whether the performance of two models is significantly different based on the Wilcoxon signed-rank test with Holm correction. A thick horizontal line groups methods with no significant difference. Details on statistical significance tests can be found in the Appendix~\ref{apd:significance_tests}.

%% file: sections/04_results.tex
\begin{figure}[t!]
    \centering
    \begin{subfigure}[t]{0.48\textwidth}
        \centering
        \includegraphics[width=\textwidth, trim=10 65 70 0, clip]{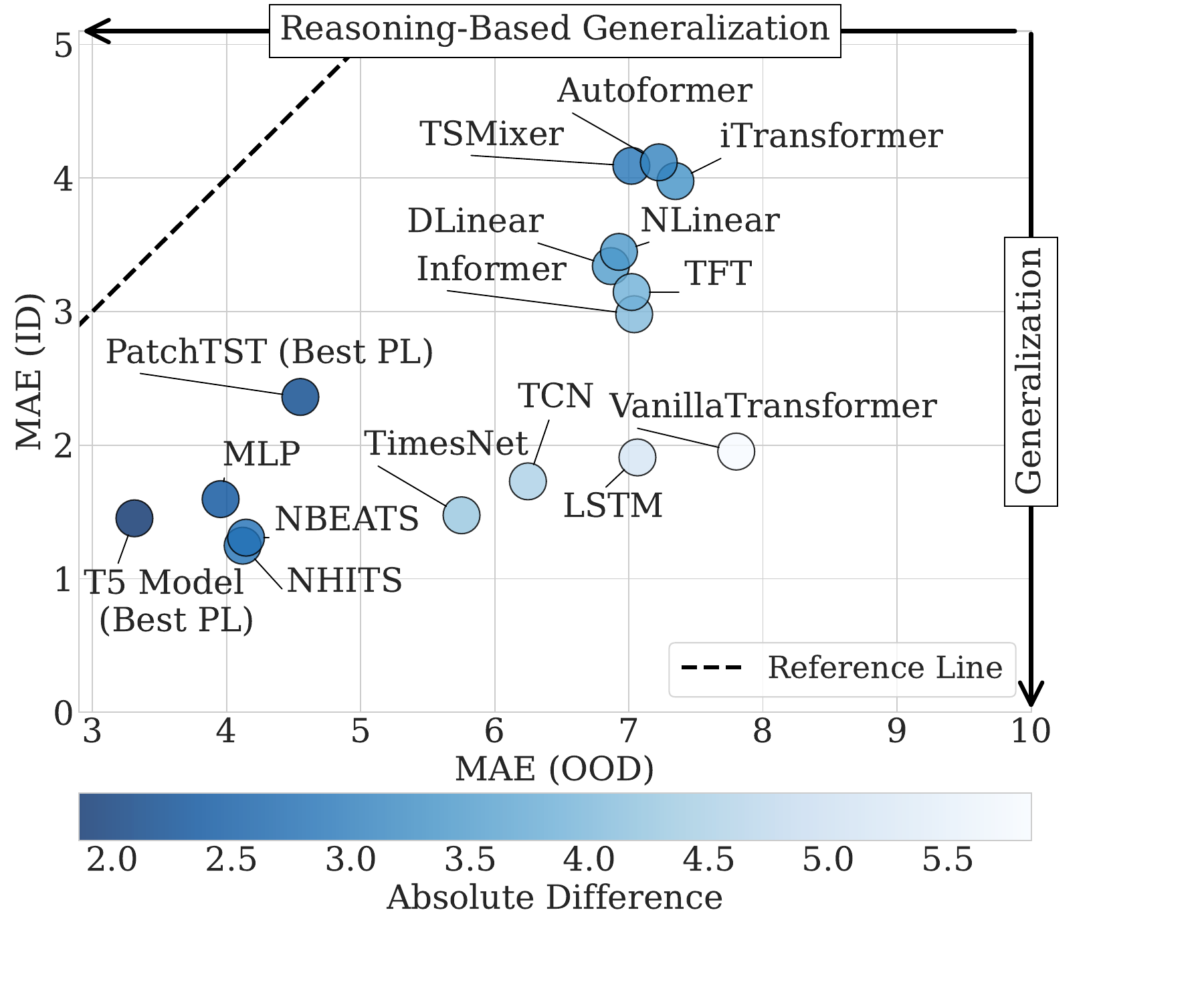}
        \caption{
            The MAE across datasets for ID and OOD data. The absolute difference in MAE between ID and OOD data is shown in blue, with lighter colors indicating larger deviations.
        }
        \label{fig:error_scatter}
    \end{subfigure}
    \hfill
    \begin{subfigure}[t]{0.46\textwidth}
        \centering
        \includegraphics[width=\textwidth, trim=95 50 0 0, clip]{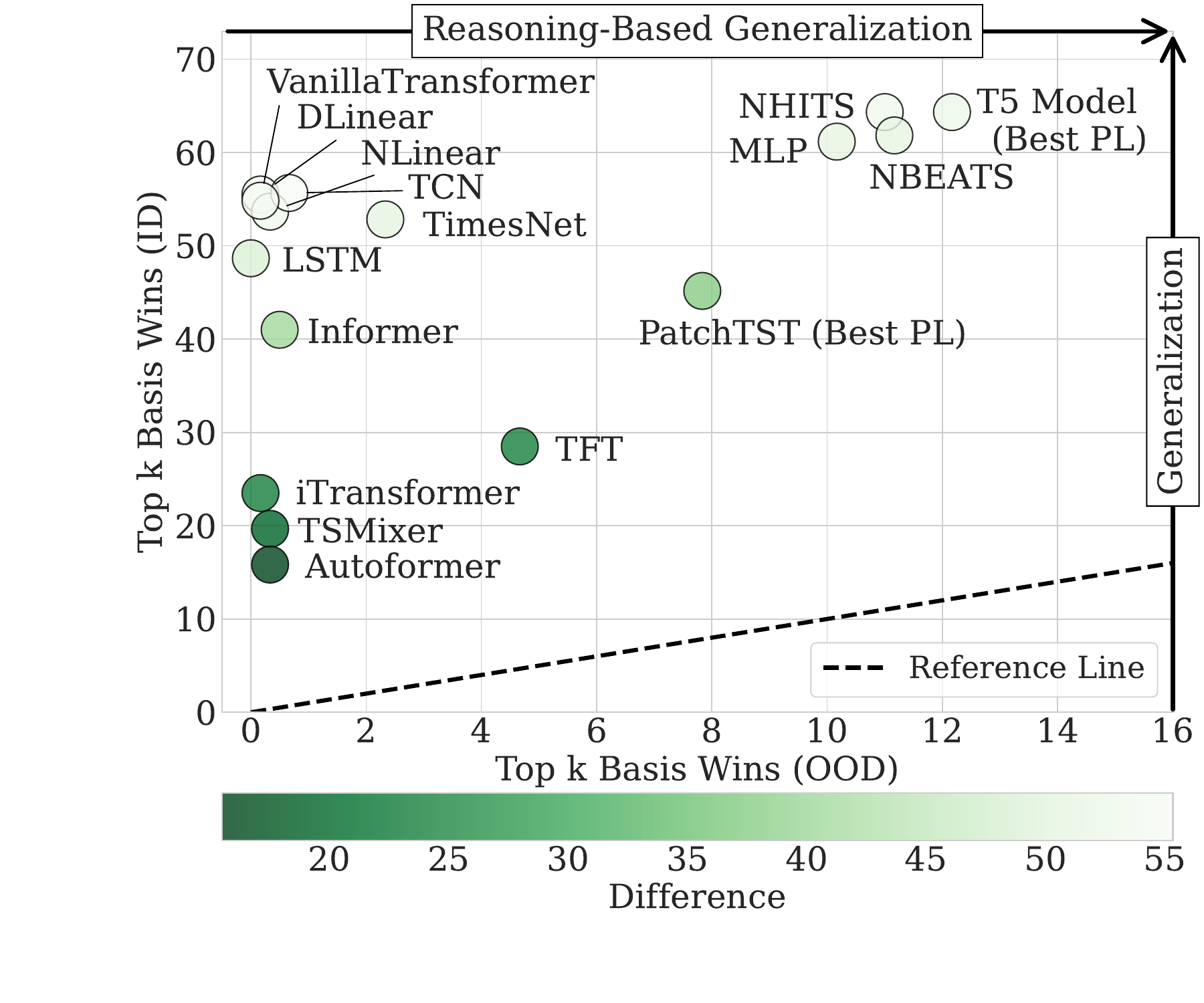}
        \caption{
            The average number of top $k$ compositions models outperform across datasets for ID and OOD data. The difference in wins is shown in green.
        }
        \label{fig:topkwins}
    \end{subfigure}
    \caption{
        Generalization performance of models on composition reasoning tasks across ID and OOD settings. 
        Patch-based Transformers and MLP-based models consistently outperform others in reasoning-focused generalization.
    }
    \label{fig:gen_comparison}
\end{figure}
\begin{figure*}[t!]
    \centering
    \begin{minipage}{0.4\textwidth}
        \centering
        \includegraphics[width=\textwidth, trim=0 0 310 0, clip]{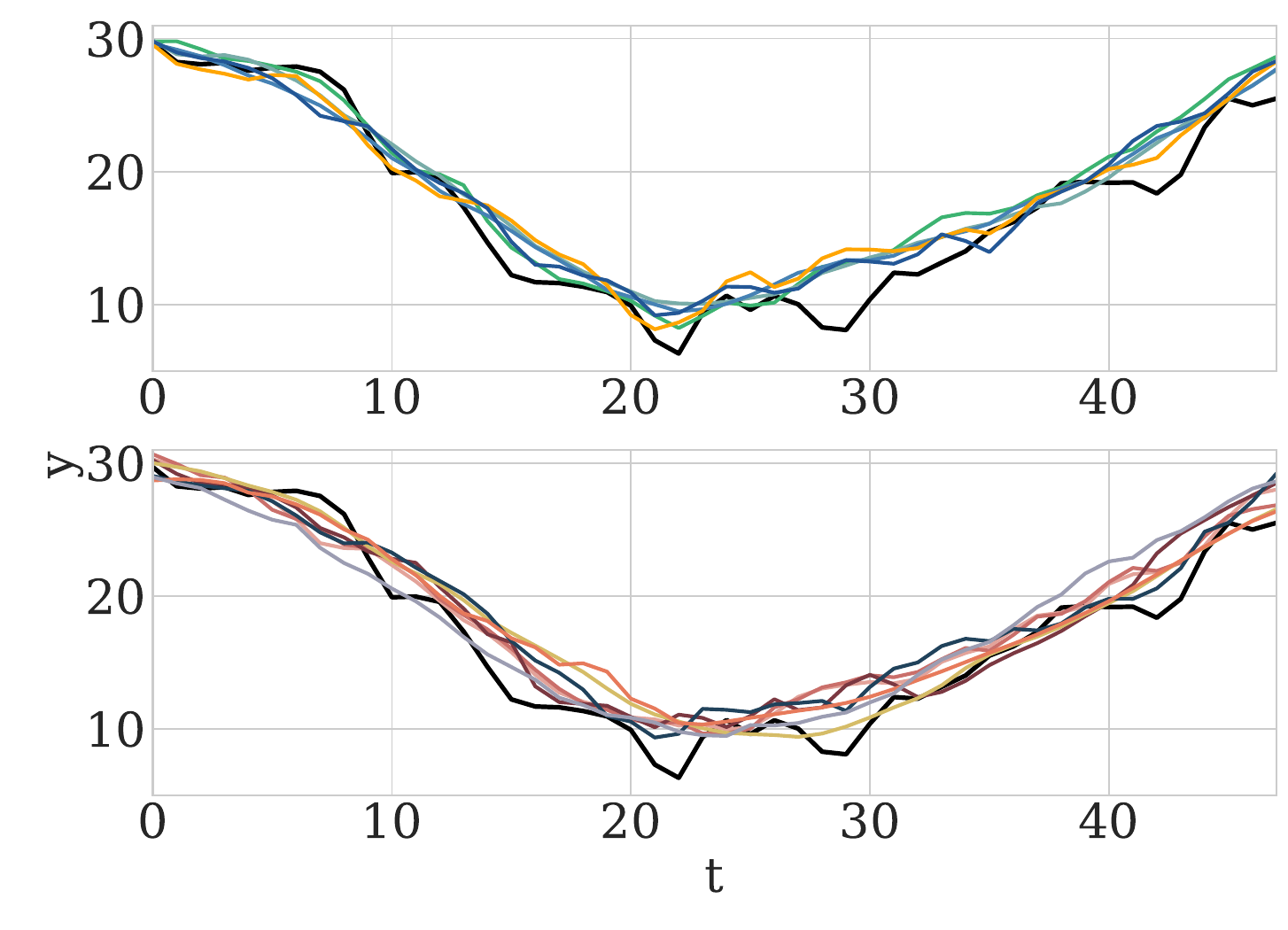}
        \subcaption{Model forecasts for in-distribution series.
        }
    \end{minipage}
    \hfill
    \begin{minipage}{0.58\textwidth}
        \centering
        \includegraphics[width=\textwidth]{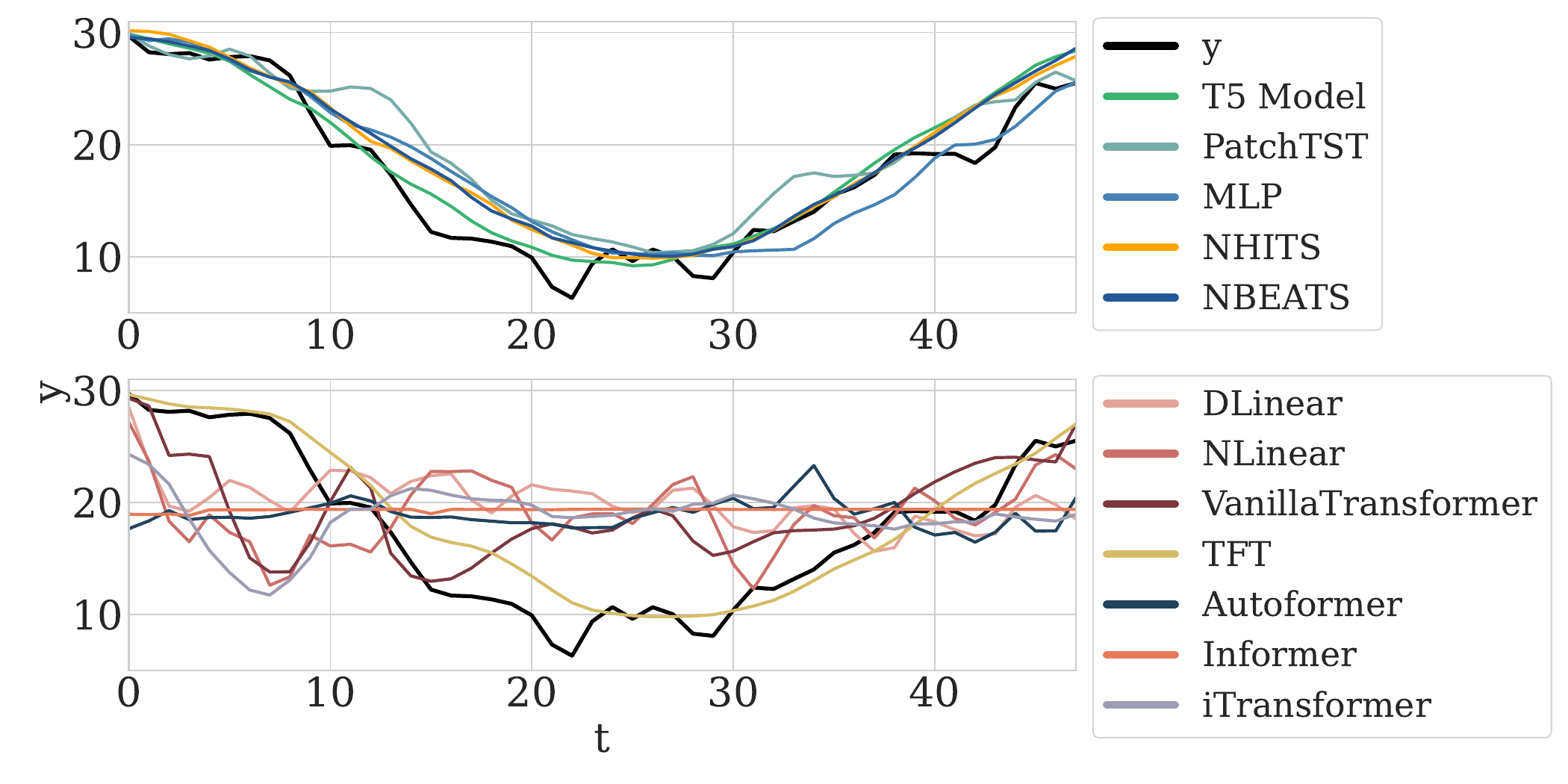}
        \subcaption{Model forecasts for out-of-distribution series}
        \label{fig:subseasonal_ood_example_forecasts_main}
    \end{minipage}

    \begin{minipage}{0.4\textwidth}
        \centering
        \includegraphics[width=\textwidth, trim=0 0 310 0, clip]{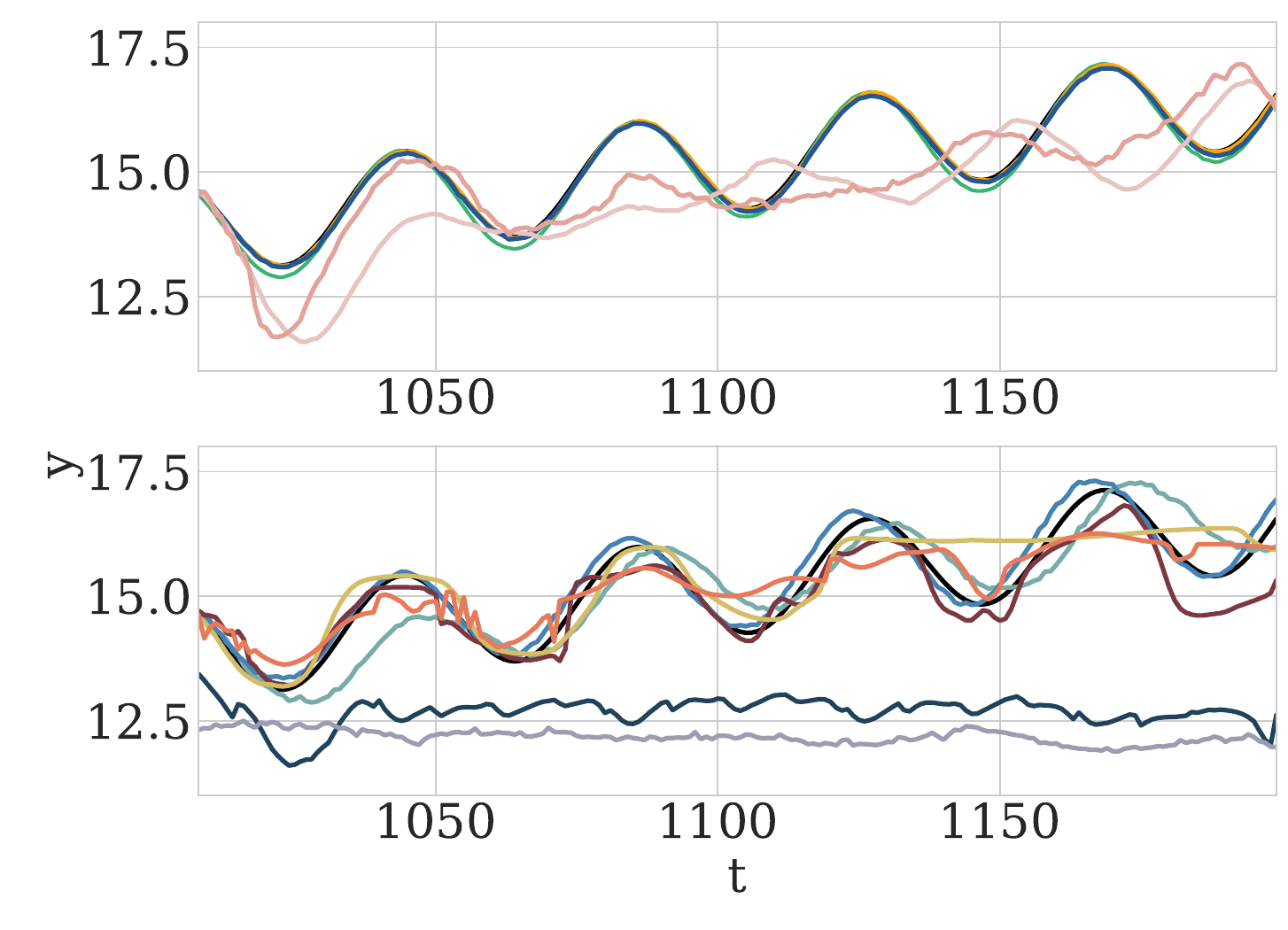}
        \subcaption{Model forecasts for in-distribution series
        }
    \end{minipage}
    \hfill
    \begin{minipage}{0.58\textwidth}
        \centering
        \includegraphics[width=\textwidth]{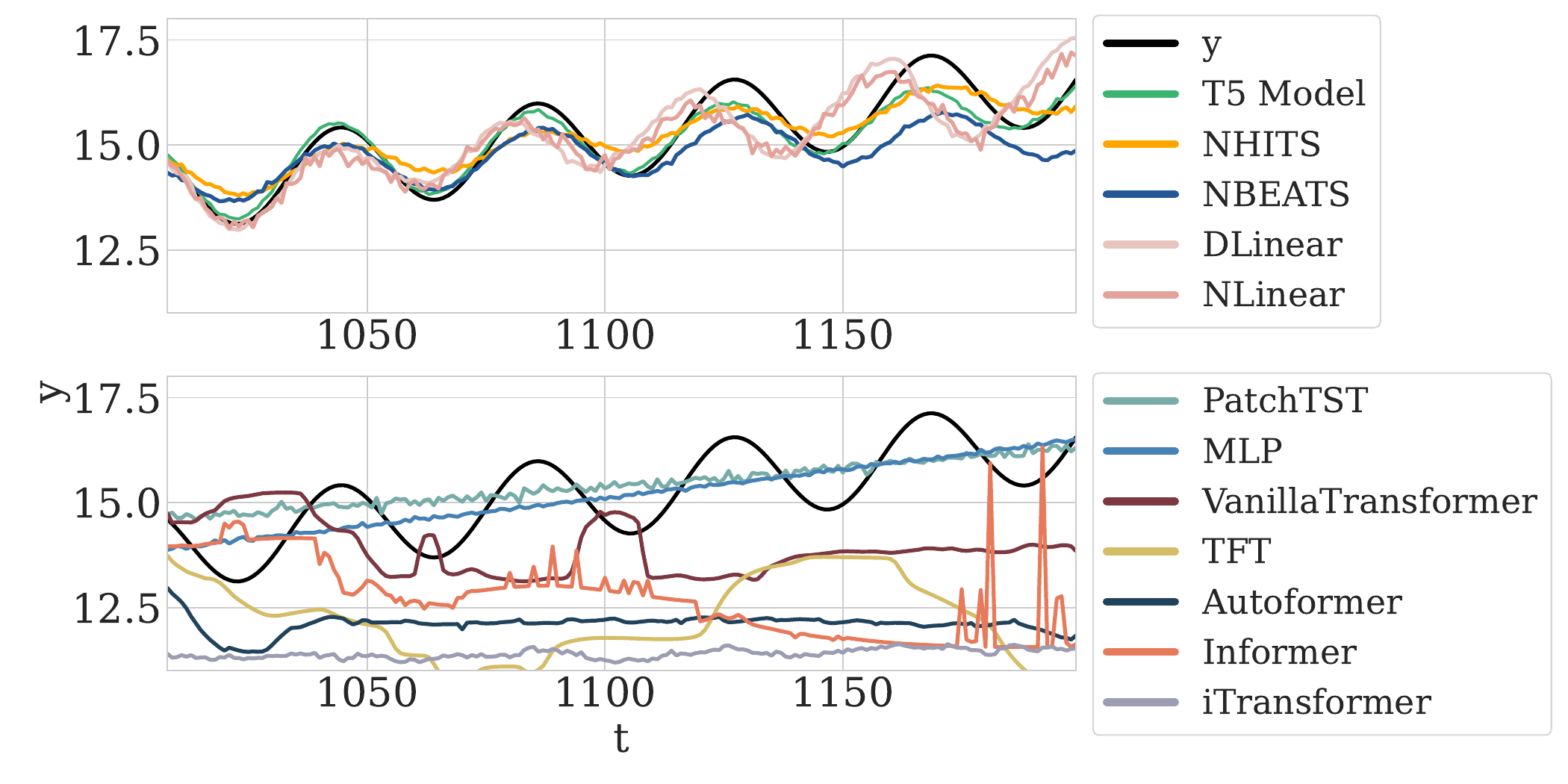}
        \subcaption{Model forecasts for out-of-distribution series}
        \label{fig:trend_ood_example_forecasts_main}
    \end{minipage}
    \caption{\textbf{(a, c)} Forecasts on the Subseasonal (top) and synthetic trend-seasonality (bottom) datasets using models trained with the traditional forecasting paradigm. \textbf{(b, d)} Forecasts from models trained with the compositional reasoning forecasting paradigm. The \Tfive and residualized MLP-based models (\NHITS, \NBEATS; top) generalize well to out-of-distribution series, unlike most Transformer variants and linear models (bottom), except for \DLinear and \NLinear on the synthetic dataset in \textbf{(d)}.}
\label{fig:ood_forecast_example_main}
\end{figure*}

\textbf{Patch-based Transformers \& MLP-based models show sparks of compositional reasoning.} Transformer-based models that `patch' input time series, including \PatchTST and \Tfive, as well as MLP-based models, such as \MLP, \NHITS and \NBEATS, achieve lower MAE as shown in Fig.~\ref{fig:error_scatter} and Fig.~\ref{fig:cd_baselines_component}. The \Tfive model exhibits the best generalization performance in both OOD forecasting tasks with \NHITS as the second-best model with an average MAE difference of 17\% across datasets. These models also achieve the highest number of wins as one of the top three models for compositional reasoning tasks, as shown in Table~\ref{tab:composition_main}. Standard deviation results for all models in Table~\ref{tab:composition_main} are provided in Table~\ref{tab:composition_baseline_results_table} in Appendix~\ref{apd:composition_full_table_results}. Example forecasts for these models on the subseasonal dataset are shown in Fig.~\ref{fig:subseasonal_ood_example_forecasts_main} and forecast examples for all 6 datasets are shown in Figs.~\ref{fig:forecast_examples1_apd} and~\ref{fig:forecast_examples2_apd} in Appendix~\ref{apd:forecast_examples}. 
The \Tfive and \NBEATS\ also achieve the highest Top-$k$ Basis Wins on OOD data as shown in Table~\ref{tab:composition_main}. 

\begin{wraptable}[15]{r}{5.5cm}
\caption{We compare model embedding similarity using CKA and find that models trained on both frequency (F) and trend (T) components (OOD) exhibit higher similarity to ID models trained on signal compositions than OOD models trained on either component alone.}
\label{wrap-tab:1}
\resizebox{0.4\textwidth}{!}{%
\begin{tabular}{lccc}
\toprule  
ID & OOD (F \& T) & OOD (F) & OOD (T) \\
\midrule
1.00 & \textbf{0.86} & 0.75 & 0.25 \\
-- & 1.00 & 0.67 & 0.16 \\
-- & -- & 1.00 & 0.19 \\
-- & -- & -- & 1.00 \\
\bottomrule
\end{tabular}}\label{tab:cka_similarity}
\end{wraptable}

\textbf{Compositionality is reflected in latent space embeddings.}
The compositional abilities of \Tfive, \NHITS, and \NBEATS extend to nonstationary data, as evidenced by their ranking among the top three models for forecasting trend and sinusoid compositions, as shown in Table~\ref{tab:composition_nonstationary_results_table}, with forecasts demonstrating these compositions, as illustrated in Fig.~\ref{fig:trend_ood_example_forecasts_main}. We find that the latent space representations of the \Tfive model trained jointly on both trend and frequency components exhibit higher similarity—measured using Centered Kernel Alignment (CKA)—to those of the \Tfive model trained on the ground truth signal compositions. In contrast, the \Tfive trained solely on trend or sinusoidal signals shows lower alignment with the ground truth activations, as shown in Table~\ref{tab:cka_similarity}. This suggests that the \Tfive model leverages information from both components to forecast the target signal, indicating compositionality in the latent space. Additionally, we evaluate the similarity of OOD \Tfive embeddings after fine-tuning on ID data and find that the model trained on both trend and frequency components retains the highest similarity to its original representations, compared to models trained on individual components alone. This implies that only minimal adaptation was required to achieve performance comparable to the ID model, further supporting the presence of compositional structure in the latent space prior to fine-tuning. More details are provided in section~\ref{apd:cka_embds}.

\input{tables/main_table}

\textbf{Zero-shot T5 in OOD scenarios can improve upon statistical model baselines trained on ID data.} Zero-shot forecasts for the \Tfive, \NHITS, \NBEATS, and \MLP models in OOD scenarios via the compositional reasoning forecasting paradigm can outperform moving average forecasts for \ARIMA models trained on ID data across all 6 datasets. These same models can also outperform \ETS model forecasts for 2 of 6 datasets. All other models can outperform both statistical baselines for 1 of 6 datasets. This result underscores the effectiveness of \Tfive and residualized MLP-based architectures in zero-shot forecasting for stationary OOD data. It also suggests that TSFMs pre-trained on various concepts could, with further development, eventually serve as sufficient options alongside efficient statistical models like ETS for effective generalization in OOD scenarios.

\textbf{Better ``reasoners" can also be computationally efficient.} The patch-based Transformers, such as \PatchTST and \Tfive, generally demonstrate lower FLOPs compared to many other Transformer, RNN, and CNN models. However, the \Tfive model exhibits the second-largest size in terms of total trainable parameters, whereas models such as \MLP, \NHITS, and \NBEATS have much smaller footprints, as shown in Fig.~\ref{fig:flops_model_comparison}. Despite \Tfive achieving the best generalization, the \NHITS\ and \NBEATS\ models are approximately 97\% more computationally efficient in terms of floating-point operations per second (FLOPs) and 86\% smaller in terms of trainable parameters than \Tfive. Performance versus FLOPs plots for both ID and OOD scenarios are provided in Appendix~\ref{apd:flops_plots}.

\textbf{Input patching enables Transformer-based TSFMs, while other tokenization methods degrade performance.} We observe that the \Tfive with fixed-length patches significantly outperforms other tokenization methods in compositional reasoning tasks, as shown in Fig.~\ref{fig:cd_tokenization_ablation_component}. Patch-based tokenization also ranks highest for experiments on ID series. For other \Tfive ablations, the model size and projection layer also demonstrate statistically significant differences across methods as shown in Figs.~\ref{fig:cd_size_ablation_component},
~\ref{fig:cd_tokenization_ablation_component}, ~\ref{fig:cd_proj_ablation_component}, respectively. Other design decisions, such as attention type, token length, positional encoding, loss functions, scaling functions, context length, and input decomposition did not demonstrate statistically significant differences as shown in Fig.~\ref{fig:cd_diagrams_apd} in Appendix~\ref{apd:cd_diagrams}.

\textbf{Architecture choices that perform well for ID series do not necessarily perform well for OOD series.} Some models demonstrate consistently higher performance across both ID and OOD data, such as models like \NHITS, \NBEATS, \MLP and the \Tfive. However, our ablation studies with the \Tfive reveal that certain architecture components choices that perform well for ID series do not necessarily perform well for OOD series. In particular, smaller models rank higher in performance than larger models in compositional reasoning tasks across datasets, whereas larger models outperform smaller models in the general train/test paradigm, as shown in Figs.~\ref{fig:cd_size_ablation_aggregate} and~\ref{fig:cd_size_ablation_component}. Similarly, although using residual networks for projection layers may improve performance on in-distribution (ID) tasks, it performs significantly worse on compositional reasoning tasks. CD diagrams for each architecture design component are included in Figs.~\ref{fig:cd_diagrams_baselines_apd} and~\ref{fig:cd_diagrams_apd} in Appendix~\ref{apd:cd_diagrams}. 

%% file: tables/main_table.tex
\begin{table*}[t!]
\centering
\caption{Mean Absolute Error (MAE) averaged over 3 random seeds for model forecasts. The out-of-distribution (OOD) column presents MAE results for models trained via the compositional reasoning forecasting paradigm. The in-distribution (ID) column presents MAE results for models trained via the traditional forecasting paradigm. \PatchTST\ and the \Tfive\ with the best patch length (PL) from Tables~\ref{tab:composition_baseline_results_table} and \ref{tab:composition_t5_results_table} are included. Best results are highlighted in \textbf{bold}, second best results are \underline{underlined}. The count of instances across datasets where the model ranks in the top three for performance is shown in the second to last column with non-zero entries in \textcolor{blue}{blue}. The average number of top $k$ compositions the model can outperform over the datasets is shown in the last column with nonzero entries in \textcolor{purple}{purple}.}
\label{tab:composition_main}
\resizebox{1.0\textwidth}{!}{
\begin{tabular}{l|cc|cc|cc|cc|cc|cc||cc|cc}
\toprule
\multirow{2}{*}{\textbf{Model}} & \multicolumn{2}{c}{\textbf{Synthetic Sinusoid}} & \multicolumn{2}{c}{\textbf{ECL}} & \multicolumn{2}{c}{\textbf{ETTm2}} & \multicolumn{2}{c}{\textbf{Solar}} & \multicolumn{2}{c}{\textbf{Subseasonal}} & \multicolumn{2}{c||}{\textbf{Loop Seattle}} & \multicolumn{2}{c}{\textbf{\small{Top 3 Win Count}}} & \multicolumn{2}{c}{\textbf{\small{Top-$k_{max}$ Basis Win}}} \\
% \cline{2-17}
{} & \textbf{OOD} & \textbf{ID} & \textbf{OOD} & \textbf{ID} & \textbf{OOD} & \textbf{ID} & \textbf{OOD} & \textbf{ID} & \textbf{OOD} & \textbf{ID} & \textbf{OOD} & \textbf{ID} & \textbf{\small{OOD}} & \textbf{\small{ID}} & \textbf{\small{OOD}} & \textbf{\small{ID}} \\
\hline
\ARIMA & -- & 15.538 & -- & 0.822 & -- & 0.332 & -- & 9.687 & -- & 7.855 & -- & 8.638 & \small{--} & \small{0} & \small{--} & \small{0.5} \\
\ETS & -- & 16.075 & -- & 0.105 & -- & 0.211 & -- & 1.730 & -- & 2.067 & -- & 5.575 & \small{--} & \small{0} & \small{--} & \small{0.8} \\ \midrule
\DLinear & 12.991 & 12.460 & 0.820 & \underline{0.103} & 0.330 & 0.135 & 9.925 & \textbf{1.555} & 8.042 & 1.496 & 9.085 & 4.293 & \small{0} & \small{\textcolor{blue}{2}} & \small{0.2} & \small{\textcolor{purple}{55.5}} \\
\NLinear & 13.287 & 13.056 & 0.801 & 0.104 & 0.325 & 0.136 & 9.681 & \underline{1.569} & 8.436 & 1.509 & 9.026 & 4.307 & \small{0} & \small{\textcolor{blue}{1}} & \small{0.3} & \small{\textcolor{purple}{53.7}} \\
\MLP & \underline{8.647} & 2.475 & \underline{0.283} & 0.106 & 0.253 & 0.114 & \underline{4.826} & 1.559 & 1.886 & 1.456 & 7.839 & 3.864 & \small{\textcolor{blue}{3}} & \small{\textcolor{blue}{1}} & \small{\textcolor{purple}{10.2}} & \small{\textcolor{purple}{61.2}} \\
\NHITS & 8.924 & \textbf{1.106} & 0.295 & \textbf{0.101} & \textbf{0.214} & \textbf{0.100} & 5.682 & 1.592 & 1.858 & \underline{1.135} & \underline{7.747} & 3.448 & \small{\textcolor{blue}{4}} & \small{\textcolor{blue}{4}} & \small{\textcolor{purple}{11.0}} & \small{\textcolor{purple}{64.3}} \\
\NBEATS & 8.907 & 1.383 & 0.294 & \underline{0.103} & \underline{0.216} & \underline{0.102} & 5.852 & 1.599 & \underline{1.840} & 1.177 & 7.763 & 3.479 & \small{\textcolor{blue}{5}} & \small{\textcolor{blue}{4}} & \small{\textcolor{purple}{11.2}} & \small{\textcolor{purple}{61.8}} \\
\TSMixer & 14.466 & 15.090 & 0.799 & 0.129 & 0.335 & 0.182 & 9.877 & 1.979 & 7.770 & 1.602 & 8.865 & 5.565 & \small{0} & \small{0} & \small{0.3} & \small{\textcolor{purple}{19.7}} \\  \midrule
\LSTM & 13.410 & 4.238 & 0.835 & 0.110 & 0.337 & 0.135 & 10.241 & 1.717 & 8.095 & 1.545 & 9.465 & 3.703 & \small{0} & \small{0} & \small{0} & \small{\textcolor{purple}{48.7}} \\  \midrule
\TCN & 11.478 & 3.833 & 0.837 & 0.106 & 0.339 & 0.135 & 9.868 & 1.642 & 6.170 & 1.234 & 8.792 & \underline{3.422} & \small{0} & \small{\textcolor{blue}{1}} & \small{0.7} & \small{\textcolor{purple}{55.7}} \\
\TimesNet & 9.788 & \underline{2.451} & 0.518 & 0.104 & 0.313 & 0.109 & 9.914 & 1.714 & 4.109 & 1.500 & 9.872 & 2.970 & \small{0} & \small{\textcolor{blue}{2}} & \small{\textcolor{purple}{2.3}} & \small{\textcolor{purple}{52.8}} \\  \midrule
\VanillaTransformer & 12.279 & 4.935 & 0.919 & 0.106 & 0.334 & 0.136 & 11.956 & 1.667 & 9.641 & 1.276 & 11.675 & 3.591 & \small{0} & \small{0} & \small{0.2} & \small{\textcolor{purple}{54.8}} \\
\iTransformer & 15.478 & 15.203 & 0.829 & 0.157 & 0.326 & 0.196 & 9.822 & 1.805 & 8.447 & 1.628 & 9.182 & 4.871 & \small{0} & \small{0} & \small{0.2} & \small{\textcolor{purple}{23.5}} \\
\Autoformer & 15.301 & 15.018 & 0.795 & 0.137 & 0.330 & 0.294 & 10.348 & 2.108 & 7.933 & 2.390 & 8.634 & 4.758 & \small{0} & \small{0} & \small{0.3} & \small{\textcolor{purple}{15.8}} \\
\Informer & 14.353 & 10.144 & 0.787 & 0.128 & 0.321 & 0.141 & 8.351 & 1.662 & 6.878 & 1.564 & 11.549 & 4.241 & \small{0} & \small{0} & \small{0.5} & \small{\textcolor{purple}{41.0}} \\
\TFT & 14.531 & 9.745 & 0.445 & 0.115 & 0.312 & 0.117 & 12.873 & 2.106 & 2.684 & 1.454 & 11.280 & 5.340 & \small{0} & \small{0} & \small{\textcolor{purple}{4.7}} & \small{\textcolor{purple}{28.5}} \\
\PatchTST\ (Best PL) & 10.696 & 6.959 & 0.482 & 0.122 & 0.247 & 0.138 & 5.726 & 1.633 & 2.185 & 1.659 & 7.965 & 3.653 & \small{\textcolor{blue}{1}} & \small{0} & \small{\textcolor{purple}{7.8}} & \small{\textcolor{purple}{45.2}} \\
\Tfive\ (Best PL) & \textbf{7.177} & 2.480 & \textbf{0.239} & \underline{0.103} & 0.259 & 0.103 & \textbf{3.899} & 1.578 & \textbf{1.714} & \textbf{1.097} & \textbf{6.589} & \textbf{3.351} & \small{\textcolor{blue}{5}} & \small{\textcolor{blue}{4}} & \small{\textcolor{purple}{12.2}} & \small{\textcolor{purple}{64.3}} \\
\bottomrule
\end{tabular}
}
\end{table*}

%% file: sections/05_discussion.tex
\begin{wrapfigure}[21]{r}{0.45\textwidth}
    \centering
    \vspace{-1.2em} % optional tweak to adjust vertical alignment
    \includegraphics[
        width=0.48\textwidth,
        trim=43 10 0 0,
        clip
    ]{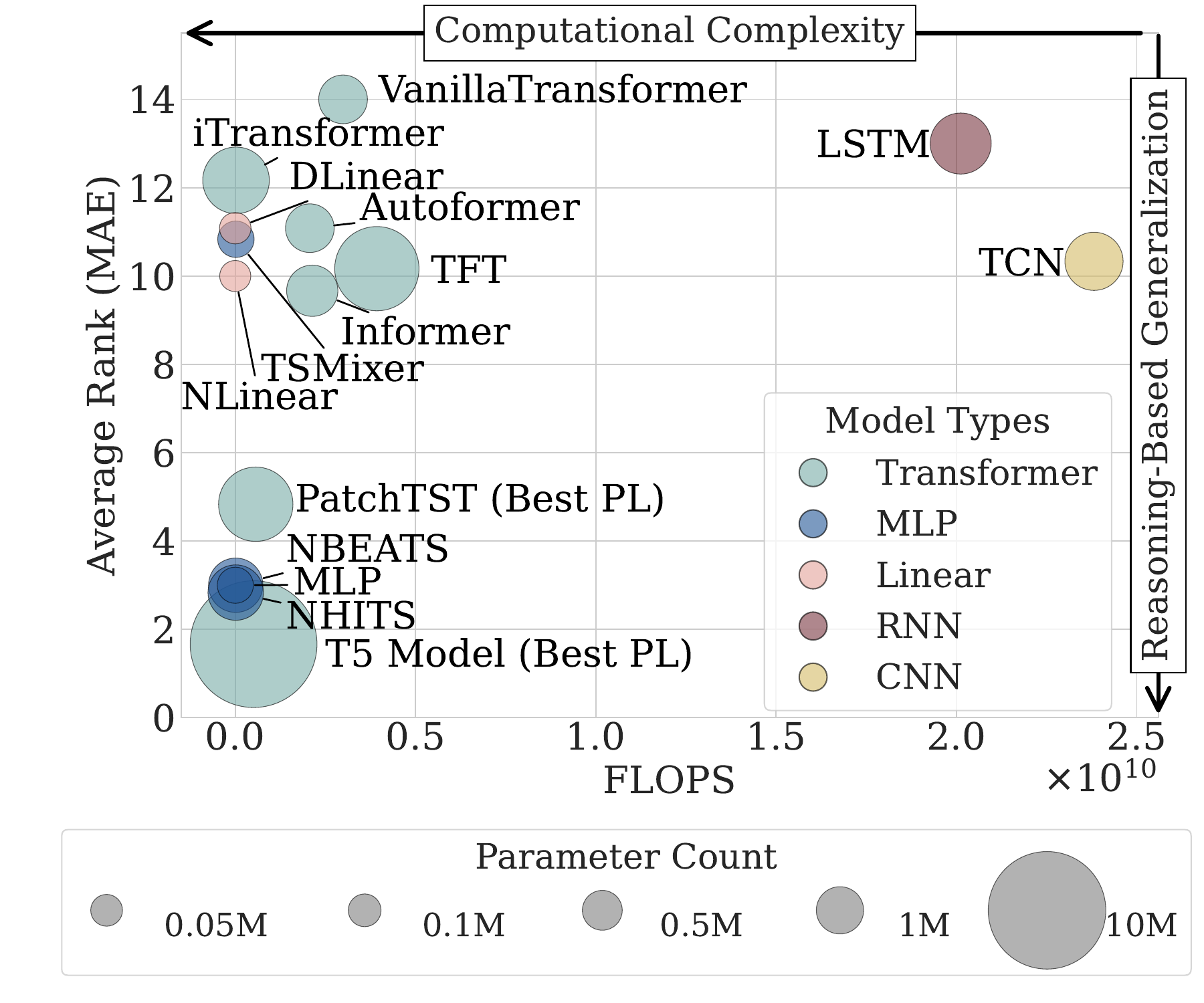}
    \caption{
        Comparison of average rank across datasets and random seeds versus model efficiency, measured in FLOPs. The marker size reflects the number of trainable parameters. 
    }
    \label{fig:flops_model_comparison}
\end{wrapfigure}

Through our experiments, we find that: (1) Transformer-based models adapted from LLMs, such as T5, and residualized MLP-based  architectures, such as 
\NHITS, rank among the best in generalization and compositional reasoning tasks; (2) input patching unlocks reasoning capabilities in Transformer-based TSFMs, whereas other input tokenization methods degrade \Tfive performance; (3) compositional reasoning capabilities do not necessarily scale with Transformer model size; and (4) only 7 out of 16 deep learning models demonstrate glimpses of compositional reasoning, which, while promising, highlights the need for future work to enhance compositional reasoning capabilities in TSFMs. We find that model forecast error in OOD scenarios is moderately correlated ($\sim 0.69$) with error in ID scenarios, motivating further exploration of potential links between compositional reasoning capabilities and generalization.

\input{figures/cd_figures}

Our findings provide critical insights for the future development of size- and parameter-efficient compositional reasoning-capable TSFMs. Sparks of compositional reasoning in the \Tfive and \NBEATS, despite their distinct architectures, highlight that reasoning abilities cannot be assumed across entire model classes (e.g., MLPs, Transformers) and may differ across architectural variants within them. Based on our experiments, practitioners aiming to develop more reasoning-capable TSFMs should explore input series tokenization strategies that preserve local information, such as `patching.' Also, hierarchical decomposition, as used in \NBEATS\ and \NHITS, simplifies complex temporal patterns into structured residuals for additive modeling, demonstrating robust generalization. Architectures with concept-specialized stacked processing, such as \NHITS\ and \NBEATS, show greater potential for enhancing compositional reasoning compared to approaches like moving average filters for extracting seasonality, as evidenced by the performance of \DLinear, \Autoformer, and the moving average ablation of \Tfive\ (Table~\ref{tab:composition_t5_results_table}). Furthermore, future research should prioritize innovating new TSFM component methods over simply scaling model size. While the \texttt{T5-efficient-base} model ranks the highest in performance for ID forecasting tasks, increased model size does not translate to improved compositional reasoning performance as shown by the statistically significant difference between \texttt{t5-mini} and \texttt{t5-base} in Fig.~\ref{fig:cd_size_ablation_component_main}. 

\vspace*{-3pt}
\paragraph{Limitations and Future Work.}\label{apd:limitations}
This study has two main limitations related to data and model ablations. First, due to computational constraints, we evaluated compositional capabilities using five real-world open-source datasets. While informative, additional datasets could reveal further differences between architectural components. Second, ablations focus solely on the best-performing \Tfive. Future work can extend these to other TSFM architectures and explore new components. To facilitate this, we provide open-source code and benchmarks for easy integration of datasets and model variations.

% Future work can also explore alternative hyperparameter configurations and convergence-based stopping criteria for model training, beyond early stopping with a fixed number of validation iterations. 

%% file: figures/cd_figures.tex
\begin{wrapfigure}[14]{l}{0.52\textwidth}
    \centering
    \vspace{-1.5em} % Optional: adjust vertical position
    % \begin{subfigure}[t]{0.48\textwidth}
    %     \centering
    %     \includegraphics[width=\textwidth, trim=9 0 0 0, clip]{images/cd_baselines_component.pdf}
    %     \caption{Model Choice}
    %     \label{fig:cd_baselines_component_main}
    % \end{subfigure}
    % \hfill
    \begin{subfigure}[t]{0.48\textwidth}
        \centering
        \includegraphics[width=\textwidth, trim=9 0 0 0, clip]{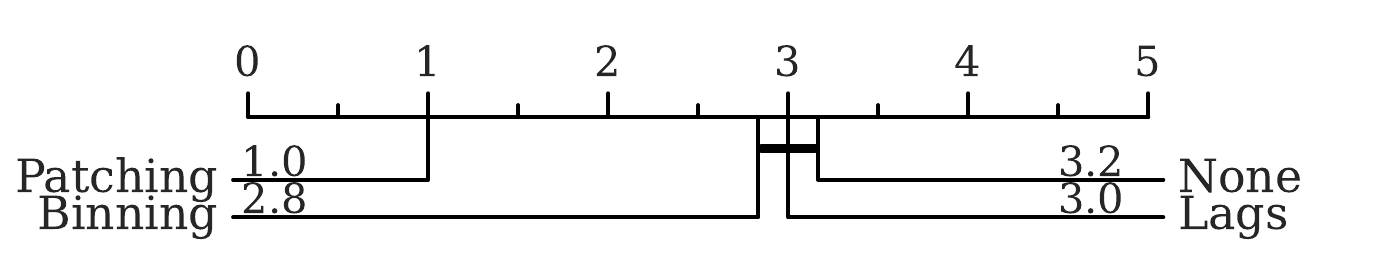}
        \caption{\Tfive\ Tokenization}
        \label{fig:cd_tokenization_ablation_component_main}
    \end{subfigure}

    \vspace{0.5em}

    \begin{subfigure}[t]{0.48\textwidth}
        \centering
        \includegraphics[width=\textwidth, trim=40 0 0 0, clip]{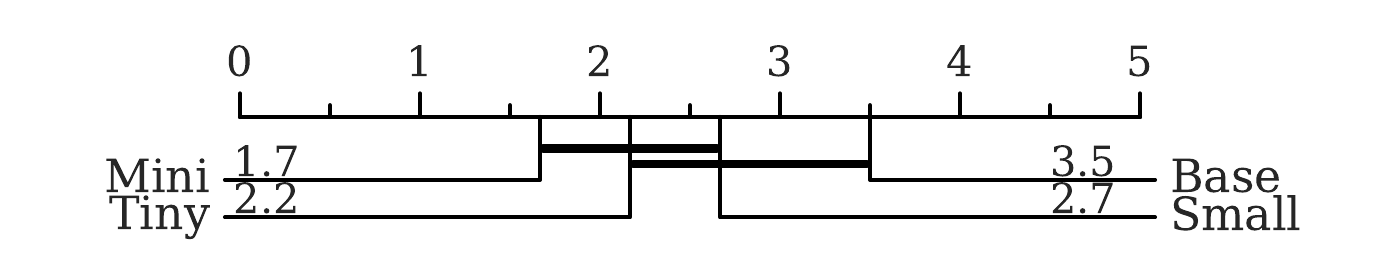}
        \caption{\Tfive\ Size}
        \label{fig:cd_size_ablation_component_main}
    \end{subfigure}
    \hfill

    \caption{
        \textbf{(a, b)} Performance ranks on compositional reasoning tasks across datasets for \Tfive\ design choices. Lower ranks are better; thick lines denote non-significant differences.
    }
    \label{fig:main_comparison}
\end{wrapfigure}

%% file: sections/appendixA.tex
\subsection{Reproducibility Statement}\label{section:reproducibility_statement}

The models were implemented using the \texttt{Neuralforecast} library~\citep{olivares2022library_neuralforecast}. The open-source code for generating critical difference diagrams is available at \url{https://github.com/hfawaz/cd-diagram}. All datasets used in this study are publicly accessible and can be downloaded by following the instructions at \url{https://github.com/SalesforceAIResearch/gift-eval}. Both \texttt{Neuralforecast} and the Gift-Eval data repository are licensed under the Apache License 2.0. Model training and evaluation were conducted on a computing cluster equipped with 128 AMD EPYC 7502 CPUs, 503 GB of RAM, and 8 NVIDIA RTX A6000 GPUs, each with 49 GiB of RAM. To ensure reproducibility, the model implementations, training framework, and datasets are open-source, supporting future research on model reasoning. The complete codebase is available at \url{https://github.com/PotosnakW/neuralforecast/tree/tsfm_reasoning}. 

\subsection{Impact Statement}\label{section:impact_statement}
This study investigates whether time series models can perform implicit reasoning during zero-shot inference by synthesizing learned concepts to generalize to more complex data patterns. Our findings reveal that certain models can generalize effectively in well-designed OOD scenarios, highlighting reasoning abilities that go beyond basic pattern memorization. As this work is exploratory and not tied to specific applications, we do not foresee any negative societal impacts. In contrast, our insights could aid in the development of more data- and computationally-efficient deep learning architectures. Additionally, our results can help identify the limitations of time series models, ensuring that they are not used in scenarios where poor generalization is likely.

\subsection{Extended Related Work}\label{apd:extend_prior_work}

\paragraph{Reasoning in LLMs.} While models can recall individual facts well in single-hop Q\&A tasks, they generally struggle with multi-hop reasoning that requires `chain-of-thought' logic. \citet{soheeyang2024multihopreasoning} found evidence of latent multi-hop reasoning for specific fact compositions, but noted it is highly contextual. \citet{wang2024grokked} demonstrated that Transformers can learn implicit reasoning, but only after extended training beyond typical overfitting thresholds. Composition tasks have also been studied in machine translation, testing whether models can generalize learned command components to new conjunctions, such as repeating actions \citep{lake2018generalization}.

Other types of reasoning studied in LLMs include \textit{comparison} and \textit{inverse search} \citep{allenshu2023physics3.2, wang2024grokked}.  \textit{Comparison} reasoning involves models evaluating two or more entities to make judgments to answer prompts of whether the attribute value of one entity is greater or smaller than that of another \citep{allenshu2023physics3.2, wang2024grokked} or to complete numerical answers with appropriate `greater-than' logic \cite{hanna2023gptgreaterthan}. Inverse search tests a model's ability to generate predictions in the reverse order of the training task. For example, this could involve applying the model to identify an entity based on its attributes when it was originally trained to predict the attributes of entities. Allen-Zhu et al. found that generative models struggle with inverse search unless trained specifically for it \citep{allenshu2023physics3.2}.

\paragraph{Time Series Reasoning with LLMs.} Prior work has explored the use of LLMs for time series forecasting \citep{hartvigsen2024llmforecasting}, including efforts to determine whether textual context can enhance forecasting performance \citep{althoff2024llmtimeseriesreasoningstruggle, Williams2025llmtimeseriestext}. Other studies have investigated LLMs' ability to interpret time series concepts—such as slope and frequency—and translate them into meaningful natural language features \citep{cai2024timeseriesexam, chow2024llmstimeseriesreasoning}. Additionally, some research has examined LLMs’ capacity for etiological reasoning, or identifying the underlying scenario that generated a given time series \citep{althoff2024llmtimeseriesreasoningstruggle}. These works differ from ours in that they focus on LLMs’ reasoning about or with time series. In contrast, our work investigates whether time series forecasting models themselves can reason through the composition of time series patterns.

\paragraph{Frequency-Based Modeling.} While not focused on model reasoning, prior work in time series forecasting has used frequency-based features to improve performance. \NBEATS reconstructs and forecasts time series by predicting Fourier coefficients needed for reconstruction, while \BasisFormer similarly predicts these coefficients \citep{ni2024basisformer}. Yang et al. leverage the Fourier basis expansion to provide frequency features directly in the time domain \citep{yang2024rethinkingfourier}. \TimesNet utilizes the Fast Fourier Transform (FFT) to identify and extract significant frequency components indicative of periodic patterns \citep{wu2023timesnettemporal2dvariationmodeling}. Previous models aim to improve architectures by incorporating frequency-based features during training. In contrast, our study focuses on evaluating whether models can logically generalize to unseen periodic patterns through the synthesis of frequency-based features—specifically, sine and cosine basis functions.

\subsection{Models}\label{apd:first_models}

\noindent\textbf{DLinear (\DLinear)} - Linear layers are employed to model the trend and seasonal components, with the decomposition achieved through a moving average filter to separate the time series into its trend and seasonal components \citep{zeng_2023_dlinear}. 

\noindent\textbf{NLinear (\NLinear)}  - A Linear layer is employed to model the series. The model first subtracts the input by the last value of the sequence. Then, the input goes through a linear layer, and the subtracted part is added back before making
the final prediction \citep{zeng_2023_dlinear}. 

\noindent\textbf{Multi Layer Perceptron (\MLP)} - A neural network architecture composed of stacked Fully Connected Neural Networks trained with backpropagation \citep{nair2010_mlp, fukushima1975_mlp, rosenblatt1958_mlp}. 

\noindent\textbf{Neural Hierarchical Interpolation for Time Series (\NHITS)} - A deep learning model that applies multi-rate input pooling, hierarchical interpolation, and backcast residual connections together to generate additive predictions with different signal bands \citep{challu_olivares2022_nhits}.

\noindent\textbf{Neural Basis Expansion Analysis for Time Series (\NBEATS)} - A deep learning, MLP-based model that leverages both backward and forward residual connections. The network can decomposes the input signal into trend and seasonality components using polynomial and harmonic basis projections or substitute the polynomial and harmonic basis for identity basis based on the specified configuration \citep{oreshkin2020nbeats}.

\noindent\textbf{Time-Series Mixer (\TSMixer)} - TSMixer is a deep learning, MLP-based model that uses stacked mixing layers to learn and combine temporal and cross-sectional representations through mixing operations along both time and feature dimensions \citep{chen2023tsmixer}.

\noindent\textbf{Long Short-Term Memory Recurrent Neural Network (\LSTM)} - A recurrent neural network (\RNN) architecture that transforms hidden states from a multi-layer \LSTM\ encoder into contexts which are used to generate forecasts using \MLP s  \citep{sak2014_lstm}.

\noindent\textbf{Temporal Convolution Network (\TCN)} - A 1D causal-convolutional network architecture that transforms hidden states into contexts which are used as inputs to \MLP\ decoders to generate forecasts. Causal convolutions are used to generate contexts by convolving the prediction at time $t$ only with elements from time $t$ and earlier \citep{bai2018_tcn, oord2016_tcn}.

\noindent\textbf{TimesNet (\TimesNet)} - A deep learning architecture that transforms the original 1D time series into a set of
2D tensors based on multiple periods to capture intra- and inter-period variations modeled by 2D kernels \citep{wu2023timesnettemporal2dvariationmodeling}.

\noindent\textbf{VanillaTransformer (\VanillaTransformer)} - An encoder-decoder architecture with a multi-head attention mechanism that uses autoregressive features from a convolution network, window-relative positional embeddings from harmonic functions, and absolute positional embeddings from calendar data. An MLP decoder outputs time series predictions in a single pass \citep{vaswani_2021_attentionisallyouneed, zhou2021informerefficienttransformerlong}.

\noindent\textbf{iTransformer (\iTransformer)} - An attention-based deep learning architecture that applies attention and feed-forward networks to inverted dimensions by embedding time points into variate tokens. The attention mechanism capture multivariate correlations while the feed-forward network learns nonlinear representations for each token \citep{liu2024itransformerinvertedtransformerseffective}.

\noindent\textbf{Autoformer (\Autoformer)} - An encoder-decoder architecture with a multi-head attention mechanism that uses autoregressive features from a convolution network and absolute positional embeddings from calendar data. Decomposed trend and seasonal components are obtained using a moving average filter and an Auto-Correlation mechanism is used to identify periodic dependencies and aggregate similar sub-series \citep{wu_2021_autoformer, vaswani_2021_attentionisallyouneed}.

\noindent\textbf{Informer (\Informer)} - An encoder-decoder architecture with a multi-head attention mechanism that has three key features: a ProbSparse self-attention mechanism with \(O(L \log L)\) complexity, a self-attention distilling process, and an MLP decoder that outputs time-series predictions in a single pass. It uses autoregressive features from a convolution network, window-relative positional embeddings from harmonic functions, and absolute positional embeddings from calendar data \citep{zhou2021informerefficienttransformerlong, 
vaswani_2021_attentionisallyouneed}.

\noindent\textbf{Temporal Fusion Transformer (\TFT)} - An attention-based deep learning architecture that learns temporal relationships at different scales using \LSTM s for local processing and self-attention layers to model long-term dependencies. It also leverages variable selection networks and a series of gating layers to suppress unnecessary processing in the architecture \citep{lim2021_tft}. 

\noindent\textbf{Patch Time Series Transformer (\PatchTST)} - An encoder-only, multi-head attention-based architecture that separates input time series into sub-series level patches as input tokens. Each channel is dedicated to a single univariate time series, and all channels use the same embedding and Transformer weights. \cite{nie2023patchtst}

\noindent\textbf{T5-Efficient} - A series of transformer-based architectures developed to enhance the efficiency of the original T5. Code and configuration files are open-source and obtained from the HuggingFace library \citep{wolf_2020_hgtransformers}.

\subsection{TSFM Model Design Decision Comparison}\label{apd:TSFM_model_decision_decisions_comparison}
TSFMs involve combinations of various architectural component design decisions. We outline the design decisions of various TSFMs, including Chronos \citep{ansari2024chronos, ansari2024chronos}, LagLlama \citep{rasul2024lagllama}, Moirai \citep{moirai2024}, MOMENT \citep{goswami2024moment}, Timer \citep{liutimer}, TimesFM \citep{das2024TimesFM}, and TinyTimeMixers \citep{ekambaram2024ttms} in Table~\ref{tab:tsfm_design_decisions_comparison}. These decisions encompass architectural choices, input series tokenization methods, projection layer types, and positional encodings, among others.
\input{tables/TSFM_decision_decisions_table}

\subsection{TSFM Design Decision Experiments}\label{apd:experiment_table}
Comparing the performance of open-source TSFMs on compositional reasoning tasks alone does not allow us to isolate performance improvements attributable to specific components. To address this, we implement a deep learning Transformer architecture with a modular design, built on the open-source T5-efficient backbone \citep{wolf_2020_hgtransformers} to enable controlled experimentation with common TSFM design choices. An outline of the various types of components and methods for each design decision ablation is shown in Table~\ref{tab:tsfm_design_decisions_experiments}.
\input{tables/experiment_table}

\subsection{Synthetic Composition Benchmarks}\label{apd:synthetic_data_parameters}
We leverage open-source code from \MOMENT ~\citep{goswami2024moment} to generate synthetic sinusoidal time series with varying frequency $b$ and amplitude $a$: $\mathcal{F}_{\text{stationary}} = \{\mathcal{Y} : \mathcal{Y}(t) = a\sin(b 2\pi t), a\cos(b 2\pi t)\}$, where $a \in [1, 32]$ and $b \in [3, 32]$.

To take a first step towards evaluating compositional reasoning capabilities for nonstationary data, we introduce a synthetic dataset that includes nonstationary trend signals. The synthetic dataset, characterized by controlled variations in frequency and amplitude as well as trend, and serves as a simplified benchmark for testing compositional capabilities of time series forecasting algorithms with nonstationary and periodic series. We consider a set of sinusoidal functions and trend functions, $\mathcal{F}_{\text{nonstationary}} = \{mt\}$, where $m \in [-32, 32]$. For $N$ time series each with $2$ signal compositions, one sinusoid and one trend signal, the functions $\{\sin, \cos\}$ and parameters $a$, $b$, and $m$ are randomly sampled $N=100$ times without replacement to generate the time series for each of the $N=100$ series. For Trend Dataset \#1, $m \in [-32, 32]$. In Trend Dataset \#2, the training set is limited to $m \in [1, 32]$, while the test set covers $m \in [-32, -1]$, allowing us to evaluate whether the models can capture compositions with reversed patterns.

\subsection{Real-World Data}\label{apd:realworld_data_parameters}
Table \ref{tab:realworld_dataset_characteristics} summarizes the characteristics of the real-world datasets used in our experiments. The ``Pre" column denotes the dataset properties prior to preprocessing, while ``Post" reflects the properties after preprocessing. For the real-world datasets, we selected a horizon length of 48, following the GIFT-Eval benchmark \citep{aksu2024giftevalbenchmark}. Please see section \ref{section:methods_data} for more details regarding data preprocessing.

\begin{table}[ht]
\centering
\caption{Dataset characteristics including domain, frequency, number (\#) of series and number of forecast targets. The number of unique series in each dataset before (Pre) and after (Post) preprocessing are shown, as we selected 100 subseries to ensure consistent dataset sizes with an equal number of series and a training length of 1056 timesteps.}
\resizebox{1.0\textwidth}{!}{\begin{tabular}{r|ccccccccc}
\toprule
\textbf{Dataset} & \textbf{Domain} & \textbf{Frequency} & \textbf{\# Series (Pre)} & \textbf{\# Series (Post)} & \textbf{\# Targets} & \textbf{\# Subseries (Post)}  & \textbf{Series Length (Post)} & \textbf{Prediction Horizon} \\ 
\midrule
Synthetic Sinusoid & N/A & N/A & N/A & N/A & 1 & 100 & 1,200 & 192 \\ 
ECL & Energy & Hourly & 321 & 47 & 1 & 100 & 1056 & 48 \\ 
ETTm2 & Energy & 15 minute & 1 & 1 & 7 & 100 & 1056 & 48 \\ 
Solar & Energy & Hourly & 5166 & 99 & 1 & 100 & 1056 & 48 \\ 
Subseasonal & Climate & Weekly & 862 & 56 & 4 & 100 & 1056 & 48 \\ 
Loop Seattle & Transportation & Hourly & 323 & 36 & 1 & 100 & 1056 & 48 \\ 
\bottomrule
\end{tabular}}
\label{tab:realworld_dataset_characteristics}
\end{table}

\subsection{Model Training and Parameters}\label{apd:first_hyperparameters}
All models are open-source. Models trained on synthetic data can be found in the \texttt{Neuralforecast} library \citep{olivares2022library_neuralforecast}. 
All models were trained and evaluated on a computing cluster consisting of 128 AMD EPYC 7502 CPUs, 503 GB of RAM, and 8 NVIDIA RTX A6000 GPUs each with 49 GiB RAM. The synthetic datasets used in our study will be released publicly with the full paper. The T5 backbone and configuration files are available in the Hugging Face library \citep{wolf_2020_hgtransformers}. 

We train the \Tfive with the following parameters outlined in Table~\ref{tab:hyperparameters} to ensure consistent evaluation across architectures. The training loss functions used in the `loss function' ablation experiment include Mean absolute Error (MAE), Mean Squared Error (MSE), Huber Loss, and Distribution Loss with the Student's t-distribution. $H$ refers to the forecast horizon, $t$ refers to the time point at which forecasts are generated, $\textbf{y}$ refers to the target signal values, and $\hat{\textbf{y}}$ refers to the model's predicted values:
\[
\begin{minipage}{0.35\textwidth}
\begin{align*}
    \mathrm{MAE}(\textbf{y}_i, \hat{\textbf{y}}_i) &= \frac{1}{H}\sum_{i=t+1}^{t+H}|\textbf{y}_i - \hat{\textbf{y}}_i|,\\
    \mathrm{MSE}(\textbf{y}_i, \hat{\textbf{y}}_i) &= \frac{1}{H}\sum_{i=t+1}^{t+H}(\textbf{y}_i - \hat{\textbf{y}}_i)^2,
\end{align*}
\end{minipage}%
\hfill
\begin{minipage}{0.64\textwidth}
\begin{align*}
    \mathrm{Huber}(\textbf{y}_i, \hat{\textbf{y}}_i, \delta) &= 
    \begin{cases}
    \frac{1}{2}(\textbf{y}_i - \hat{\textbf{y}}_i)^2, &\text{if } |\textbf{y}_i - \hat{\textbf{y}}_i| \leq \delta, \\
    \delta \cdot (|\textbf{y}_i - \hat{\textbf{y}}_i| - \frac{1}{2}\delta),  &\text{otherwise}.
    \end{cases}, \\
    \mathrm{DistributionLoss}(\theta) &= -\log(P(\textbf{y}_i | \theta)),
\end{align*}
\end{minipage}%
\hfill
\]
where the probability density function (PDF) of the Student's \emph{t}-distribution with location parameter \(\mu\), scale parameter \(\sigma\), and degrees of freedom \(\nu\) is given by:

\[
\mathbb{P}(\textbf{y}_\tau \mid \mu, \sigma, \nu) = \frac{\Gamma\left(\frac{\nu + 1}{2}\right)}{\sqrt{\nu \pi} \, \Gamma\left(\frac{\nu}{2}\right) \sigma} \left( 1 + \frac{1}{\nu} \left(\frac{\textbf{y}_\tau - \mu}{\sigma}\right)^2 \right)^{-\frac{\nu + 1}{2}},
\]

\begin{table}[t!]
    \caption{Common hyperparameter search space. \textsuperscript{\textdagger} Based on the T5-efficient-tiny architecture}
    \centering
    \label{tab:hyperparameters}
    \begin{tabular}{rl} 
    \toprule
      \textbf{Hyperparameter} & \textbf{Considered Values}\\
      \hline
      Input size :& 256 (192 for tokenization experiments) \\
      Learning rate : & 1e-4 \\
      Batch size :& 4 \\
      Windows batch size :& 256 \\
      Dropout :& 0.0 \\
      Training steps :& 10000 \\
      Validation check steps :& 100 \\
      %Early stop patience steps & 5 \\
      Early stop patience steps :& 20 \\
      Random seed :& \{1, 5, 10\} \\
      Hidden size :& 256\textsuperscript{\textdagger} \\
      Linear hidden size :& 1024\textsuperscript{\textdagger} \\
      Model Encoder layers :& 4\textsuperscript{\textdagger} \\
      Model Decoder layers :& 0 \\
      Number of attention heads :& 4\textsuperscript{\textdagger} \\
      Patch length :& 96 for fixed-length patches (1 for other tokenization methods) \\
      Stride :& 8 for fixed-length patches (1 for other tokenization methods) \\
      \bottomrule
    \end{tabular}
  \centering
\end{table}

\subsection{Evaluation Metrics}\label{apd:eval_metrics}
We use \textbf{Mean Absolute Error (MAE)} to evaluate model performance. Here, $H$ refers to the forecast horizon, $t$ refers to the time point at which forecasts are generated, $\textbf{y}$ refers to the target signal values, and $\hat{\textbf{y}}$ refers to the model's predicted values:\\
\begin{align}
    \mathrm{MAE}(\textbf{y}_{\tau}, \hat{\textbf{y}}_{\tau}) &= \frac{1}{H}\sum_{{\tau}=t+1}^{t+H}|\textbf{y}_{\tau} - \hat{\textbf{y}}_{\tau}|.
\end{align}

\subsection{Significance Tests}\label{apd:significance_tests}

For comparisons of three or more models, the Friedman test ($\alpha=0.2)$ is used to determine whether there are significant performance differences, testing the null hypothesis that all methods perform equally on average. If the null hypothesis is rejected, a post-hoc analysis is conducted using the Wilcoxon signed-rank test with Holm correction. The Wilcoxon test highlights specific pairs of models with statistically significant performance differences, while the Holm correction controls for Type I errors across multiple hypothesis tests. 

%% file: tables/TSFM_decision_decisions_table.tex
\begin{table*}[h]
\centering
\caption{Key design choices of widely used time series foundation models. All these models are based on the Transformer Architecture, with the exception of TinyTimeMixers~\citep{ekambaram2024ttms} which is an all-MLP architecture. All models which use Reversible Instance Normalization (RevIn), do so without the affine learnable parameters.}
\label{tab:tsfm_design_decisions_comparison}
\resizebox{1.0\textwidth}{!}{
\begin{tabular}{r c | c | c | c | c | c | c | c }
\toprule
\multicolumn{2}{c|}{\textbf{TSFM Component}} & \textbf{\texttt{Chronos}} & \textbf{\texttt{LagLlama}} & \textbf{\texttt{Moirai}} & \textbf{\texttt{MOMENT}} & \textbf{\texttt{Timer}} & \textbf{\texttt{TimesFM}} & \textbf{\texttt{TinyTimeMixers}} \\
\midrule
\multirow{3}{*}{\textbf{Architecture}}
        & Encoder-Only & 
            - & 
            \textcolor{green}{\checkmark} & 
            \textcolor{green}{\checkmark} & 
            \textcolor{green}{\checkmark} & 
            - &
            - &
            - \\
        & Decoder-Only & - & \textcolor{green}{\checkmark} & 
        - & 
        - & 
        \textcolor{green}{\checkmark} & 
        \textcolor{green}{\checkmark} & 
        -\\
        & Encoder-Decoder & \textcolor{green}{\checkmark} & - & - & - & - & - & \textcolor{green}{\checkmark}\\
\hline
% \multirow{2}{*}{\textbf{Backbone}}
%         & Transformer & 
%             \textcolor{green}{\checkmark} & 
%             \textcolor{green}{\checkmark} & 
%             \textcolor{green}{\checkmark} & 
%             \textcolor{green}{\checkmark} & 
%             \textcolor{green}{\checkmark} &
%             \textcolor{green}{\checkmark} &
%             - \\
%         & MLP & - & - & - & - & - & - & \textcolor{green}{\checkmark} \\
% \hline
\multirow{5}{*}{\textbf{Tokenization}}
        & Fixed-Length Patches & - & - & - & \textcolor{green}{\checkmark} & \textcolor{green}{\checkmark} & \textcolor{green}{\checkmark} & - \\
        & Multi-Scale Patches & - & - & \textcolor{green}{\checkmark} & - & - & - & - \\
        & Adaptive-Length Patches & - & - & - & - & - & - & \textcolor{green}{\checkmark} \\
        & Binning & \textcolor{green}{\checkmark} & - & - & - & - & - & - \\
        & Lags & - & \textcolor{green}{\checkmark} & - & - & - & - & - \\
\hline
\multirow{2}{*}{\textbf{Projection Layer}}
        & Linear & 
        - & 
        \textcolor{green}{\checkmark} & 
        \textcolor{green}{\checkmark} & 
        \textcolor{green}{\checkmark} & 
        \textcolor{green}{\checkmark} &  
        - & 
        \textcolor{green}{\checkmark} \\
        & Residual & - & - & - & - & - & \textcolor{green}{\checkmark} & -  \\
\hline
\multirow{3}{*}{\textbf{Positional Encoding}}
        & Relative & - & - & - & \textcolor{green}{\checkmark} & - & - & - \\
        & SinCos & - & - & - & \textcolor{green}{\checkmark} & \textcolor{green}{\checkmark} & \textcolor{green}{\checkmark} & (optional) \\
        & RoPE & - & \textcolor{green}{\checkmark} & \textcolor{green}{\checkmark} & - & - & - & - \\
\hline
\multirow{5}{*}{\textbf{Loss}}
        & MAE & - & - & - & - & - & - & - \\
        & MSE & - & - & - & \textcolor{green}{\checkmark} & \textcolor{green}{\checkmark} & \textcolor{green}{\checkmark} & \textcolor{green}{\checkmark} \\
        & Huber & - & - & - & - & - & - & - \\
        & Distribution & - & \textcolor{green}{\checkmark} (StudentT) & \textcolor{green}{\checkmark} (Mixture) & - & - & - & - \\
        & Cross-Entropy & \textcolor{green}{\checkmark} & - & - & - & - & - & - \\
\hline
\multirow{4}{*}{\textbf{Temporal Scaler}}
        & RevIN & - & - & \textcolor{green}{\checkmark} & \textcolor{green}{\checkmark} & \textcolor{green}{\checkmark} & \textcolor{green}{\checkmark} & \textcolor{green}{\checkmark} \\
        & Robust & - & \textcolor{green}{\checkmark} & - & - & - & - & - \\
        & Mean Absolute & \textcolor{green}{\checkmark} & - & - & - & - & - & - \\
\bottomrule
\end{tabular}
}
\end{table*}

%% file: tables/experiment_table.tex
\begin{table*}[h]
    \caption{Hyper-parameters used in the TSFM experiments, with default hyper-parameters in \textbf{bold}.}
    \centering
    \label{tab:tsfm_design_decisions_experiments}
    \resizebox{0.7\textwidth}{!}{\begin{tabular}{r l} 
    \toprule
      \textbf{Experiment} & \textbf{Experiment Parameters} \\ \midrule
      \textbf{Model Size} : & T5-efficient-\{\textbf{tiny}, mini, small, base\}\\

      \textbf{Attention} : & \{\textbf{Bidirectional}, Casual\}\\

      \textbf{Projection/Head Layer} : & \{\textbf{Linear}, Residual\}\\

      \textbf{Tokenization} : & \{None, \textbf{Fixed-patch length}, Binning, Lags (l=2)\}\\

      \textbf{Fixed-Patch Length} : & \{8, 16, 32, 64, \textbf{96}, 128\}\\

      \textbf{Positional Encoding} : & \{Relative, SinCos, \textbf{Relative+SinCos}, RoPE\}\\

      \textbf{Loss Function} : & \{\textbf{MAE}, MSE, Huber, Distribution (StudentT, q=[80, 90])\}\\

      \textbf{Scaler} : & \{\textbf{RevIN} (standard, non-learnable), Robust\}\\

      \textbf{Context Length} : & \{\textbf{256}, 512\}\\

      \textbf{Decomposition} : & \{\textbf{None}, Moving Avg.\}\\
      \bottomrule
    \end{tabular}}
  \centering
\end{table*}

%% file: sections/appendixB.tex
\subsection{Analysis of Compositionality in the Latent Space}\label{apd:cka_embds}

We evaluate the compositionality of the model in the latent space by comparing the similarity of model embeddings using Centered Kernel Alignment (CKA). We hypothesize that the out-of-distribution (OOD) model trained on both frequency and trend components will exhibit latent representations that are more similar to those of the in-distribution (ID) model trained on compositions of signals (i.e., the sum of trend and frequency signals). Such similarity would indicate that the OOD model is effectively leveraging information from both components to forecast the target signal, demonstrating compositionality in the latent space. To test this hypothesis, we extract embeddings from three different OOD models: 1) models trained on only sinusoid signals, 2) models trained on only trend signals, and 3) models trained on both sinusoid and trend signals, by feeding all models the forecast input composed of summed trend and sinusoid signals. We similarly extract embeddings from the ID model. We use the best-performing model: the \Tfive{} model with a patch length of 128.

We use linear Centered Kernel Alignment (CKA) to quantify the similarity between neural network embeddings. Given two representation matrices \( X, Y \in \mathbb{R}^{n \times d} \), linear CKA is defined as:

\[
\text{CKA}(X, Y) = \frac{\| X_c^\top Y_c \|_F^2}{\| X_c^\top X_c \|_F \cdot \| Y_c^\top Y_c \|_F}
\]

where \( X_c \) and \( Y_c \) are mean-centered versions of \( X \) and \( Y \), and \( \| \cdot \|_F \) denotes the Frobenius norm. Linear CKA is invariant to isotropic scaling and orthogonal transformations, making it well-suited for comparing learned representations across different models.

We find that the latent space representations of the \Tfive model trained jointly on both trend and frequency components exhibit higher similarity—measured using Centered Kernel Alignment (CKA)—to those of the \Tfive model trained on the ground truth signal compositions. In contrast, \Tfive models trained exclusively on either trend or sinusoidal signals show lower alignment with the ground truth activations, as shown in Table~\ref{tab:cka_similarity}. This result is consistent with our hypothesis.

We further hypothesize that among the OOD models—originally trained on sinusoids, trend, or both components and then fine-tuned on ID data—the model trained on both signals will exhibit the highest similarity to its original (pre–fine-tuning) embeddings. This would suggest that only minimal adaptation is needed to match the performance of the ID model, further supporting the idea that compositional structure is already present in the latent space before fine-tuning. To test this, we fine-tune the \Tfive{} model on an ID training set composed of trend and sinusoid signal compositions. Each model is fine-tuned for up to 1,000 steps with early stopping. We then extract embeddings before and after fine-tuning and compute CKA similarity between the two stages. The resulting CKA similarity scores are 0.84, 0.33, and \textbf{0.85} for models originally trained on sinusoids, trend, and both components, respectively. This result is consistent with our hypothesis.

\subsection{Model Rank and Pairwise Comparisons}\label{apd:cd_diagrams}
\input{figures/cd_figures_apd}

\newpage
\subsection{Model Forecasts}\label{apd:forecast_examples}
We include example forecasts for each of the 6 datasets used in this study.

\begin{figure*}[t!]
    \centering

    % Row 1
    \begin{subfigure}[t]{0.4\textwidth}
        \centering
        \includegraphics[width=\textwidth, trim=0 0 310 0, clip]{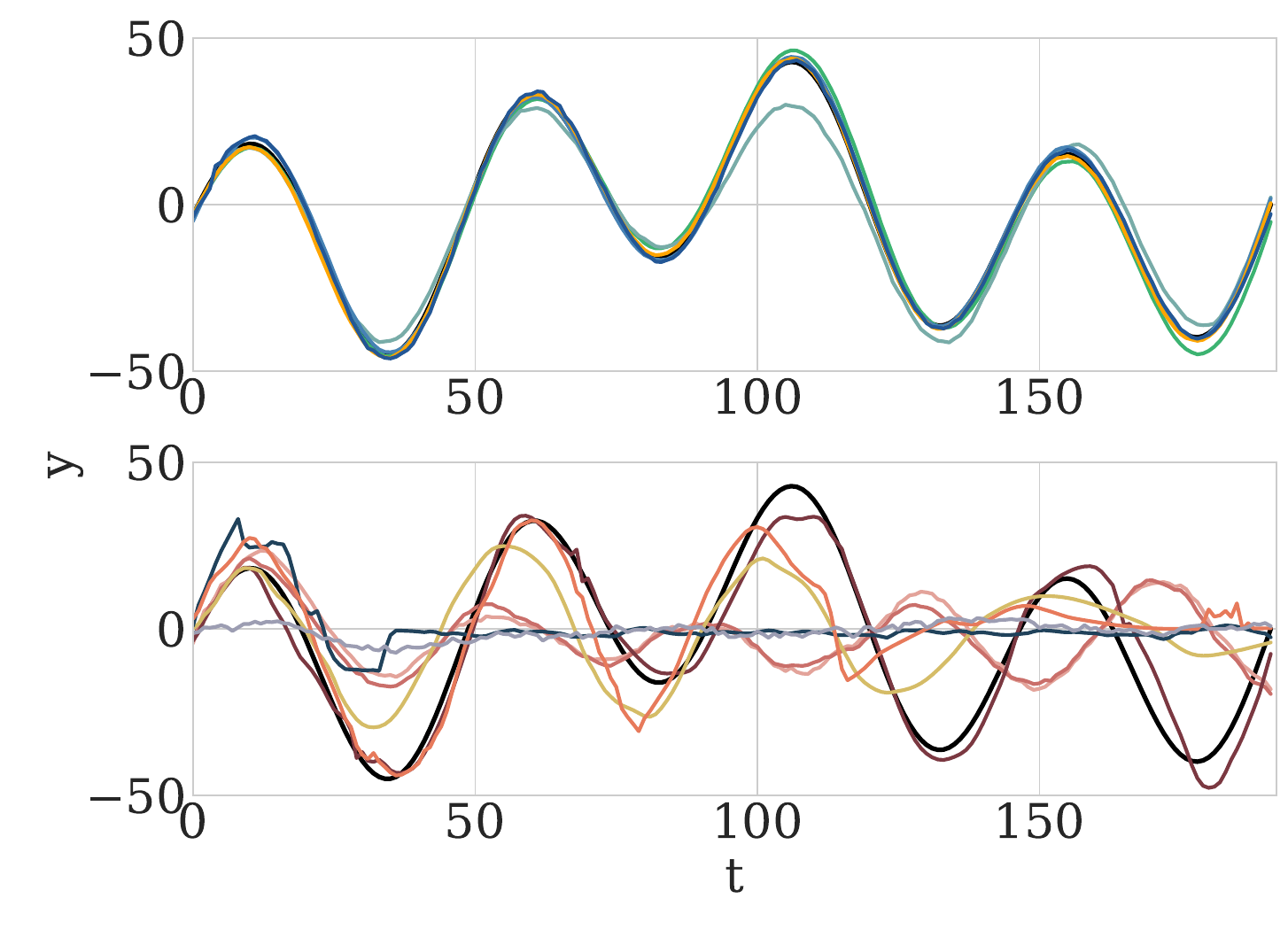}
        \caption{In-distribution Sinusoid}
        \label{fig:sine_id_forecast_example}
    \end{subfigure}
    \hfill
    \begin{subfigure}[t]{0.58\textwidth}
        \centering
        \includegraphics[width=\textwidth]{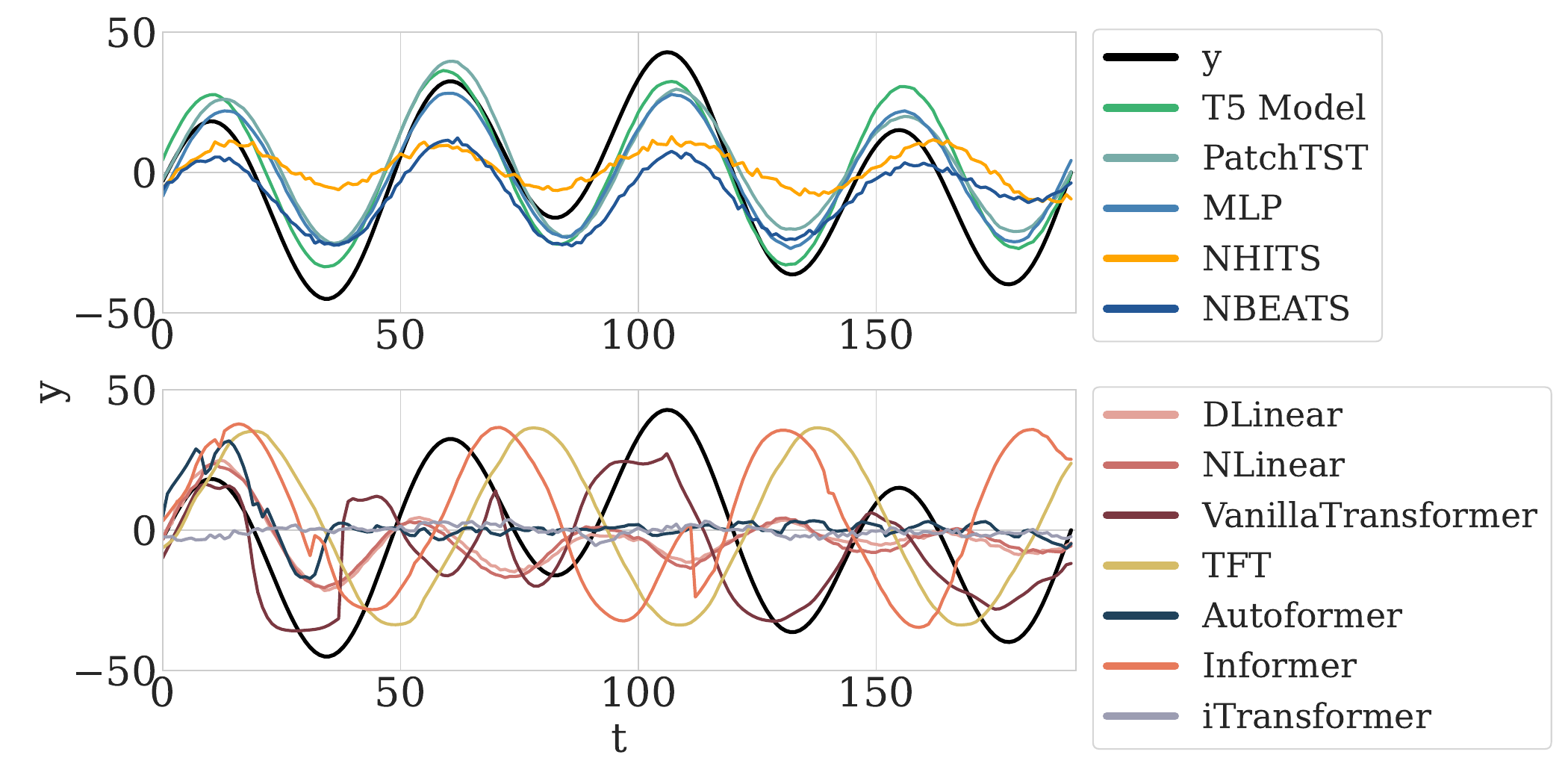}
        \caption{Out-of-distribution Sinusoid}
        \label{fig:sine_ood_forecast_example}
    \end{subfigure}

    \vspace{1.2em}

    % Row 2
    \begin{subfigure}[t]{0.4\textwidth}
        \centering
        \includegraphics[width=\textwidth, trim=0 0 310 0, clip]{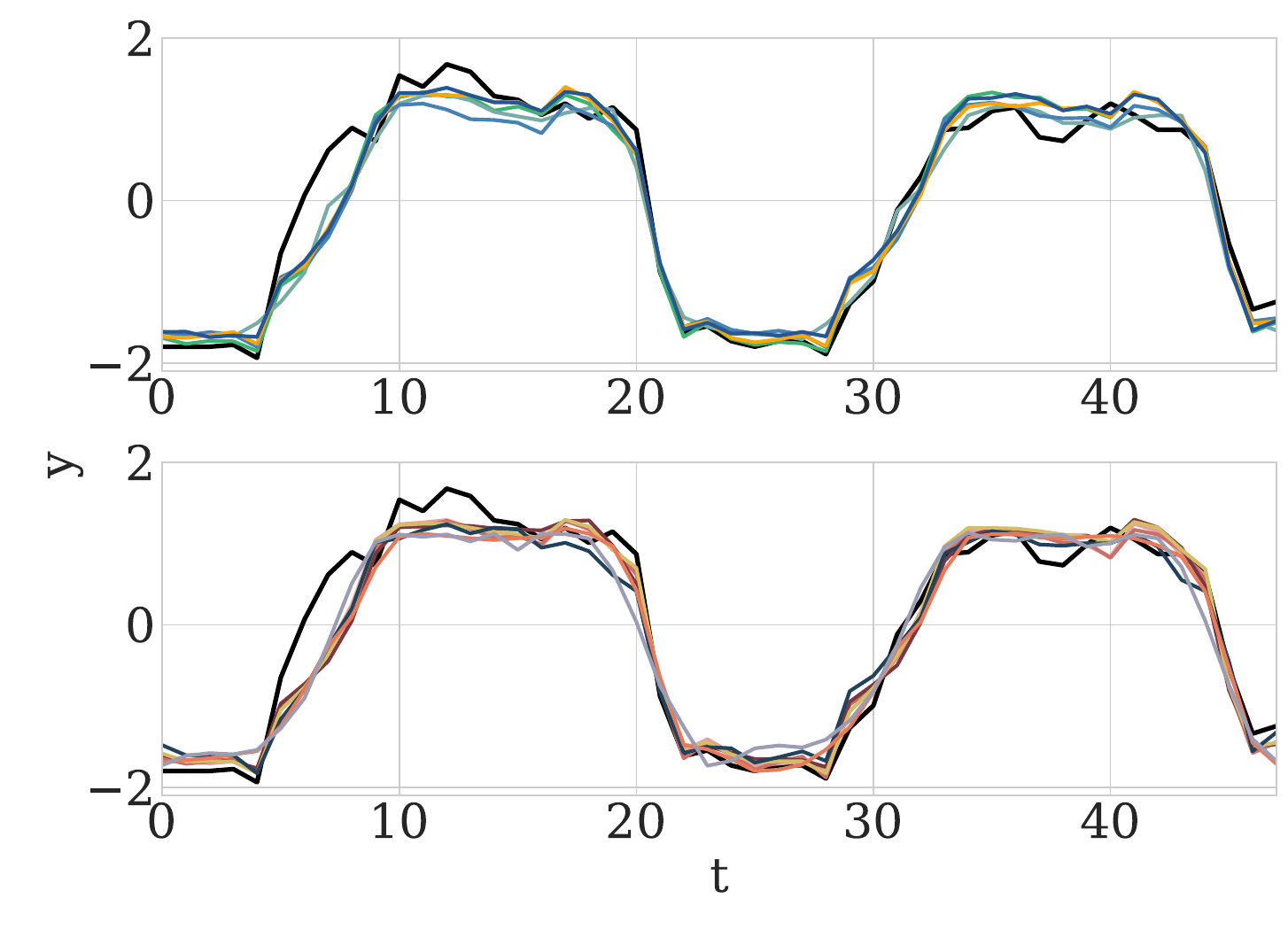}
        \caption{In-distribution ECL}
        \label{fig:ecl_id_forecast_example}
    \end{subfigure}
    \hfill
    \begin{subfigure}[t]{0.58\textwidth}
        \centering
        \includegraphics[width=\textwidth]{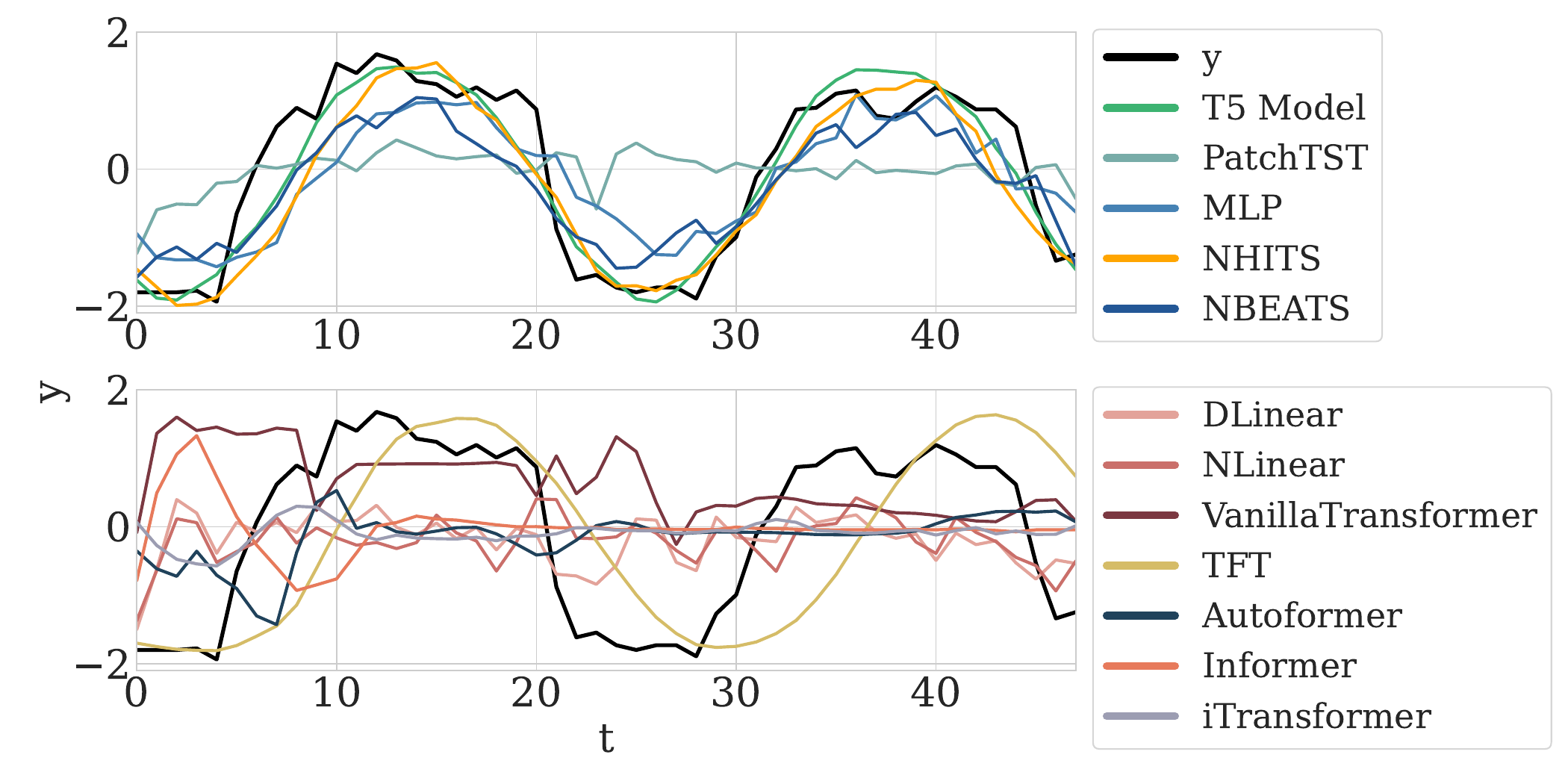}
        \caption{Out-of-distribution ECL}
        \label{fig:ecl_ood_forecast_example}
    \end{subfigure}

    \vspace{1.2em}

    % Row 3
    \begin{subfigure}[t]{0.40\textwidth}
        \centering
        \includegraphics[width=\textwidth, trim=0 0 310 0, clip]{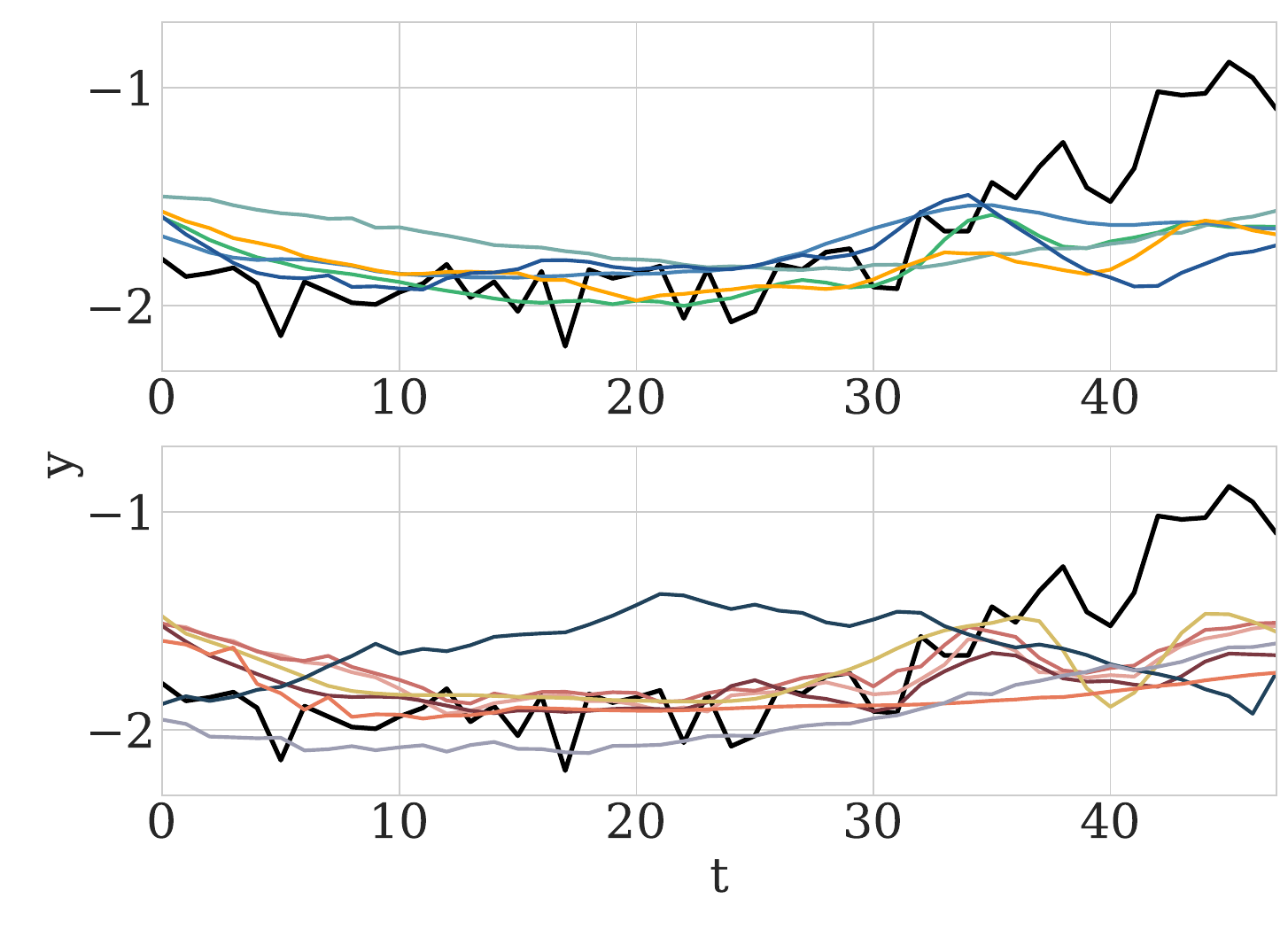}
        \caption{In-distribution ETTm2}
        \label{fig:ettm2_id_forecast_example}
    \end{subfigure}
    \hfill
    \begin{subfigure}[t]{0.58\textwidth}
        \centering
        \includegraphics[width=\textwidth]{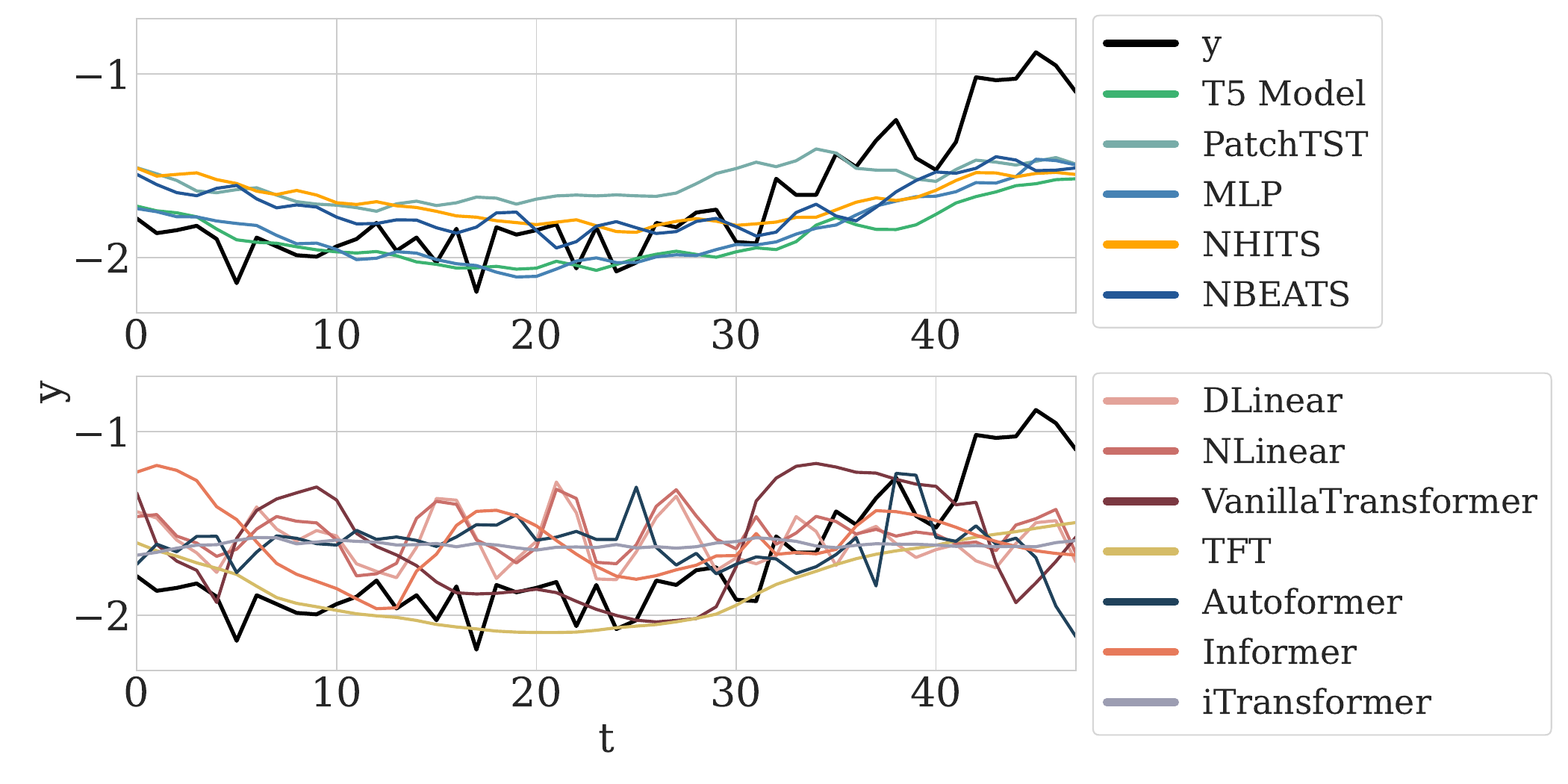}
        \caption{Out-of-distribution ETTm2}
        \label{fig:ettm2_ood_forecast_example}
    \end{subfigure}

    \caption{\textbf{(a, c, e)} In-distribution forecasts for Sinusoid, ECL, and ETTm2 datasets. \textbf{(b, d, f)} Corresponding out-of-distribution forecasts using the compositional reasoning paradigm. Top-performing models (Patch-based Transformers, MLPs) generalize well OOD, unlike other variants.}
    \label{fig:forecast_examples1_apd}
\end{figure*}

\begin{figure*}[t!]
    \centering

    % Row 1
    \begin{subfigure}[t]{0.4\textwidth}
        \centering
        \includegraphics[width=\textwidth, trim=0 0 310 0, clip]{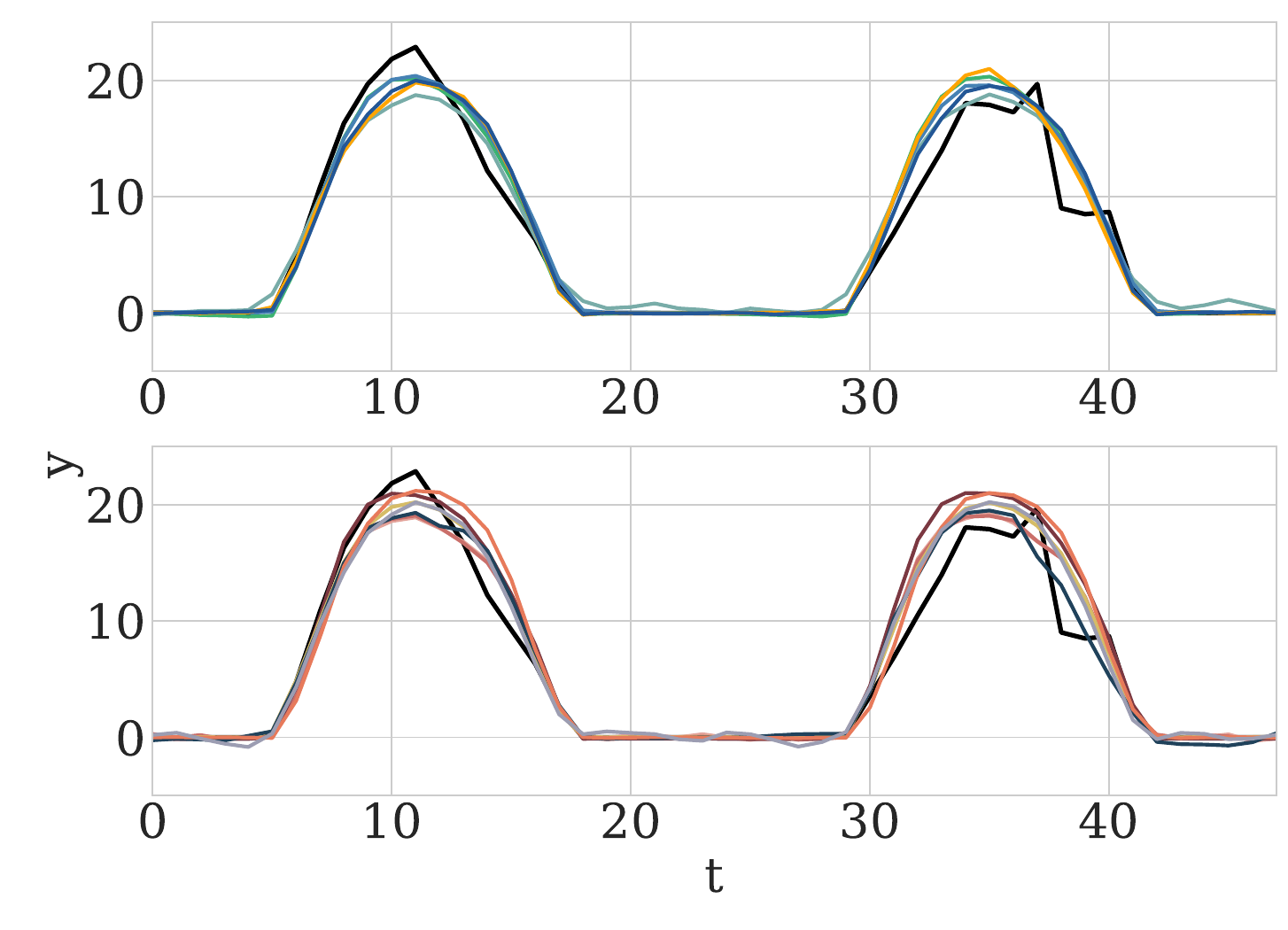}
        \caption{In-distribution Sinusoid}
        \label{fig:solar_id_forecast_example}
    \end{subfigure}
    \hfill
    \begin{subfigure}[t]{0.58\textwidth}
        \centering
        \includegraphics[width=\textwidth]{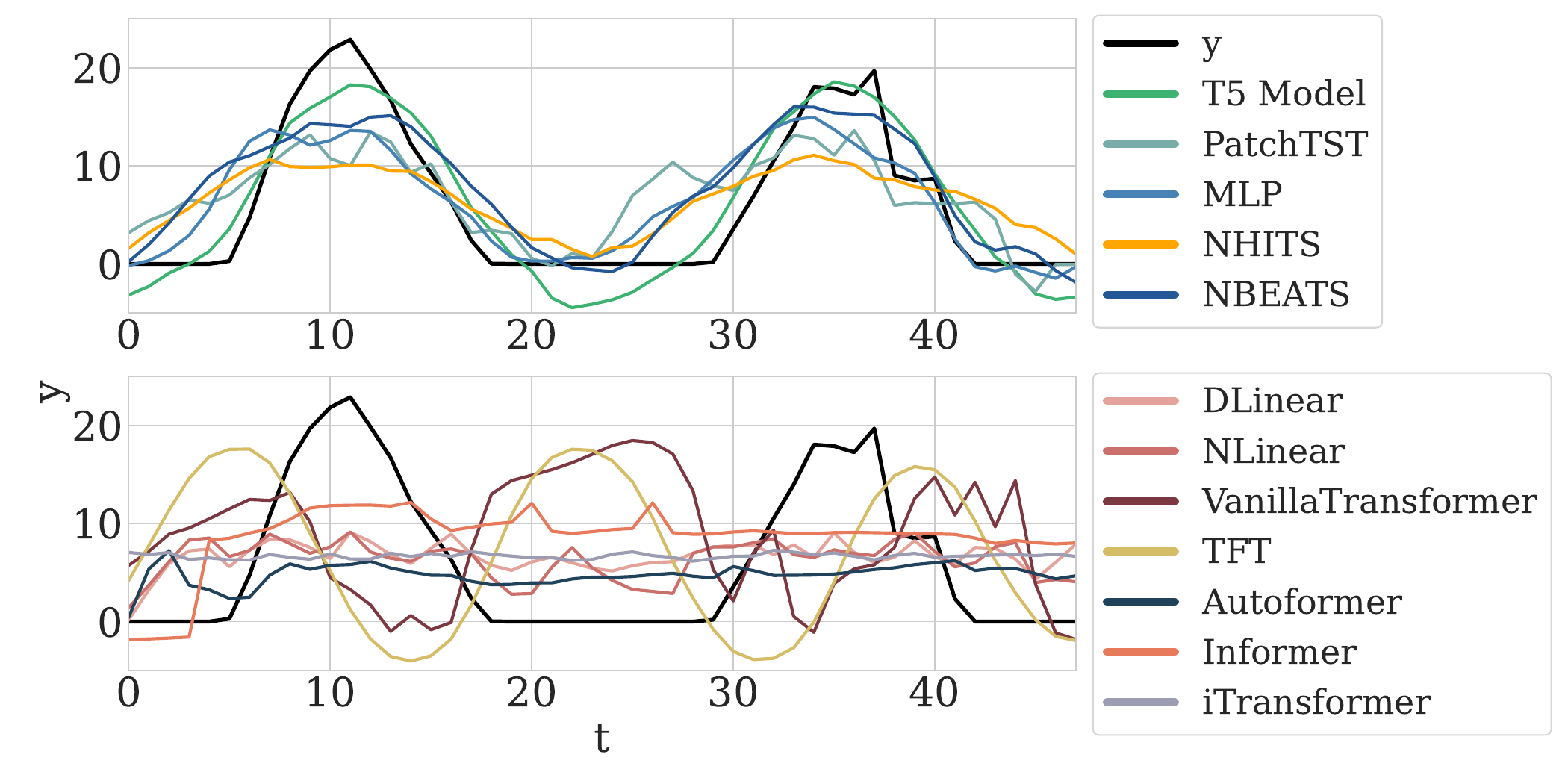}
        \caption{Out-of-distribution Sinusoid}
        \label{fig:solar_ood_forecast_example}
    \end{subfigure}

    \vspace{1.2em}

    % Row 2
    \begin{subfigure}[t]{0.4\textwidth}
        \centering
        \includegraphics[width=\textwidth, trim=0 0 310 0, clip]{images/subseasonal_id_forecast_example.pdf}
        \caption{In-distribution ECL}
        \label{fig:subseasonal_id_forecast_example}
    \end{subfigure}
    \hfill
    \begin{subfigure}[t]{0.58\textwidth}
        \centering
        \includegraphics[width=\textwidth]{images/subseasonal_ood_forecast_example.pdf}
        \caption{Out-of-distribution ECL}
        \label{fig:subseasonal_ood_forecast_example}
    \end{subfigure}

    \vspace{1.2em}

    % Row 3
    \begin{subfigure}[t]{0.40\textwidth}
        \centering
        \includegraphics[width=\textwidth, trim=0 0 310 0, clip]{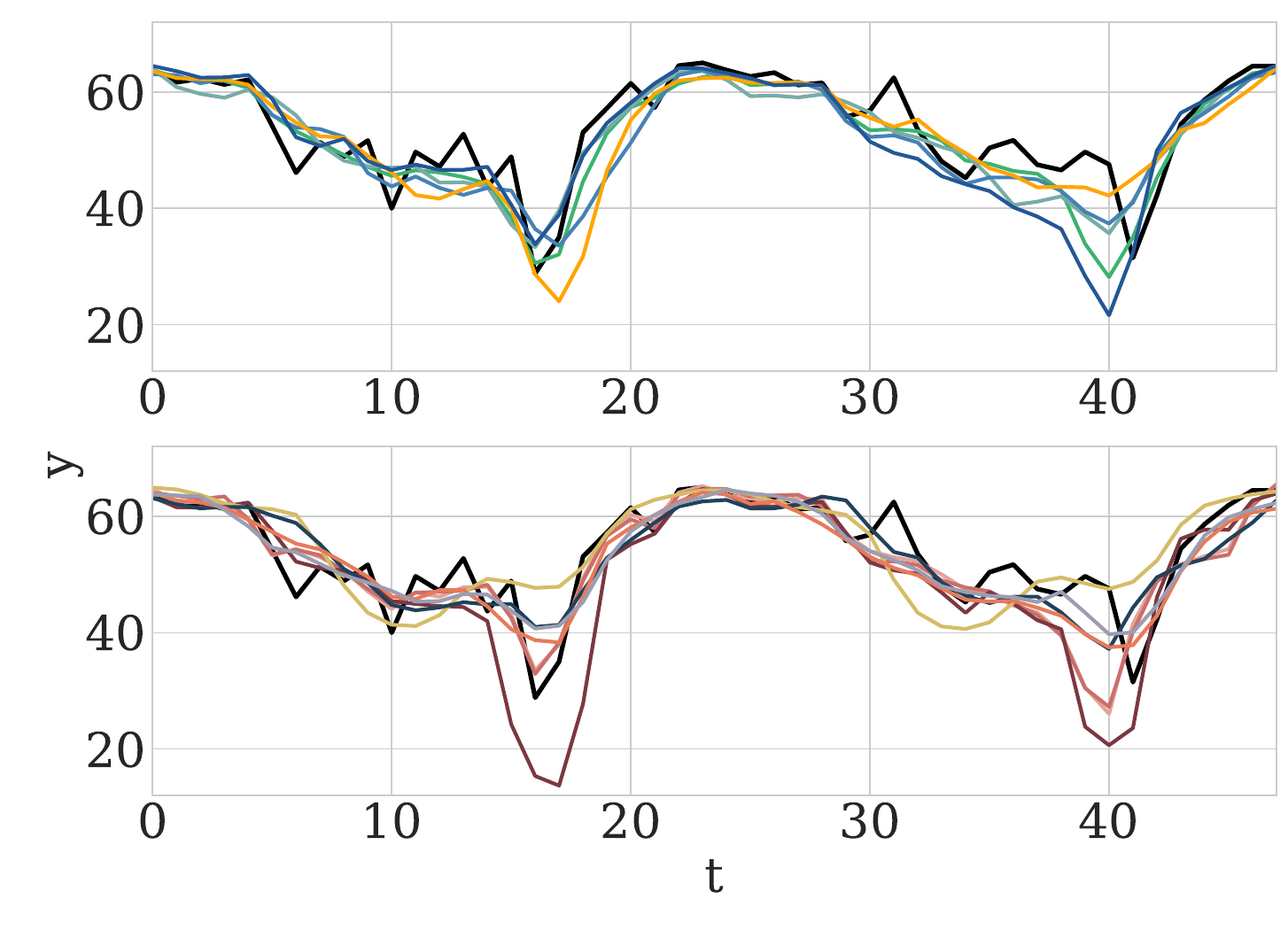}
        \caption{In-distribution ETTm2}
        \label{fig:loopseattle_id_forecast_example}
    \end{subfigure}
    \hfill
    \begin{subfigure}[t]{0.58\textwidth}
        \centering
        \includegraphics[width=\textwidth]{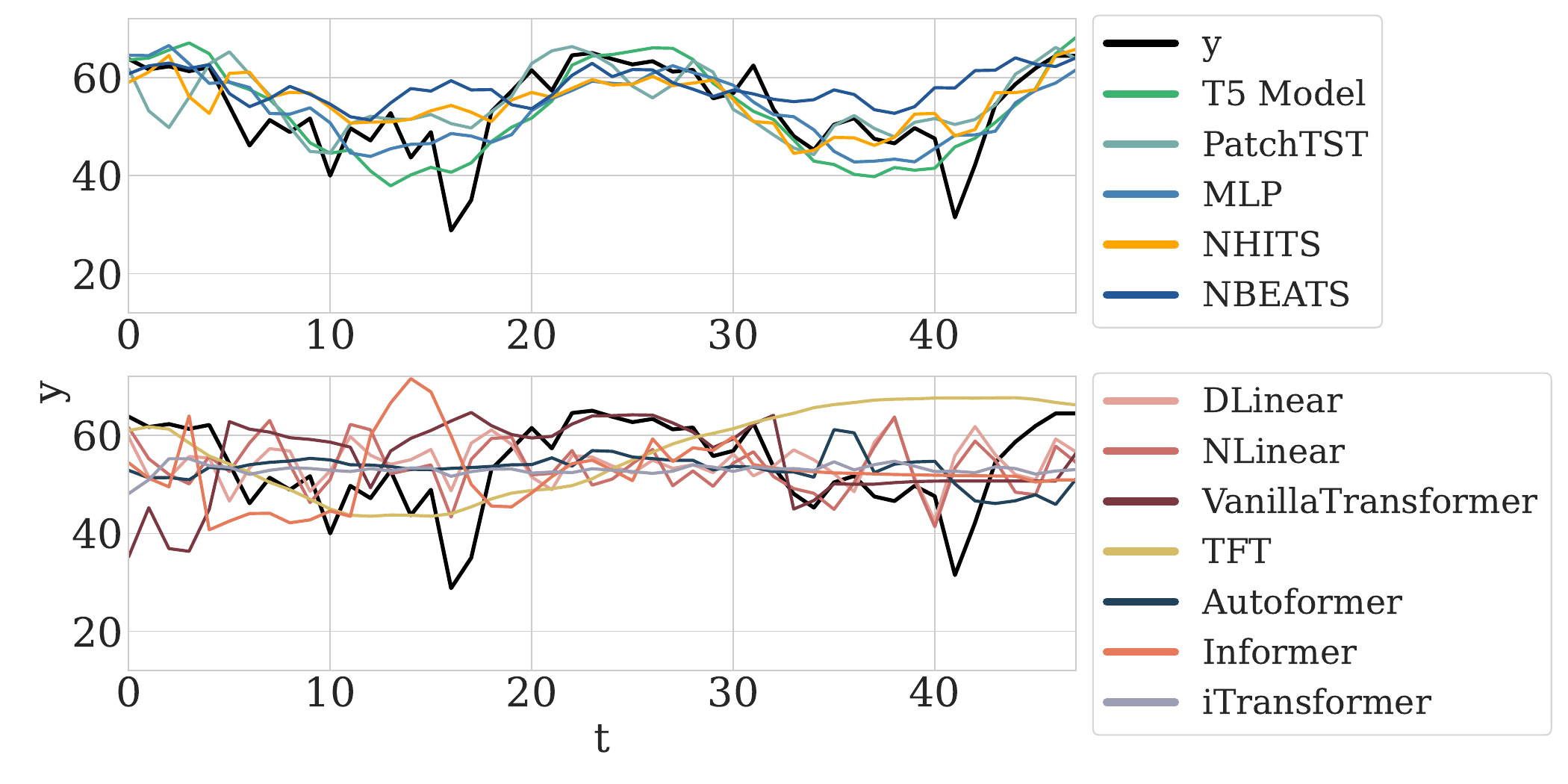}
        \caption{Out-of-distribution ETTm2}
        \label{fig:loopseattle_ood_forecast_example}
    \end{subfigure}

    \caption{\textbf{(a, c, e)} In-distribution forecasts for Solar, Subseasonal, and Loop Seattle datasets. \textbf{(b, d, f)} Corresponding out-of-distribution forecasts using the compositional reasoning paradigm. Top-performing models (Patch-based Transformers, MLPs) generalize well OOD, unlike other variants.}
    \label{fig:forecast_examples2_apd}
\end{figure*}

\newpage
\subsection{Model Performance, Efficiency, and Size Comparison}\label{apd:flops_plots}
We rank model performance across datasets and compare rank with model computational complexity in terms of floating-point operations per second (FLOPs) and model size in terms of the total number of trainable parameters. 

\begin{figure*}[t!]
    \centering

    \begin{minipage}{0.495\textwidth}
        \centering
        \includegraphics[width=\textwidth]{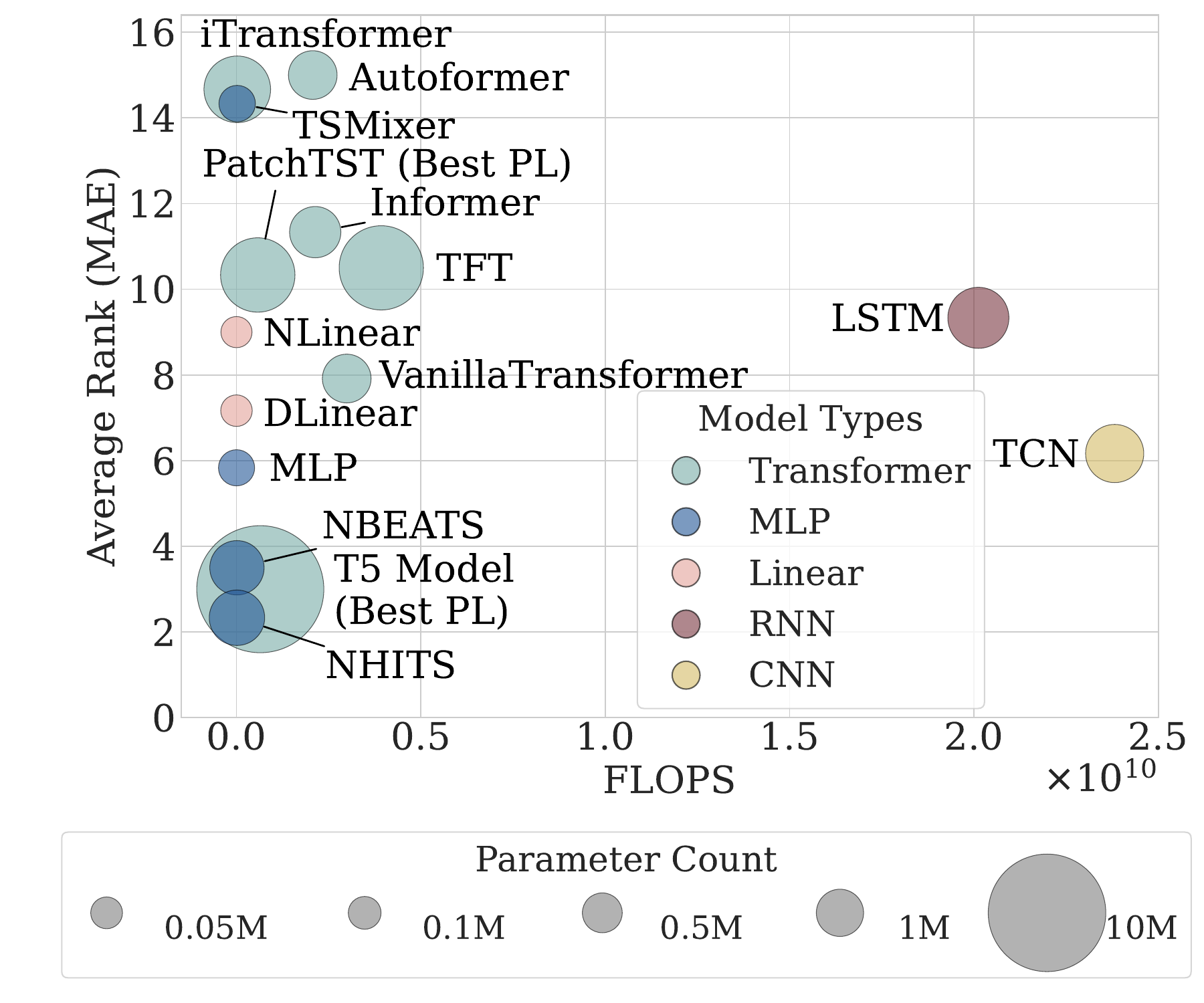}
        \subcaption{In-distribution series results (excluding \TimesNet)}
        \label{fig:flops_aggregate}
    \end{minipage}%
    \hfill
    \begin{minipage}{0.495\textwidth}
        \centering
        \includegraphics[width=\textwidth]{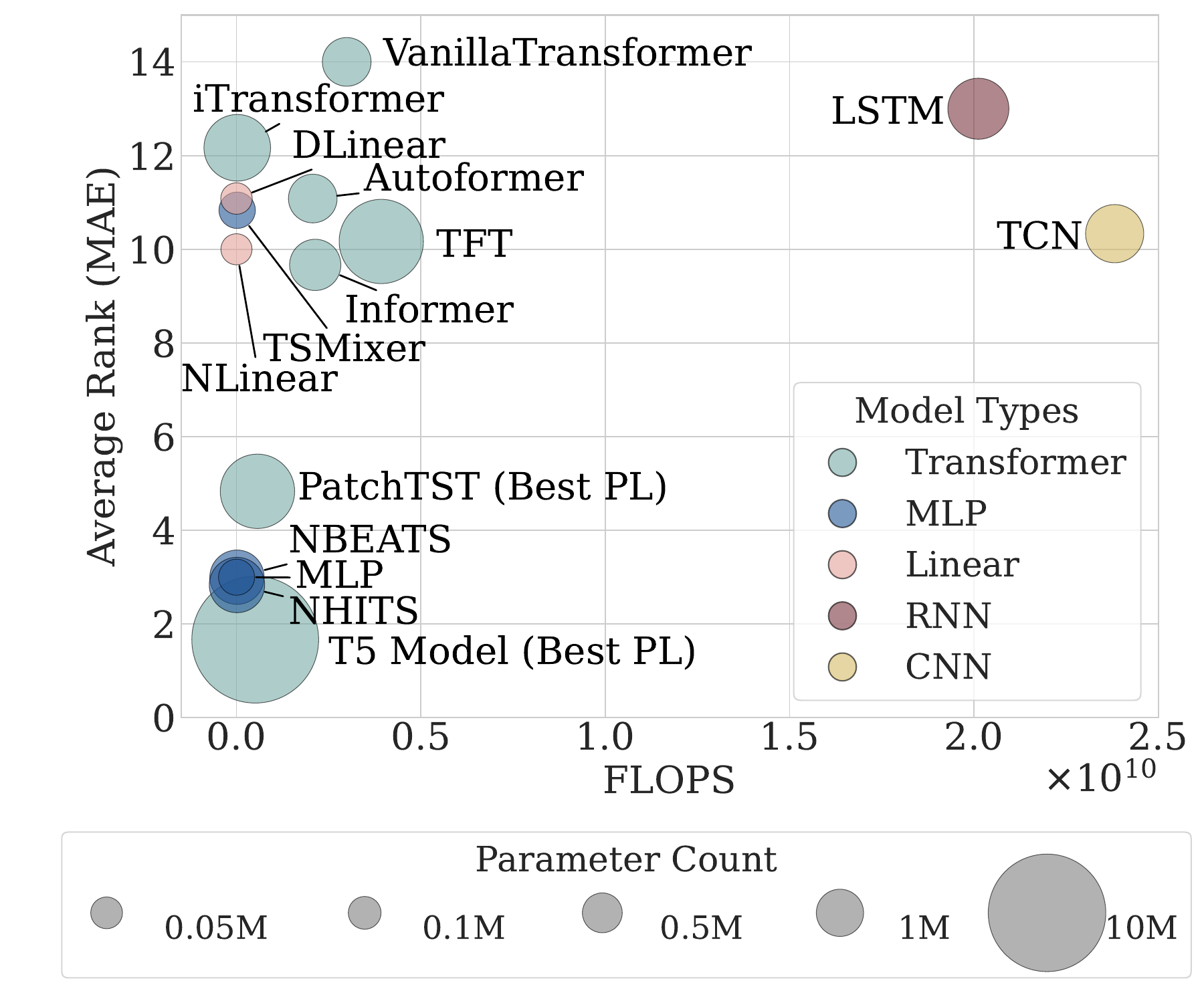}
        \subcaption{Out-of-distribution series results (excluding \TimesNet)}
        \label{fig:flops_component}
    \end{minipage}

    \vspace{1ex} % Vertical space between rows

    \begin{minipage}{0.495\textwidth}
        \centering
        \includegraphics[width=\textwidth]{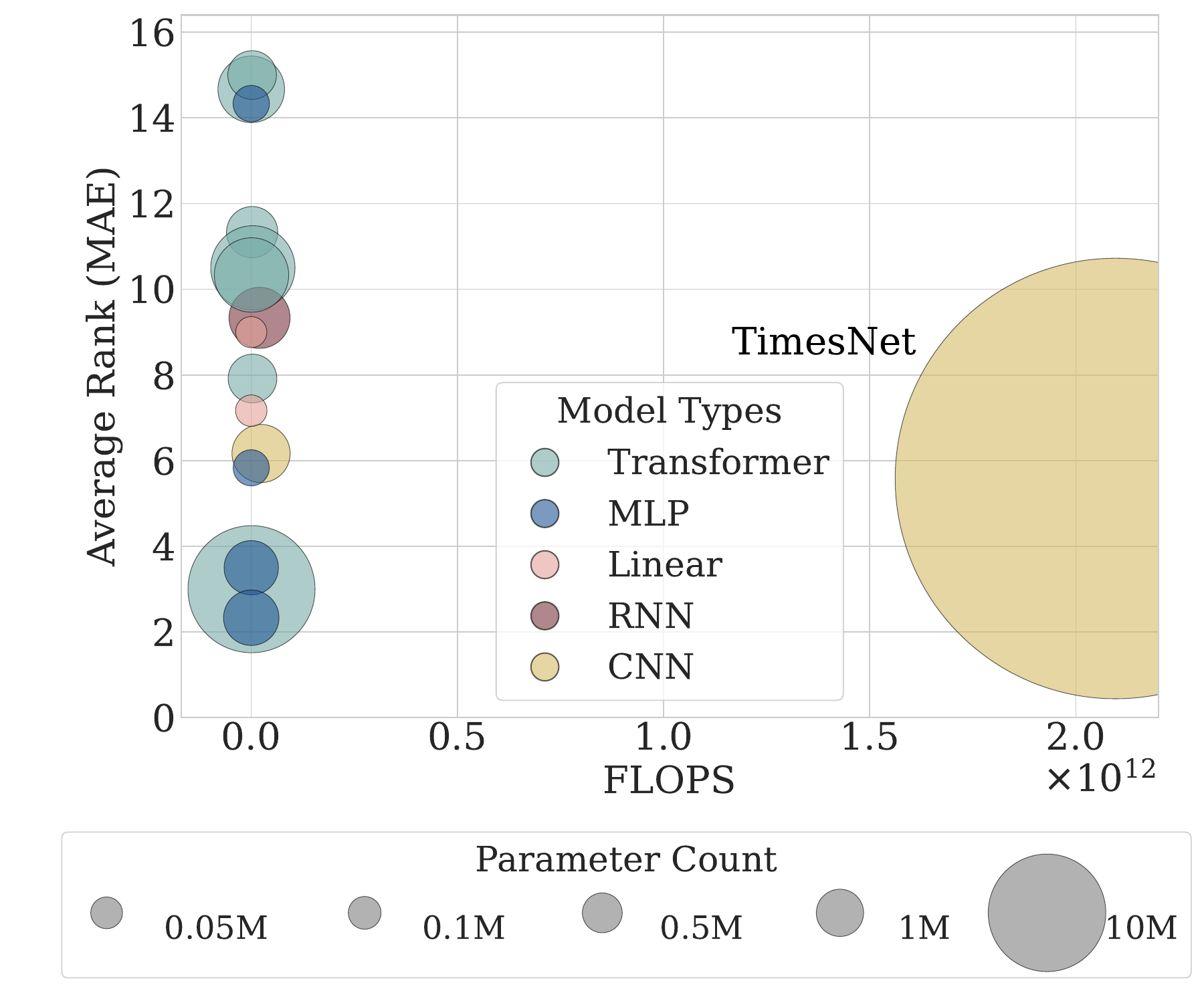}
        \subcaption{In-distribution series results (including \TimesNet)}
        \label{fig:flops_aggregate_with_timesnet}
    \end{minipage}%
    \hfill
    \begin{minipage}{0.495\textwidth}
        \centering
        \includegraphics[width=\textwidth]{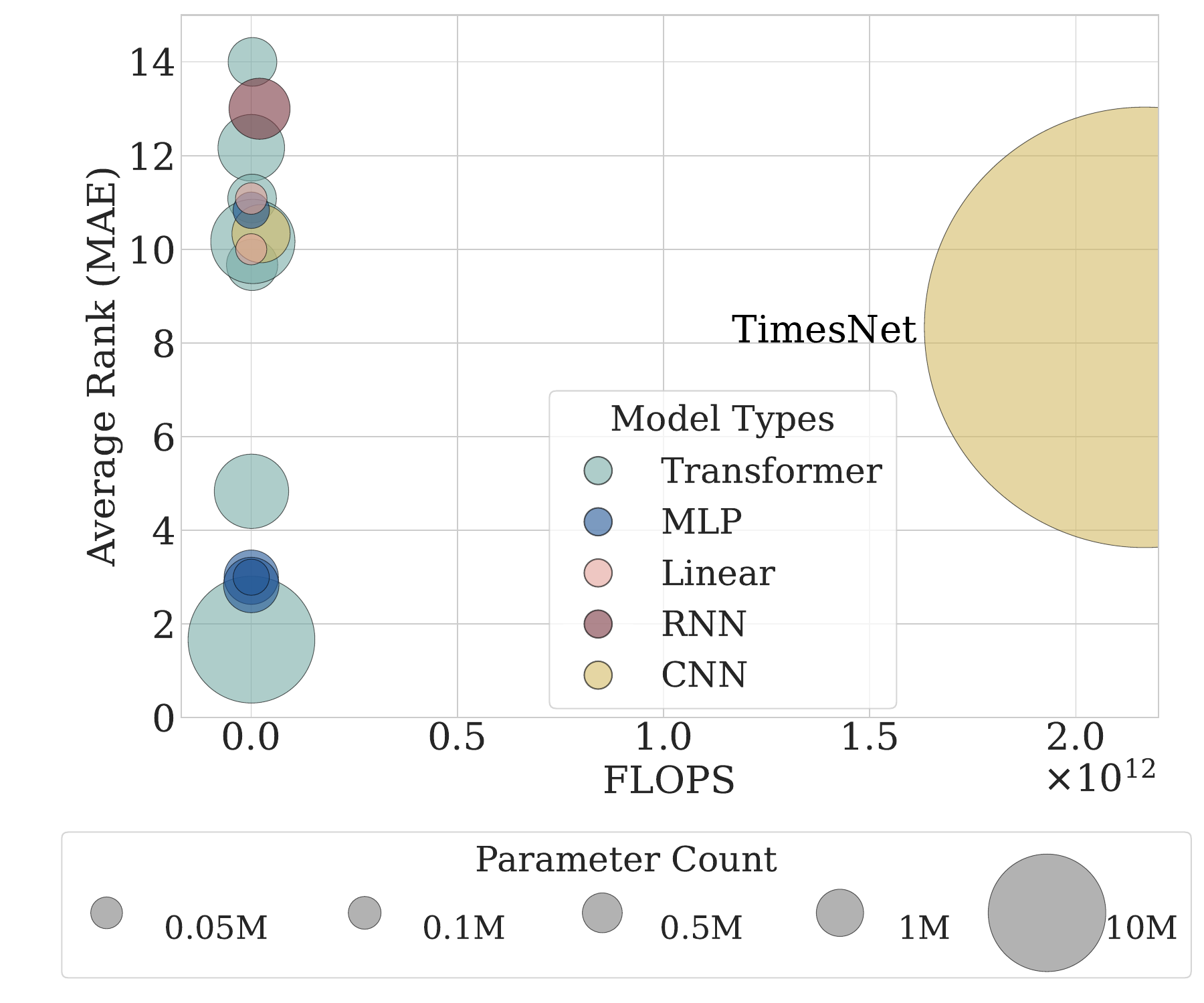}
        \subcaption{Out-of-distribution series results (including \TimesNet)}
        \label{fig:flops_component_with_timesnet}
    \end{minipage}

    \caption{Comparison of average rank across datasets and random seeds versus model computational complexity, measured by floating-point operations per second (FLOPs). The size of each point represents the number of trainable parameters, highlighting the trade-offs between model complexity and performance. \textbf{(a)} In-distribution and \textbf{(b)} out-of-distribution results for all models, excluding \TimesNet, are shown to provide a clearer comparison by mitigating the parameter size skew. \textbf{(c)} In-distribution and \textbf{(d)} out-of-distribution results for all models, including \TimesNet.}
    \label{fig:flops_model_comparison_apd}
\end{figure*}

\newpage

\input{tables/nonstationary_table_apd}

\newpage
\subsection{Model Composition Reasoning Results}\label{apd:composition_full_table_results}
We include the complete table results with MAE error mean and standard deviation measured across three random seeds. The results of the compositional reasoning task for 16 widely adopted time series forecasting models are included in Table~\ref{tab:composition_baseline_results_table}. Composition reasoning task results for controlled ablations of architecture components used in TSFMs are shown in Table~\ref{tab:composition_t5_results_table}.

\input{tables/composition_full_table_results}

\input{tables/composition_full_table_t5_results}

%% file: figures/cd_figures_apd.tex
\begin{figure*}[htbp]
    \centering
   \begin{minipage}{0.48\textwidth}
        \centering
        \includegraphics[width=\textwidth]{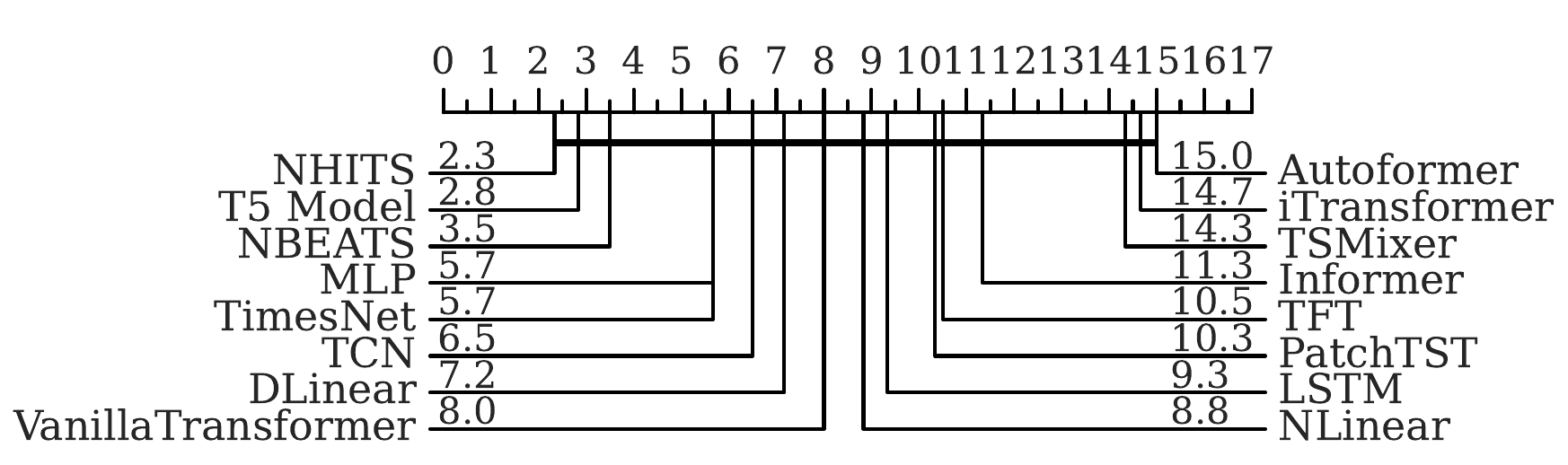}
        \subcaption{Model choice for in-distribution series (p-value: 2.71e-8)}
        \label{fig:cd_baselines_aggregate}
    \end{minipage}%
    \hfill
    \begin{minipage}{0.48\textwidth}
        \centering
        \includegraphics[width=\textwidth]{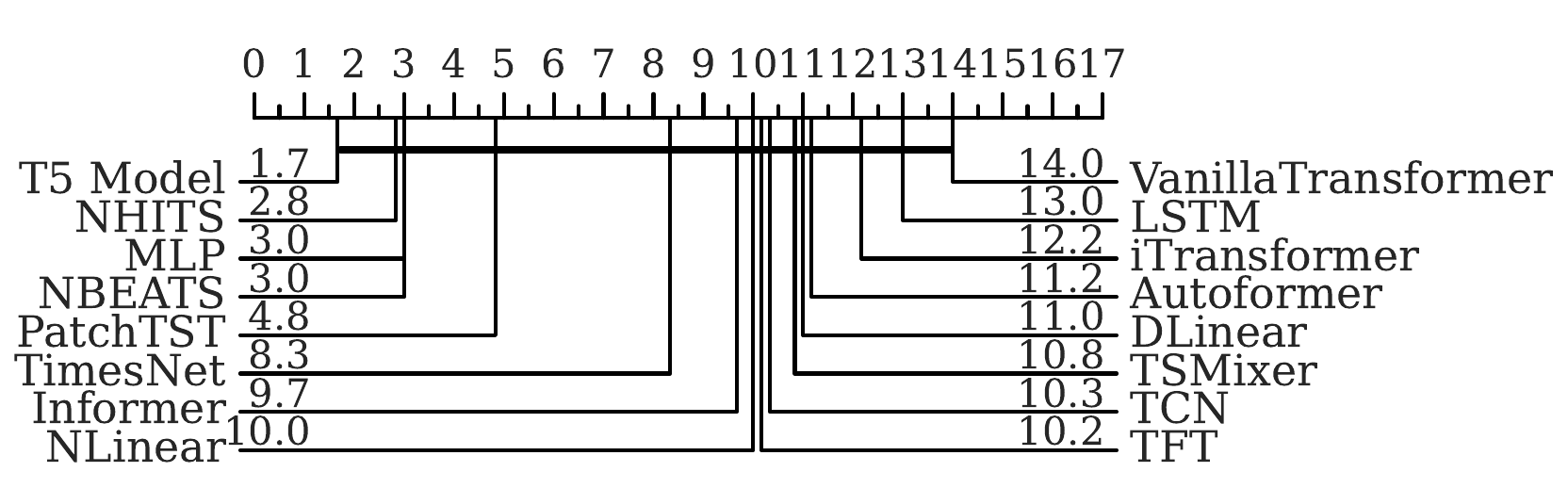}
        \subcaption{Model choice for out-of-distribution series (p-value: 3.52e-8)}
        \label{fig:cd_baselines_component}
    \end{minipage}

\caption{Critical Difference (CD) diagrams illustrate model ranks and pairwise statistical comparisons of model performance on compositional reasoning tasks across all datasets. Lower ranks indicate better performance. A thick horizontal line groups models that are not significantly different. The statistical tests used to generate the CD diagrams are detailed in Section \ref{section:evaluation}. \textbf{(a, b)} The patch-based Transformer models and MLP-based models outperform other models in both traditional and compositional reasoning forecasting paradigms. The Friedman p-value is included in the subcaptions.}
    \label{fig:cd_diagrams_baselines_apd}
\end{figure*}

\begin{figure*}[htbp]
    \centering

    \vspace{1ex} % Vertical space between rows

    \begin{minipage}{0.48\textwidth}
        \centering
        \includegraphics[width=\textwidth]{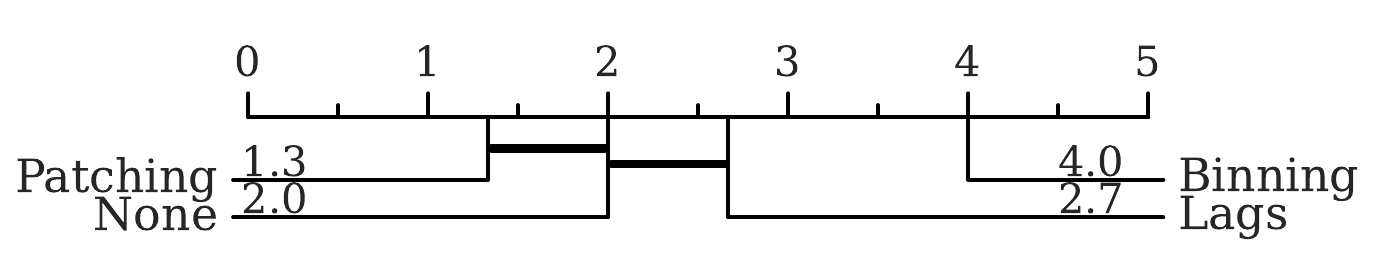}
        \subcaption{Tokenization for in-distribution series (p-value=2.91e-3)}
        \label{fig:cd_tokenization_ablation_aggregate}
    \end{minipage}%
    \hfill
    \begin{minipage}{0.48\textwidth}
        \centering
        \includegraphics[width=\textwidth]{images/cd_tokenization_ablation_component.pdf}
        \subcaption{Tokenization for out-of-distribution series (p-value=1.17e-2)}
        \label{fig:cd_tokenization_ablation_component}
    \end{minipage}

    \vspace{1ex} % Vertical space between rows

    \begin{minipage}{0.48\textwidth}
        \centering
        \includegraphics[width=\textwidth]{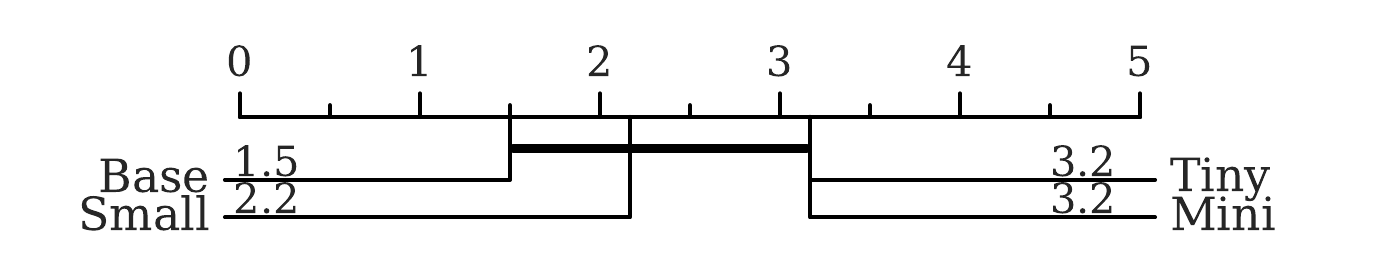}
        \subcaption{Model size for in-distribution series (p-value: 6.58e-2)}
        \label{fig:cd_size_ablation_aggregate}
    \end{minipage}%
    \hfill
    \begin{minipage}{0.48\textwidth}
        \centering
        \includegraphics[width=\textwidth]{images/cd_size_ablation_component.pdf}
        \subcaption{Model Size for out-of-distribution series (p-value=8.58e-2)}
        \label{fig:cd_size_ablation_component}
    \end{minipage}

    \vspace{1ex} % Vertical space between rows

    \begin{minipage}{0.48\textwidth}
        \centering
        \includegraphics[width=\textwidth]{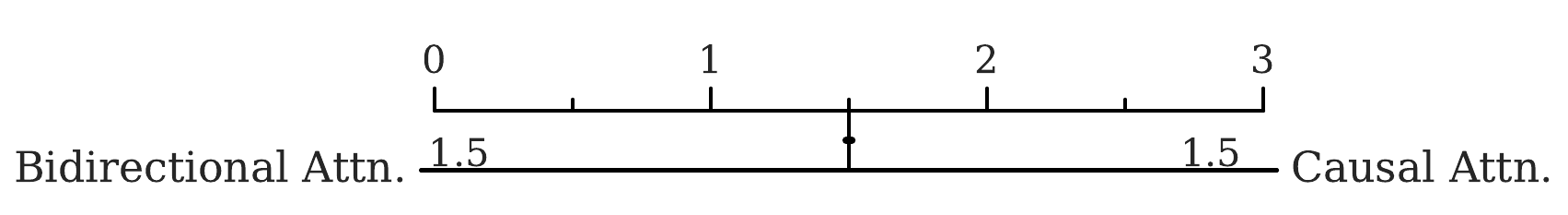}
        \subcaption{Attn. type for in-distribution series}
        \label{fig:cd_attn_ablation_aggregate}
    \end{minipage}%
    \hfill
    \begin{minipage}{0.48\textwidth}
        \centering
        \includegraphics[width=\textwidth]{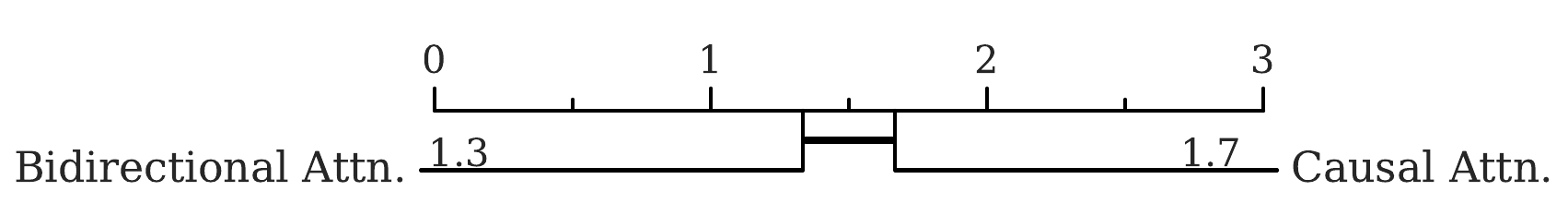}
        \subcaption{Attn. type for out-of-distribution series}
        \label{fig:cd_attn_ablation_component}
    \end{minipage}

    \vspace{1ex} % Vertical space between rows

    \begin{minipage}{0.48\textwidth}
        \centering
        \includegraphics[width=\textwidth]{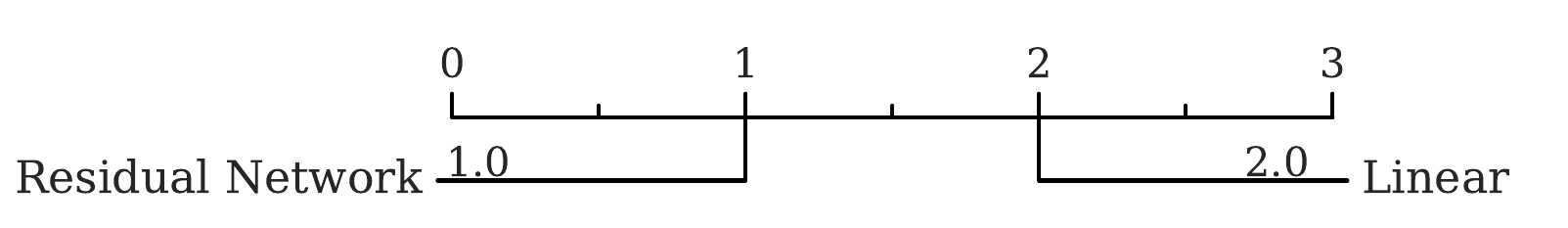}
        \subcaption{Projection layer for in-distribution series}
        \label{fig:cd_proj_ablation_aggregate}
    \end{minipage}%
    \hfill
    \begin{minipage}{0.48\textwidth}
        \centering
        \includegraphics[width=\textwidth]{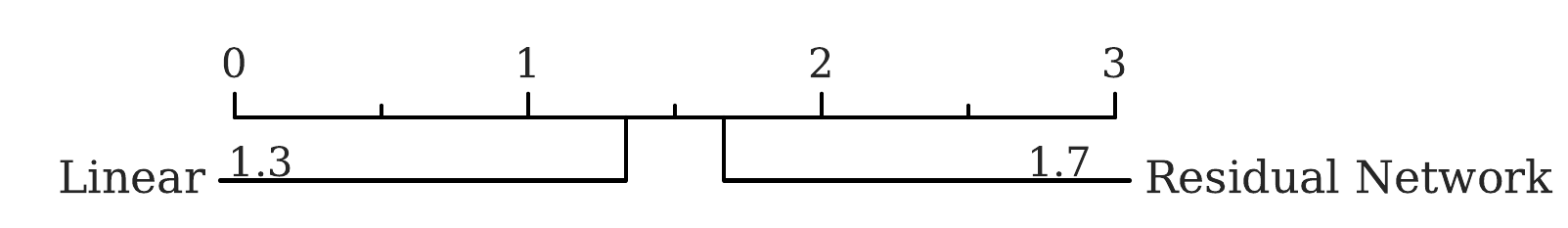}
        \subcaption{Projection layer for out-of-distribution series}
        \label{fig:cd_proj_ablation_component}
    \end{minipage}

    \vspace{1ex} % Vertical space between rows

    \begin{minipage}{0.48\textwidth}
        \centering
        \includegraphics[width=\textwidth]{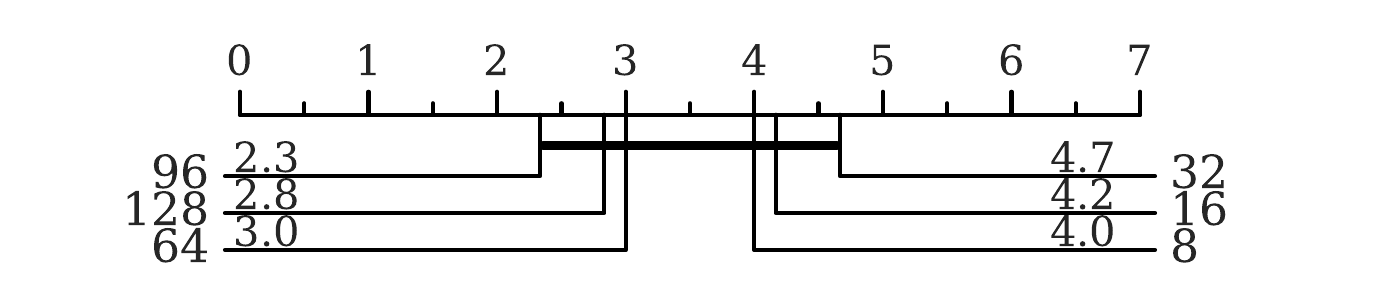}
        \subcaption{Token length for in-distribution series (p-value=0.22)}
        \label{fig:cd_tokenlen_ablation_aggregate}
    \end{minipage}%
    \hfill
    \begin{minipage}{0.48\textwidth}
        \centering
        \includegraphics[width=\textwidth]{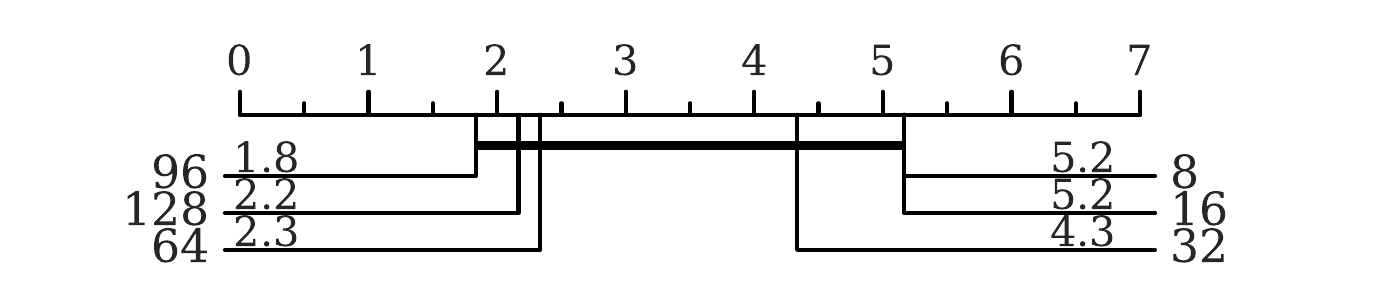}
        \subcaption{Token length for out-of-distribution series (p-value=8.62e-4)}
        \label{fig:cd_tokenlen_ablation_component}
    \end{minipage}

    \vspace{1ex} % Vertical space between rows

    \begin{minipage}{0.48\textwidth}
        \centering
        \includegraphics[width=\textwidth]{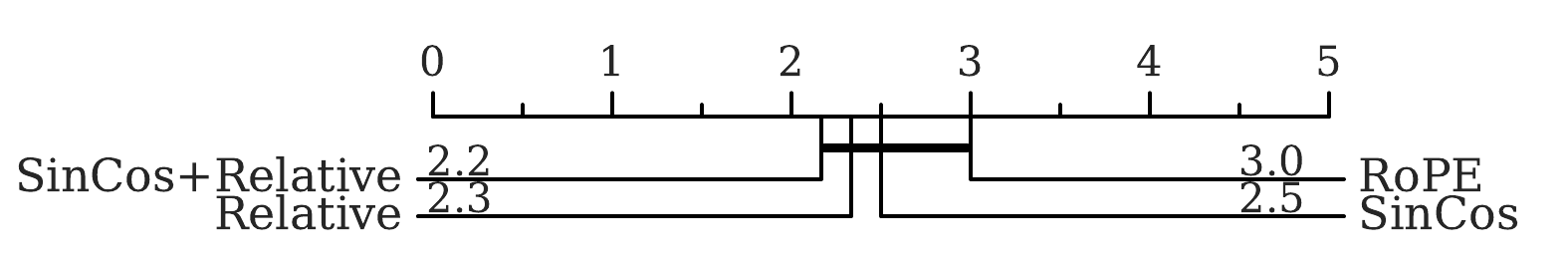}
        \subcaption{Positional encoding for in-distribution series (p-value=0.71)}
        \label{fig:cd_pe_ablation_aggregate}
    \end{minipage}%
    \hfill
    \begin{minipage}{0.48\textwidth}
        \centering
        \includegraphics[width=\textwidth]{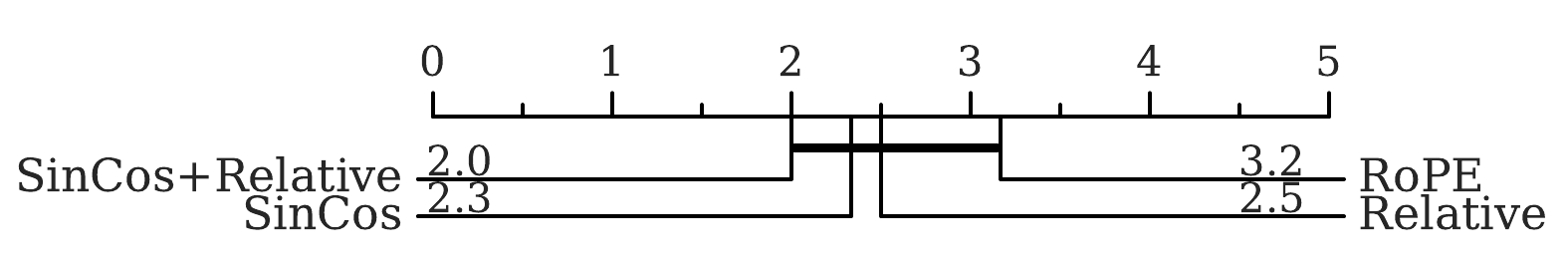}
        \subcaption{Positional encoding for out-of-distribution series (p-value=0.46)}
        \label{fig:cd_pe_ablation_component}
    \end{minipage}
    
    \vspace{1ex} % Vertical space between rows

    \begin{minipage}{0.48\textwidth}
        \centering
        \includegraphics[width=\textwidth]{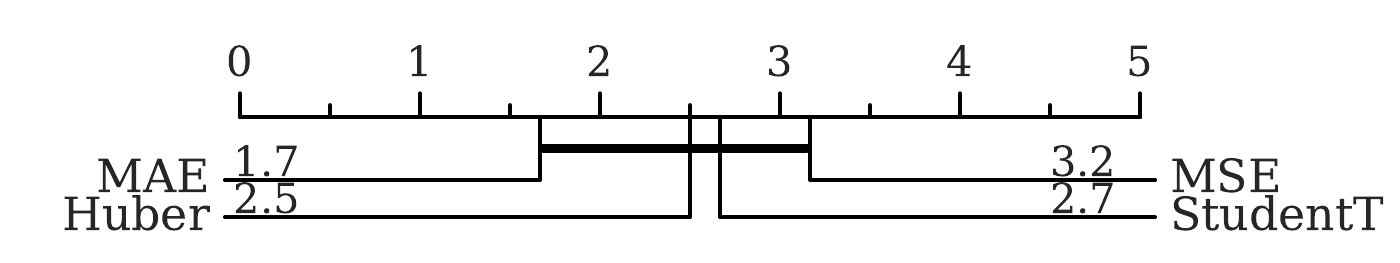}
        \subcaption{Loss function for in-distribution series (p-value=0.24)}
        \label{fig:cd_loss_ablation_aggregate}
    \end{minipage}%
    \hfill
    \begin{minipage}{0.48\textwidth}
        \centering
        \includegraphics[width=\textwidth]{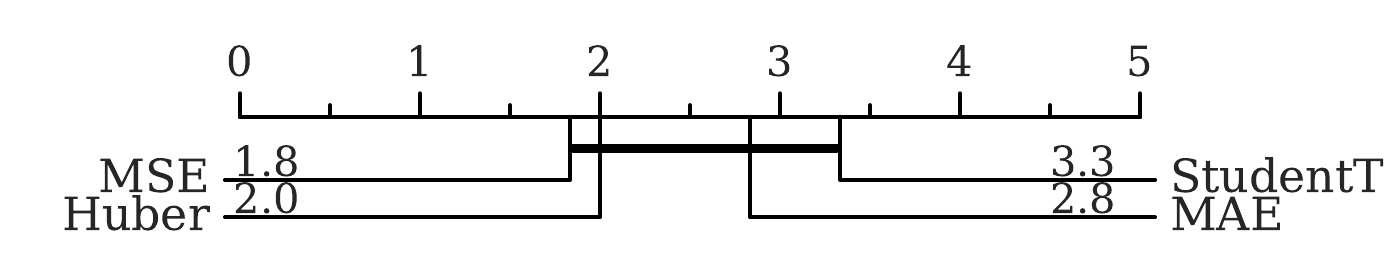}
        \subcaption{Loss function for out-of-distribution series (p-value=0.14)}
        \label{fig:cd_loss_ablation_component}
    \end{minipage}

    \vspace{1ex} % Vertical space between rows

    \begin{minipage}{0.48\textwidth}
        \centering
        \includegraphics[width=\textwidth]{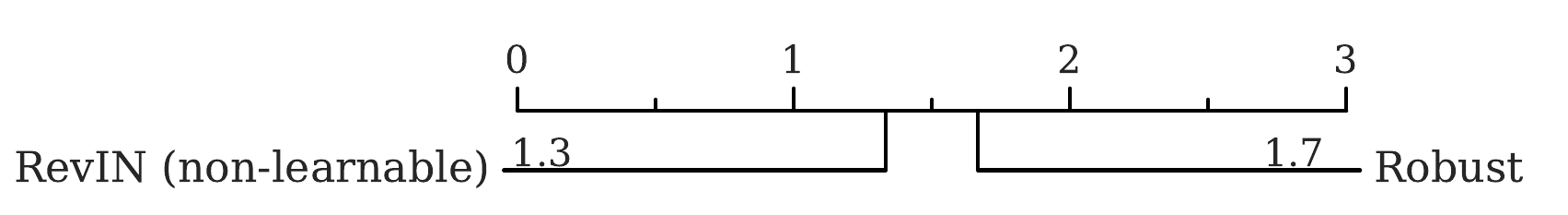}
        \subcaption{Scaler for in-distribution series}
        \label{fig:cd_scaler_ablation_aggregate}
    \end{minipage}%
    \hfill
    \begin{minipage}{0.48\textwidth}
        \centering
        \includegraphics[width=\textwidth]{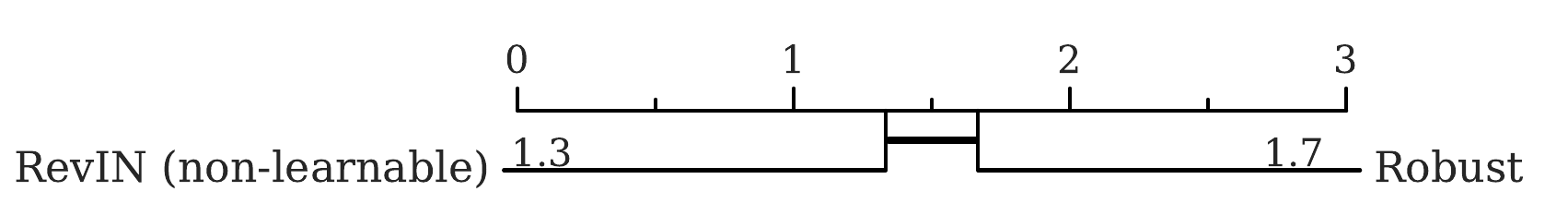}
        \subcaption{Scaler function for out-of-distribution series}
        \label{fig:cd_scaler_ablation_component}
    \end{minipage}

    \vspace{1ex} % Vertical space between rows

    \begin{minipage}{0.48\textwidth}
        \centering
        \includegraphics[width=\textwidth]{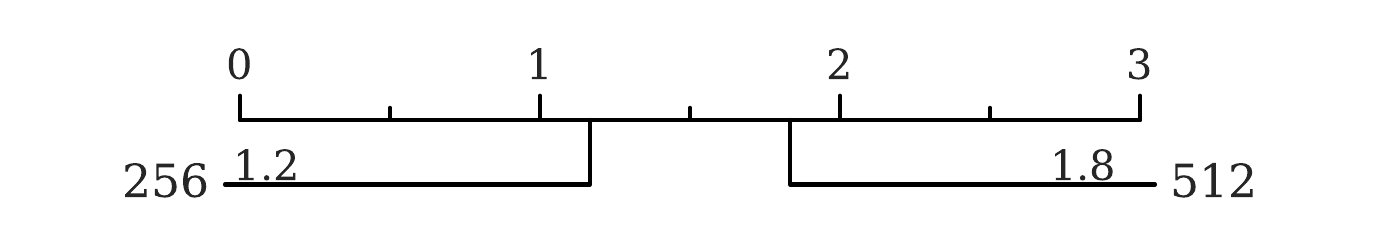}
        \subcaption{Context length function for in-distribution series}
        \label{fig:cd_contextlen_ablation_aggregate}
    \end{minipage}%
    \hfill
    \begin{minipage}{0.48\textwidth}
        \centering
        \includegraphics[width=\textwidth]{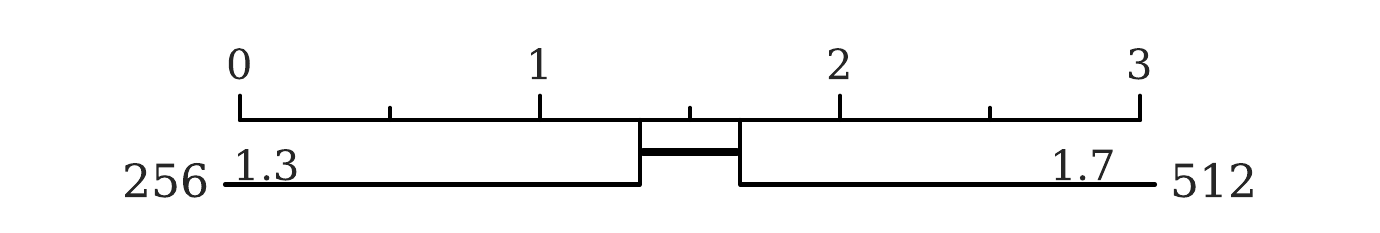}
        \subcaption{Context length for out-of-distribution series}
        \label{fig:cd_contextlen_ablation_component}
    \end{minipage}

    \vspace{1ex} % Vertical space between rows

    \begin{minipage}{0.48\textwidth}
        \centering
        \includegraphics[width=\textwidth]{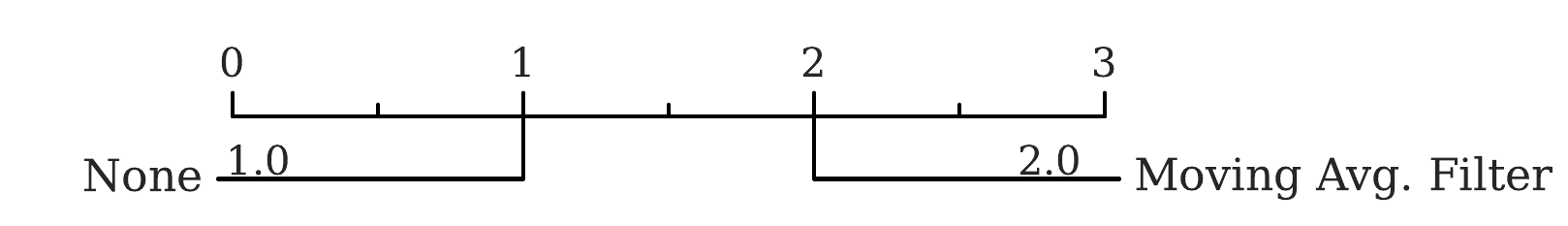}
        \subcaption{Input decomposition for in-distribution series}
        \label{fig:cd_decomp_ablation_aggregate}
    \end{minipage}%
    \hfill
    \begin{minipage}{0.48\textwidth}
        \centering
        \includegraphics[width=\textwidth]{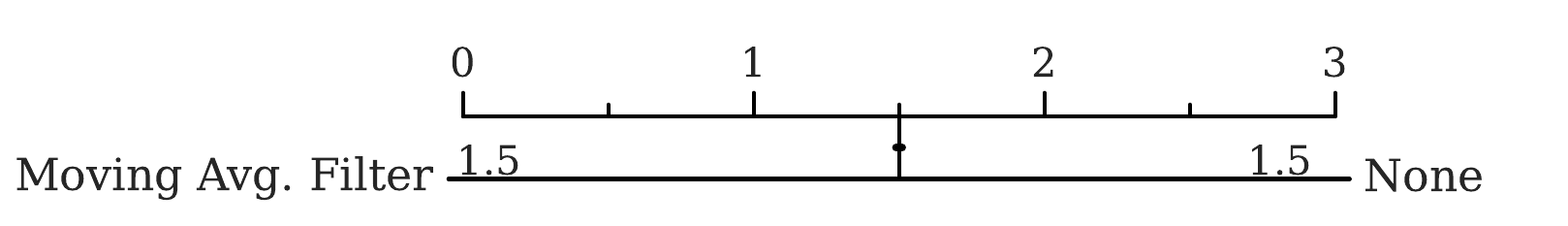}
        \subcaption{Input decomposition function for out-of-distribution series}
        \label{fig:cd_decomp_ablation_component}
    \end{minipage}

    \caption{Critical Difference (CD) diagrams illustrate model ranks and pairwise statistical comparisons of model performance on compositional reasoning tasks across all datasets. Lower ranks indicate better performance. A thick horizontal line groups models that are not significantly different. The statistical tests used to generate the CD diagrams are detailed in Section \ref{section:evaluation}. For analyses comparing three or more methods, the Friedman p-value is included in the subcaptions.}
    \label{fig:cd_diagrams_apd}
\end{figure*}

%% file: tables/nonstationary_table_apd.tex
\begin{table}[!ht] 
\centering
\caption{Mean Absolute Error (MAE) averaged over 3 random seeds (with standard deviation in parentheses) for compositional reasoning task with nonstationary synthetic data. The out-of-distribution (OOD) column presents MAE results for models trained via the compositional reasoning forecasting paradigm. The in-distribution (ID) column presents MAE results for models trained via the traditional forecasting paradigm. The count of instances across datasets where the model ranks in the top three for performance is shown in the second to last column with non-zero entries in \textcolor{blue}{blue}.}
\label{tab:composition_nonstationary_results_table}
\resizebox{0.8\textwidth}{!}{
\begin{tabular}{ll|cc|cc|cc}
\toprule
\multicolumn{2}{c|}{\multirow{2}{*}{\textbf{Model}}} & \multicolumn{2}{c}{\textbf{Trend Dataset \#1}} & \multicolumn{2}{c}{\textbf{Trend Dataset \#2}} & \multicolumn{2}{c}{\textbf{Top 3 Win Count}} \\
\cline{3-8}
{} & {} & \textbf{OOD} & \textbf{ID} & \textbf{OOD} & \textbf{ID} & \small{\textbf{OOD}} & \small{\textbf{ID}} \\
\hline\hline
\multirow{4}{*}{\rotatebox[origin=c]{90}{\textbf{Statistical}}} & \multirow{2}{*}{\ARIMA} & \multirow{2}{*}{--} & 13.542 & \multirow{2}{*}{--} & 12.195 & \multirow{2}{*}{--} & \multirow{2}{*}{\small{0}} \\
{} & {} & {} & (--) & {} & (--) & {} & {} \\
\cline{2-8}
{} & \multirow{2}{*}{\ETS} & \multirow{2}{*}{--} & 11.419 & \multirow{2}{*}{--} & 10.805 & \multirow{2}{*}{--} & \multirow{2}{*}{\small{0}} \\
{} & {} & {} & (--) & {} & (--) & {} & {} \\
\hline
\multirow{4}{*}{\rotatebox[origin=c]{90}{\textbf{Linear}}} & \multirow{2}{*}{\DLinear} & 12.006 & 10.320 & 11.815 & 10.080 & \multirow{2}{*}{\small{0}} & \multirow{2}{*}{\small{0}} \\
{} & {} & \small{(0.168)} & \small{(0.075)} & \small{(0.093)} & \small{(0.176)} & {} & {} \\
\cline{2-8}
{} & \multirow{2}{*}{\NLinear} & 12.244 & 10.569 & 12.270 & 10.385 & \multirow{2}{*}{\small{0}} & \multirow{2}{*}{\small{0}} \\
{} & {} & \small{(0.033)} & \small{(0.616)} & \small{(0.027)} & \small{(0.430)} & {} & {} \\
\hline
\multirow{8}{*}{\rotatebox[origin=c]{90}{\textbf{MLP-Based}}} & \multirow{2}{*}{\MLP} & 3.633 & 0.949 & 3.305 & 0.873 & \multirow{2}{*}{\small{0}} & \multirow{2}{*}{\small{\textcolor{blue}{2}}} \\
{} & {} & \small{(0.219)} & \small{(0.098)} & \small{(0.187)} & \small{(0.184)} & {} & {} \\
\cline{2-8}
{} & \multirow{2}{*}{\NHITS} & 2.098 & 0.391 & 1.631 & 0.388 & \multirow{2}{*}{\small{\textcolor{blue}{2}}} & \multirow{2}{*}{\small{\textcolor{blue}{2}}} \\
{} & {} & \small{(0.311)} & \small{(0.149)} & \small{(0.173)} & \small{(0.241)} & {} & {} \\
\cline{2-8}
{} & \multirow{2}{*}{\NBEATS} & 2.063 & 0.415 & 2.027 & 0.391 & \multirow{2}{*}{\small{\textcolor{blue}{2}}} & \multirow{2}{*}{\small{\textcolor{blue}{2}}} \\
{} & {} & \small{(0.501)} & \small{(0.159)} & \small{(0.314)} & \small{(0.156)} & {} & {} \\
\cline{2-8}
{} & \multirow{2}{*}{\TSMixer} & 10.497 & 10.759 & 20.737 & 9.877 & \multirow{2}{*}{\small{0}} & \multirow{2}{*}{\small{0}} \\
{} & {} & \small{(0.070)} & \small{(0.374)} & \small{(2.975)} & \small{(1.451)} & {} & {} \\
\hline
\multirow{2}{*}{\rotatebox[origin=c]{90}{\textbf{RNN}}} & \multirow{2}{*}{\LSTM} & 12.443 & 3.264 & 11.211 & 2.600 & \multirow{2}{*}{\small{0}} & \multirow{2}{*}{\small{0}} \\
{} & {} & \small{(1.894)} & \small{(0.303)} & \small{(2.903)} & \small{(0.188)} & {} & {} \\
\hline
\multirow{4}{*}{\rotatebox[origin=c]{90}{\textbf{CNN}}} & \multirow{2}{*}{\TCN} & 9.141 & 6.429 & 8.767 & 1.835 & \multirow{2}{*}{\small{0}} & \multirow{2}{*}{\small{0}} \\
{} & {} & \small{(0.715)} & \small{(6.131)} & \small{(0.470)} & \small{(0.084)} & {} & {} \\
\cline{2-8}
{} & \multirow{2}{*}{\TimesNet} & 15.507 & 1.382 & 11.190 & 0.977 & \multirow{2}{*}{\small{0}} & \multirow{2}{*}{\small{0}} \\
{} & {} & \small{(3.226)} & \small{(0.246)} & \small{(2.295)} & \small{(0.079)} & {} & {} \\
\hline
\multirow{34}{*}{\rotatebox[origin=c]{90}{\textbf{Transformer}}} & \multirow{2}{*}{\VanillaTransformer} & 7.679 & 1.807 & 7.524 & 1.161 & \multirow{2}{*}{\small{0}} & \multirow{2}{*}{\small{0}} \\
{} & {} & \small{(0.966)} & \small{(0.460)} & \small{(1.003)} & \small{(0.230)} & {} & {} \\
\cline{2-8}
{} & \multirow{2}{*}{\iTransformer} & 35.090 & 10.783 & 24.743 & 11.578 & \multirow{2}{*}{\small{0}} & \multirow{2}{*}{\small{0}} \\
{} & {} & \small{(0.460)} & \small{(0.089)} & \small{(6.080)} & \small{(0.618)} & {} & {} \\
\cline{2-8}
{} & \multirow{2}{*}{\Autoformer} & 11.222 & 10.354 & 12.341 & 10.662 & \multirow{2}{*}{\small{0}} & \multirow{2}{*}{\small{0}} \\
{} & {} & \small{(0.980)} & \small{(0.102)} & \small{(2.518)} & \small{(0.221)} & {} & {} \\
\cline{2-8}
{} & \multirow{2}{*}{\Informer} & 10.753 & 4.135 & 10.089 & 2.505 & \multirow{2}{*}{\small{0}} & \multirow{2}{*}{\small{0}}\\
{} & {} & \small{(0.123)} & \small{(1.030)} & \small{(0.056)} & \small{(0.175)} & {} & {} \\
\cline{2-8}
{} & \multirow{2}{*}{\TFT} & 7.762 & 1.333 & 8.708 & 2.753 & \multirow{2}{*}{\small{0}} & \multirow{2}{*}{\small{0}} \\
{} & {} & \small{(0.300)} & \small{(0.380)} & \small{(1.909)} & \small{(2.792)} & {} & {} \\
\cline{2-8}
{} & \multirow{2}{*}{\PatchTST (PL=8)} & 20.470 & 4.378 & 20.707 & 4.065& \multirow{2}{*}{\small{0}} & \multirow{2}{*}{\small{0}} \\
{} & {} & \small{(2.690)} & \small{(2.668)} & \small{(4.645)} & \small{(1.636)} & {} & {} \\
\cline{2-8}
{} & \multirow{2}{*}{\PatchTST (PL=16)} & 19.892 & 6.755 & 21.906 & 2.250 & \multirow{2}{*}{\small{0}} & \multirow{2}{*}{\small{0}} \\
{} & {} & \small{(2.738)} & \small{(0.798)} & \small{(7.622)} & \small{(1.169)} & {} & {} \\
\cline{2-8}
{} & \multirow{2}{*}{\PatchTST (PL=32)} & 24.675 & 6.249 & 21.863 & 6.787 & \multirow{2}{*}{\small{0}} & \multirow{2}{*}{\small{0}} \\
{} & {} & \small{(5.563)} & \small{(3.338)} & \small{(4.351)} & \small{(1.388)} & {} & {} \\
\cline{2-8}
{} & \multirow{2}{*}{\PatchTST (PL=64)} & 29.294 & 3.844 & 28.104 & 2.868 & \multirow{2}{*}{\small{0}} & \multirow{2}{*}{\small{0}} \\
{} & {} & \small{(0.819)} & \small{(0.643)} & \small{(8.634)} & \small{(1.311)} & {} & {} \\
\cline{2-8}
{} & \multirow{2}{*}{\PatchTST (PL=96)} & 29.709 & 4.527 & 25.775 & 2.772 & \multirow{2}{*}{\small{0}} & \multirow{2}{*}{\small{0}} \\
{} & {} & \small{(2.147)} & \small{(2.222)} & \small{(9.029)} & \small{(0.722)} & {} & {} \\
\cline{2-8}
{} & \multirow{2}{*}{\PatchTST (PL=128)} & 27.825 & 5.212 & 18.710 & 4.505 & \multirow{2}{*}{\small{0}} & \multirow{2}{*}{\small{0}} \\
{} & {} & \small{(5.415)} & \small{(3.724)} & \small{(3.988)} & \small{(1.236)} & {} & {} \\
\cline{2-8}
{} & \multirow{2}{*}{\Tfive (PL=8)} & 6.074 & 1.998 & 5.405 & 1.216 & \multirow{2}{*}{\small{0}} & \multirow{2}{*}{\small{0}} \\
{} & {} & \small{(0.669)} & \small{(0.282)} & \small{(1.250)} & \small{(0.413)} & {} & {} \\
\cline{2-8}
{} & \multirow{2}{*}{\Tfive (PL=16)} & 4.812 & 1.752 & 4.806 & 1.212 & \multirow{2}{*}{\small{0}} & \multirow{2}{*}{\small{0}} \\
{} & {} & \small{(0.715)} & \small{(0.180)} & \small{(1.408)} & \small{(0.254)} & {} & {} \\
\cline{2-8}
{} & \multirow{2}{*}{\Tfive (PL=32)} & 4.686 & 1.862 & 4.205 & 1.519 & \multirow{2}{*}{\small{0}} & \multirow{2}{*}{\small{0}} \\
{} & {} & \small{(0.346)} & \small{(0.371)} & \small{(0.465)} & \small{(0.378)} & {} & {} \\
\cline{2-8}
{} & \multirow{2}{*}{\Tfive (PL=64)} & 3.464 & 1.741 & 3.528 & 1.532 & \multirow{2}{*}{\small{0}} & \multirow{2}{*}{\small{0}} \\
{} & {} & \small{(0.388)} & \small{(0.314)} & \small{(0.197)} & \small{(0.279)} & {} & {} \\
\cline{2-8}
{} & \multirow{2}{*}{\Tfive (PL=96)} & 3.262 & 1.529 & 3.624 & 1.306 & \multirow{2}{*}{\small{0}}& \multirow{2}{*}{\small{0}} \\
{} & {} & \small{(0.210)} & \small{(0.471)} & \small{(0.507)} & \small{(0.475)} & {} & {} \\
\cline{2-8}
{} & \multirow{2}{*}{\Tfive (PL=128)} & 3.046 & 1.391 & 2.913 & 1.207 & \multirow{2}{*}{\small{\textcolor{blue}{2}}} & \multirow{2}{*}{\small{0}} \\
{} & {} & \small{(0.221)} & \small{(0.142)} & \small{(0.185)} & \small{(0.145)} & {} & {} \\
\hline
\end{tabular}}
\end{table}

%% file: tables/composition_full_table_results.tex
\begin{table}[!ht] 
\centering
\caption{Mean Absolute Error (MAE) averaged over 3 random seeds (with standard deviation in parentheses) for composition reasoning tasks. The out-of-distribution (OOD) column presents MAE results for models trained via the compositional reasoning forecasting paradigm. The in-distribution (ID) column presents MAE results for models trained via the traditional forecasting paradigm. The \Tfive\ with the best patch length (PL) from Table~\ref{tab:composition_t5_results_table} is included. Best results are highlighted in \textbf{bold}, second best results are \underline{underlined}. The count of instances across datasets where the model ranks in the top three for performance is shown in the second to last column with non-zero entries in \textcolor{blue}{blue}. The average number of top $k$ compositions the model can outperform over the datasets is shown in the last column with nonzero entries in \textcolor{purple}{purple}.}
\label{tab:composition_baseline_results_table}
\resizebox{1.0\textwidth}{!}{
\begin{tabular}{ll|cc|cc|cc|cc|cc|cc||cc|cc}
\toprule
\multicolumn{2}{c|}{\multirow{2}{*}{\textbf{Model}}} & \multicolumn{2}{c}{\textbf{Synthetic Sinusoid}} & \multicolumn{2}{c}{\textbf{ECL}} & \multicolumn{2}{c}{\textbf{ETTm2}} & \multicolumn{2}{c}{\textbf{Solar}} & \multicolumn{2}{c}{\textbf{Subseasonal}} & \multicolumn{2}{c||}{\textbf{Loop Seattle}} & \multicolumn{2}{c}{\textbf{\small{Top 3 Win Count}}} & \multicolumn{2}{c}{\textbf{\small{Top-$k_{max}$ Basis Win}}} \\
\cline{3-18}
{} & {} & \textbf{OOD} & \textbf{ID} & \textbf{OOD} & \textbf{ID} & \textbf{OOD} & \textbf{ID} & \textbf{OOD} & \textbf{ID} & \textbf{OOD} & \textbf{ID} & \textbf{OOD} & \textbf{ID} & \textbf{\small{OOD}} & \textbf{\small{ID}} & \textbf{\small{OOD}} & \textbf{\small{ID}} \\
\hline\hline
\multirow{4}{*}{\rotatebox[origin=c]{90}{\textbf{Statistical}}} & \multirow{2}{*}{\ARIMA} & \multirow{2}{*}{--} & 15.538 & \multirow{2}{*}{--} & 0.822 & \multirow{2}{*}{--} & 0.332 & \multirow{2}{*}{--} & 9.687 & \multirow{2}{*}{--} & 7.855 & \multirow{2}{*}{--} & 8.638 & \multirow{2}{*}{\small{--}} & \multirow{2}{*}{\small{0}} & \multirow{2}{*}{\small{--}} & \multirow{2}{*}{\small{0.5}} \\
                      {} & {} &
                      {} & 
                      \small{(--)} & 
                      {} & 
                      \small{(--)} & 
                      {} & 
                      \small{(--)} & 
                      {} & 
                      \small{(--)} &
                      {} & 
                      \small{(--)} & 
                      {} & 
                      \small{(--)} &
                      {} &
                      {} \\
\cline{2-18}
{} & \multirow{2}{*}{\ETS} & \multirow{2}{*}{--} & 16.075 & \multirow{2}{*}{--} & 0.105 & \multirow{2}{*}{--} & 0.211 & \multirow{2}{*}{--} & 1.730 & \multirow{2}{*}{--} & 2.067 & \multirow{2}{*}{--} & 5.575 & \multirow{2}{*}{\small{--}} & \multirow{2}{*}{\small{0}} & \multirow{2}{*}{\small{--}} & \multirow{2}{*}{\small{0.8}} \\
                      {} & {} &
                      {} & 
                      \small{(--)} & 
                      {} & 
                      \small{(--)} & 
                      {} & 
                      \small{(--)} & 
                      {} & 
                      \small{(--)} &
                      {} & 
                      \small{(--)} & 
                      {} & 
                      \small{(--)} &
                      {} &
                      {} \\
\hline
\multirow{4}{*}{\rotatebox[origin=c]{90}{\textbf{Linear}}} & \multirow{2}{*}{\DLinear} & 12.991 & 12.460 & 0.820 & \underline{0.103} & 0.330 & 0.135 & 9.925 & \textbf{1.555} & 8.042 & 1.496 & 9.085 & 4.293 & \multirow{2}{*}{\small{0}} & \multirow{2}{*}{\small{\textcolor{blue}{2}}} & \multirow{2}{*}{\small{0.2}} & \multirow{2}{*}{\small{\textcolor{purple}{55.5}}} \\
                      {} & {} &
                      \small{(0.051)} & \small{(0.025)} & \small{(0.073)} & \small{(0.000)} & \small{(0.028)} & \small{(0.000)} & \small{(1.000)} & \small{(0.007)} & \small{(0.402)} & \small{(0.030)} &
                      \small{(0.327)} & 
                      \small{(0.033)} &
                      {} &
                      {} \\
\cline{2-18}
{} & \multirow{2}{*}{\NLinear} & 13.287 & 13.056 & 0.801 & 0.104 & 0.325 & 0.136 & 9.681 & \underline{1.569} & 8.436 & 1.509 & 9.026 & 4.307 & \multirow{2}{*}{\small{0}} & \multirow{2}{*}{\small{\textcolor{blue}{1}}} & \multirow{2}{*}{\small{0.3}} & \multirow{2}{*}{\small{\textcolor{purple}{53.7}}} \\
                      {} & {} &
                      \small{(0.098)} & \small{(0.060)} & \small{(0.026)} & \small{(0.001)} & \small{(0.007)} & \small{(0.002)} & \small{(1.203)} & \small{(0.010)} & \small{(0.552)} & \small{(0.011)} &
                      \small{(0.248)} & 
                      \small{(0.017)} &
                      {} &
                      {} \\
\hline
\multirow{8}{*}{\rotatebox[origin=c]{90}{\textbf{MLP-Based}}} & \multirow{2}{*}{\MLP} & \underline{8.647} & 2.475 & \underline{0.283} & 0.106 & 0.253 & 0.114 & \underline{4.826} & 1.559 & 1.886 & 1.456 & 7.839 & 3.864 & \multirow{2}{*}{\small{\textcolor{blue}{3}}} & \multirow{2}{*}{\small{\textcolor{blue}{1}}} & \multirow{2}{*}{\small{\textcolor{purple}{10.2}}} & \multirow{2}{*}{\small{\textcolor{purple}{61.2}}} \\
                      {} & {} &
                      \small{(0.258)} & \small{(0.011)} & \small{(0.020)} & \small{(0.004)} & \small{(0.011)} & \small{(0.002)} & \small{(0.043)} & \small{(0.028)} & \small{(0.118)} & \small{(0.046)} &
                      \small{(0.139)} & 
                      \small{(0.061)} &
                      {} &
                      {} \\
\cline{2-18}
{} & \multirow{2}{*}{\NHITS} & 8.924 & \textbf{1.106} & 0.295 & \textbf{0.101} & \textbf{0.214} & \textbf{0.100} & 5.682 & 1.592 & 1.858 & \underline{1.135} & \underline{7.747} & 3.448 & \multirow{2}{*}{\small{\textcolor{blue}{4}}} & \multirow{2}{*}{\small{\textcolor{blue}{4}}} & \multirow{2}{*}{\small{\textcolor{purple}{11.0}}} & \multirow{2}{*}{\small{\textcolor{purple}{64.3}}} \\
                      {} & {} &
                      \small{(0.044)} & \small{(0.036)} & \small{(0.019)} & \small{(0.001)} & \small{(0.006)} & \small{(0.003)} & \small{(0.651)} & \small{(0.026)} & \small{(0.060)} & \small{(0.008)} &
                      \small{(0.141)} & 
                      \small{(0.043)} &
                      {} &
                      {} \\
\cline{2-18}
{} & \multirow{2}{*}{\NBEATS} & 8.907 & 1.383 & 0.294 & \underline{0.103} & \underline{0.216} & \underline{0.102} & 5.852 & 1.599 & \underline{1.840} & 1.177 & 7.763 & 3.479 & \multirow{2}{*}{\small{\textcolor{blue}{5}}} & \multirow{2}{*}{\small{\textcolor{blue}{4}}} & \multirow{2}{*}{\small{\textcolor{purple}{11.2}}} & \multirow{2}{*}{\small{\textcolor{purple}{61.8}}} \\
                      {} & {} &
                      \small{(0.106)} & \small{(0.058)} & \small{(0.016)} & \small{(0.004)} & \small{(0.005)} & \small{(0.002)} & \small{(0.645)} & \small{(0.028)} & \small{(0.108)} & \small{(0.007)} &
                      \small{(0.097)} & 
                      \small{(0.034)} &
                      {} &
                      {} \\
\cline{2-18}
{} & \multirow{2}{*}{\TSMixer} & 14.466 & 15.090 & 0.799 & 0.129 & 0.335 & 0.182 & 9.877 & 1.979 & 7.770 & 1.602 & 8.865 & 5.565 & \multirow{2}{*}{\small{0}} & \multirow{2}{*}{\small{0}} & \multirow{2}{*}{\small{0.3}} & \multirow{2}{*}{\small{\textcolor{purple}{19.7}}} \\
                      {} & {} &
                      \small{(1.804)} & \small{(0.339)} & \small{(0.013)} & \small{(0.004)} & \small{(0.024)} & \small{(0.051)} & \small{(0.118)} & \small{(0.132)} & \small{(0.543)} & \small{(0.099)} &
                      \small{(0.249)} & 
                      \small{(1.038)} &
                      {} &
                      {} \\
\hline
\multirow{2}{*}{\rotatebox[origin=c]{90}{\textbf{RNN}}} & \multirow{2}{*}{\LSTM} & 13.410 & 4.238 & 0.835 & 0.110 & 0.337 & 0.135 & 10.241 & 1.717 & 8.095 & 1.545 & 9.465 & 3.703 & \multirow{2}{*}{\small{0}} & \multirow{2}{*}{\small{0}} & \multirow{2}{*}{\small{0}} & \multirow{2}{*}{\small{\textcolor{purple}{48.7}}} \\
                      {} & {} &
                      \small{(0.203)} & \small{(0.637)} & \small{(0.004)} & \small{(0.001)} & \small{(0.000)} & \small{(0.006)} & \small{(0.620)} & \small{(0.131)} & \small{(0.093)} & \small{(0.075)} &
                      \small{(1.093)} & 
                      \small{(0.054)} &
                      {} &
                      {} \\
\hline
\multirow{4}{*}{\rotatebox[origin=c]{90}{\textbf{CNN}}} & \multirow{2}{*}{\TCN} & 11.478 & 3.833 & 0.837 & 0.106 & 0.339 & 0.135 & 9.868 & 1.642 & 6.170 & 1.234 & 8.792 & \underline{3.422} & \multirow{2}{*}{\small{0}} & \multirow{2}{*}{\small{\textcolor{blue}{1}}} & \multirow{2}{*}{\small{0.7}} & \multirow{2}{*}{\small{\textcolor{purple}{55.7}}} \\
                      {} & {} &
                      \small{(0.410)} & \small{(0.193)} & \small{(0.001)} & \small{(0.002)} & \small{(0.004)} & \small{(0.001)} & \small{(0.016)} & \small{(0.112)} & \small{(3.256)} & \small{(0.031)} &
                      \small{(0.020)} & 
                      \small{(0.119)} &
                      {} &
                      {} \\
\cline{2-18}
{} & \multirow{2}{*}{\TimesNet} & 9.788 & \underline{2.451} & 0.518 & 0.104 & 0.313 & 0.109 & 9.914 & 1.714 & 4.109 & 1.500 & 9.872 & 2.970 & \multirow{2}{*}{\small{0}} & \multirow{2}{*}{\small{\textcolor{blue}{2}}} & \multirow{2}{*}{\small{\textcolor{purple}{2.3}}} & \multirow{2}{*}{\small{\textcolor{purple}{52.8}}} \\
                      {} & {} &
                      \small{(0.493)} & \small{(0.252)} & \small{(0.285)} & \small{(0.003)} & \small{(0.042)} & \small{(0.004)} & \small{(0.116)} & \small{(0.259)} & \small{(3.409)} & \small{(0.111)} &
                      \small{(0.991)} & 
                      \small{(0.360)} &
                      {} &
                      {} \\
\hline
\multirow{22}{*}{\rotatebox[origin=c]{90}{\textbf{Transformer}}} & \multirow{2}{*}{\VanillaTransformer} & 12.279 & 4.935 & 0.919 & 0.106 & 0.334 & 0.136 & 11.956 & 1.667 & 9.641 & 1.276 & 11.675 & 3.591 & \multirow{2}{*}{\small{0}} & \multirow{2}{*}{\small{0}} & \multirow{2}{*}{\small{0.2}} & \multirow{2}{*}{\small{\textcolor{purple}{54.8}}} \\
                      {} & {} &
                      \small{(0.660)} & \small{(0.080)} & \small{(0.033)} & \small{(0.002)} & \small{(0.038)} & \small{(0.005)} & \small{(2.659)} & \small{(0.032)} & \small{(0.363)} & \small{(0.071)} &
                      \small{(0.803)} & 
                      \small{(0.008)} &
                      {} &
                      {} \\
\cline{2-18}
{} & \multirow{2}{*}{\iTransformer} & 15.478 & 15.203 & 0.829 & 0.157 & 0.326 & 0.196 & 9.822 & 1.805 & 8.447 & 1.628 & 9.182 & 4.871 & \multirow{2}{*}{\small{0}} & \multirow{2}{*}{\small{0}} & \multirow{2}{*}{\small{0.2}} & \multirow{2}{*}{\small{\textcolor{purple}{23.5}}} \\
                      {} & {} &
                      \small{(0.351)} & \small{(0.782)} & \small{(0.007)} & \small{(0.007)} & \small{(0.003)} & \small{(0.001)} & \small{(0.055)} & \small{(0.093)} & \small{(0.503)} & \small{(0.027)} &
                      \small{(0.380)} & 
                      \small{(0.105)} &
                      {} &
                      {} \\
\cline{2-18}
{} & \multirow{2}{*}{\Autoformer} & 15.301 & 15.018 & 0.795 & 0.137 & 0.330 & 0.294 & 10.348 & 2.108 & 7.933 & 2.390 & 8.634 & 4.758 & \multirow{2}{*}{\small{0}} & \multirow{2}{*}{\small{0}} & \multirow{2}{*}{\small{0.3}} & \multirow{2}{*}{\small{\textcolor{purple}{15.8}}} \\
                      {} & {} &
                     \small{(0.184)} & \small{(0.130)} & \small{(0.014)} & \small{(0.009)} & \small{(0.008)} & \small{(0.046)} & \small{(0.945)} & \small{(0.638)} & \small{(0.410)} & \small{(0.551)} &
                      \small{(0.166)} & 
                      \small{(0.307)} &
                      {} &
                      {} \\
\cline{2-18}
{} & \multirow{2}{*}{\Informer} & 14.353 & 10.144 & 0.787 & 0.128 & 0.321 & 0.141 & 8.351 & 1.662 & 6.878 & 1.564 & 11.549 & 4.241 & \multirow{2}{*}{\small{0}} & \multirow{2}{*}{\small{0}} & \multirow{2}{*}{\small{0.5}} & \multirow{2}{*}{\small{\textcolor{purple}{41.0}}} \\
                      {} & {} &
                      \small{(0.174)} & \small{(2.257)} & \small{(0.082)} & \small{(0.006)} & \small{(0.016)} & \small{(0.007)} & \small{(1.560)} & \small{(0.051)} & \small{(1.340)} & \small{(0.094)} &
                      \small{(0.627)} & 
                      \small{(0.105)} &
                      {} &
                      {} \\
\cline{2-18}
{} & \multirow{2}{*}{\TFT} & 14.531 & 9.745 & 0.445 & 0.115 & 0.312 & 0.117 & 12.873 & 2.106 & 2.684 & 1.454 & 11.280 & 5.340 & \multirow{2}{*}{\small{0}} & \multirow{2}{*}{\small{0}} & \multirow{2}{*}{\small{\textcolor{purple}{4.7}}} & \multirow{2}{*}{\small{\textcolor{purple}{28.5}}} \\
                      {} & {} &
                      \small{(0.880)} & \small{(1.210)} & \small{(0.072)} & \small{(0.010)} & \small{(0.029)} & \small{(0.005)} & \small{(2.141)} & \small{(0.538)} & \small{(0.146)} & \small{(0.151)} &
                      \small{(1.182)} & 
                      \small{(0.855)} &
                      {} &
                      {} \\
\cline{2-18}
{} & \multirow{2}{*}{\PatchTST\ (PL=8)} & 13.808 & 12.036 & 0.713 & 0.428 & 0.309 & 0.156 & 9.081 & 2.350 & 6.242 & 2.007 & 11.440 & 4.755 & \multirow{2}{*}{\small{0}} & \multirow{2}{*}{\small{0}} & \multirow{2}{*}{\small{0.7}} & \multirow{2}{*}{\small{\textcolor{purple}{18.7}}} \\
                      {} & {} &
                      \small{(0.471)} & \small{(0.738)} & \small{(0.172)} & \small{(0.318)} & \small{(0.031)} & \small{(0.026)} & \small{(0.857)} & \small{(0.872)} & \small{(1.286)} & \small{(0.228)} &
                      \small{(1.564)} & 
                      \small{(0.774)} &
                      {} &
                      {} \\
\cline{2-18}
{} & \multirow{2}{*}{\PatchTST\ (PL=16)} & 14.133 & 13.633 & 0.666 & 0.279 & 0.253 & 0.150 & 10.788 & 1.633 & 5.877 & 1.904 & 9.855 & 4.989 & \multirow{2}{*}{\small{0}} & \multirow{2}{*}{\small{0}} & \multirow{2}{*}{\small{0.8}} & \multirow{2}{*}{\small{\textcolor{purple}{27.8}}} \\
                      {} & {} &
                      \small{(2.693)} & \small{(1.202)} & \small{(0.185)} & \small{(0.025)} & \small{(0.017)} & \small{(0.005)} & \small{(0.139)} & \small{(0.013)} & \small{(1.389)} & \small{(0.336)} &
                      \small{(0.618)} & 
                      \small{(1.000)} &
                      {} &
                      {} \\
\cline{2-18}
{} & \multirow{2}{*}{\PatchTST\ (PL=32)} & 13.316 & 13.412 & 0.736 & 0.273 & 0.271 & 0.168 & 8.267 & 2.721 & 4.904 & 2.385 & 9.422 & 4.924 & \multirow{2}{*}{\small{0}} & \multirow{2}{*}{\small{0}} & \multirow{2}{*}{\small{\textcolor{purple}{1.3}}} & \multirow{2}{*}{\small{\textcolor{purple}{14.3}}} \\
                      {} & {} &
                      \small{(1.751)} & \small{(2.499)} & \small{(0.248)} & \small{(0.106)} & \small{(0.010)} & \small{(0.002)} & \small{(1.302)} & \small{(1.625)} & \small{(2.018)} & \small{(0.639)} &
                      \small{(1.359)} & 
                      \small{(0.133)} &
                      {} &
                      {} \\
\cline{2-18}
{} & \multirow{2}{*}{\PatchTST\ (PL=64)} & 12.232 & 13.544 & 0.482 & 0.122 & 0.247 & 0.169 & 8.054 & 2.813 & 2.292 & 1.754 & 10.529 & 4.405 & \multirow{2}{*}{\small{\textcolor{blue}{1}}} & \multirow{2}{*}{\small{0}} & \multirow{2}{*}{\small{\textcolor{purple}{6.8}}} & \multirow{2}{*}{\small{\textcolor{purple}{27.2}}} \\
                      {} & {} &
                      \small{(0.252)} & \small{(1.055)} & \small{(0.306)} & \small{(0.011)} & \small{(0.007)} & \small{(0.040)} & \small{(2.248)} & \small{(0.724)} & \small{(0.152)} & \small{(0.353)} &
                      \small{(4.987)} & 
                      \small{(0.883)} &
                      {} &
                      {} \\
\cline{2-18}
{} & \multirow{2}{*}{\PatchTST\ (PL=96)} & 11.235 & 8.374 & 0.508 & 0.196 & 0.250 & 0.147 & 6.426 & 1.799 & 2.185 & 1.659 & 7.965 & 4.579 & \multirow{2}{*}{\small{0}} & \multirow{2}{*}{\small{0}} & \multirow{2}{*}{\small{\textcolor{purple}{7.7}}} & \multirow{2}{*}{\small{\textcolor{purple}{27.0}}} \\
                      {} & {} &
                      \small{(0.483)} & \small{(1.834)} & \small{(0.287)} & \small{(0.051)} & \small{(0.013)} & \small{(0.021)} & \small{(1.312)} & \small{(0.304)} & \small{(0.168)} & \small{(0.282)} &
                      \small{(0.355)} & 
                      \small{(0.978)} &
                      {} &
                      {} \\
\cline{2-18}
{} & \multirow{2}{*}{\PatchTST\ (PL=128)} & 10.696 & 6.959 & 0.832 & 0.161 & 0.323 & 0.138 & 5.726 & 1.964 & 2.448 & 1.678 & 9.378 & 3.653 & \multirow{2}{*}{\small{0}} & \multirow{2}{*}{\small{0}} & \multirow{2}{*}{\small{\textcolor{purple}{5.7}}} & \multirow{2}{*}{\small{\textcolor{purple}{32.7}}} \\
                      {} & {} &
                      \small{(0.907)} & \small{(2.126)} & \small{(0.010)} & \small{(0.053)} & \small{(0.037)} & \small{(0.018)} & \small{(0.567)} & \small{(0.266)} & \small{(0.467)} & \small{(0.100)} &
                      \small{(0.185)} & 
                      \small{(0.257)} &
                      {} &
                      {} \\
\cline{2-18}
{} & \multirow{2}{*}{\Tfive\ (Best PL)} & \textbf{7.177} & 2.480 & \textbf{0.239} & \underline{0.103} & 0.259 & 0.103 & \textbf{3.899} & 1.578 & \textbf{1.714} & \textbf{1.097} & \textbf{6.589} & \textbf{3.351} & \multirow{2}{*}{\small{\textcolor{blue}{5}}} & \multirow{2}{*}{\small{\textcolor{blue}{4}}} & \multirow{2}{*}{\small{\textcolor{purple}{12.2}}} & \multirow{2}{*}{\small{\textcolor{purple}{64.3}}} \\
                      {} & {} &
                      \small{(0.089)} & 
                      \small{(0.198)} & 
                      \small{(0.005)} & 
                      \small{(0.001)} & 
                      \small{(0.008)} & 
                      \small{(0.006)} & 
                      \small{(0.578)} & 
                      \small{(0.009)} &
                      \small{(0.040) } &
                      \small{0.039)} &
                      \small{(0.109)} & 
                      \small{(0.017)} &
                      {} &
                      {} \\
\bottomrule
\end{tabular}
}
\end{table}

%% file: tables/composition_full_table_t5_results.tex
\begin{table}[ht]
\centering
\caption{Mean Absolute Error (MAE) averaged over 3 random seeds (with standard deviation in parentheses) for composition reasoning tasks. The out-of-distribution (OOD) column presents MAE results for models trained via the compositional reasoning forecasting paradigm. The in-distribution (ID) column presents MAE results for models trained via the traditional forecasting paradigm. Best results are highlighted in \textbf{bold}. The count of instances across datasets where the model has the best performance is shown in the last column with non-zero entries in \textcolor{blue}{blue}.}
\label{tab:composition_t5_results_table}
\resizebox{1.0\textwidth}{!}{
\begin{tabular}{ll|cc|cc|cc|cc|cc|cc||cc}
\toprule
\multicolumn{2}{c|}{\multirow{2}{*}{\textbf{Transformer Model (T5 Backbone)}}} & \multicolumn{2}{c}{\textbf{Synthetic Sinusoid}} & \multicolumn{2}{c}{\textbf{ECL}} & \multicolumn{2}{c}{\textbf{ETTm2}} & \multicolumn{2}{c}{\textbf{Solar}} & \multicolumn{2}{c}{\textbf{Subseasonal}} & \multicolumn{2}{c||}{\textbf{Loop Seattle}} & \multicolumn{2}{c}{\textbf{\small{Win Count}}} \\
\cline{3-16}
{} & {} & \textbf{OOD} & \textbf{ID} & \textbf{OOD} & \textbf{ID} & \textbf{OOD} & \textbf{ID} & \textbf{OOD} & \textbf{ID} & \textbf{OOD} & \textbf{ID} & \textbf{OOD} & \textbf{ID} & \textbf{\small{OOD}} & \textbf{\small{ID}} \\
\hline\hline
\multirow{8}{*}{\rotatebox[origin=c]{90}{\textbf{Tokenization}}} & \multirow{2}{*}{None} & 14.032 & 4.894 & 0.685 & \textbf{0.106} & 0.369 & 0.121 & 9.969 & \textbf{1.635} & 6.401 & 1.572 & 14.327 & 3.766 & \multirow{2}{*}{\small{0}} & \multirow{2}{*}{\small{\textcolor{blue}{2}}}\\
                      {} & {} &
                      \small{(1.612)} & 
                      \small{(0.140)} & 
                      \small{(0.148)} & 
                      \small{(0.005)} & 
                      \small{(0.015)} & 
                      \small{(0.010)} & 
                      \small{(0.804)} & 
                      \small{(0.079)} &
                      \small{(1.739)} & 
                      \small{(0.049) } &
                      \small{(2.548)} & 
                      \small{(0.035)} \\
\cline{2-16}
{} & \multirow{2}{*}{Fixed Length Patches} & \textbf{8.648} & \textbf{2.611} & \textbf{0.266} & 0.107 & \textbf{0.268} & \textbf{0.100} & \textbf{3.908} & 1.663 & \textbf{1.729} & \textbf{1.154} & \textbf{7.658} & \textbf{3.118} & \multirow{2}{*}{\small{\textcolor{blue}{6}}} & \multirow{2}{*}{\small{\textcolor{blue}{4}}}\\
                      {} & {} &
                      \small{(0.072)} & 
                      \small{(0.158)} & 
                      \small{(0.019)} & 
                      \small{(0.002)} & 
                      \small{(0.003)} & 
                      \small{(0.003)} & 
                      \small{(0.204)} & 
                      \small{(0.023)} &
                      \small{(0.028)} & 
                      \small{(0.029)} &
                      \small{(0.100)} & 
                      \small{(0.026)} \\
\cline{2-16}
{} & \multirow{2}{*}{Binning} & 17.039 & 9.504 & 0.833 & 0.270 & 0.317 & 0.199 & 8.445 & 4.719 & 3.758 & 3.253 & 12.728 & 7.735 & \multirow{2}{*}{\small{0}} & \multirow{2}{*}{\small{0}}\\
                      {} & {} &
                      \small{(0.788)} & 
                      \small{(0.502)} & 
                      \small{(0.007)} & 
                      \small{(0.003)} & 
                      \small{(0.011)} & 
                      \small{(0.004)} & 
                      \small{(4.206)} & 
                      \small{(0.151)} &
                      \small{(0.365)} & 
                      \small{(0.532)} &
                      \small{(2.339)} & 
                      \small{(0.985)} \\
\cline{2-16}
{} & \multirow{2}{*}{Lags} & 13.442 & 4.599 & 0.820 & 0.120 & 0.415 & 0.126 & 10.156  & 1.669 & 4.022 & 1.376 & 11.638 & 3.897 & \multirow{2}{*}{\small{0}} & \multirow{2}{*}{\small{0}} \\
                      {} & {} &
                      \small{(0.254)} & 
                      \small{(0.328)} & 
                      \small{(0.178)} & 
                      \small{(0.005)} & 
                      \small{(0.046)} & 
                      \small{(0.006)} & 
                      \small{(1.364)} & 
                      \small{(0.030)} &
                      \small{(0.187)} & 
                      \small{(0.043)} &
                      \small{(1.556)} & 
                      \small{(0.074)} \\
\hline\hline
\multirow{8}{*}{\rotatebox[origin=c]{90}{\textbf{Model Size}}} & \multirow{2}{*}{Tiny} & \textbf{7.644} & 2.628 & \textbf{0.239} & 0.105 & 0.274 & 0.109 & 3.899 & \textbf{1.641} & 1.899 & 1.097 & \textbf{6.701} & 3.355 & \multirow{2}{*}{\small{\textcolor{blue}{3}}} & \multirow{2}{*}{\small{\textcolor{blue}{1}}} \\
                      {} & {} &
                      \small{(0.025)} & 
                      \small{(0.108)} & 
                      \small{(0.005)} & 
                      \small{(0.002)} & 
                      \small{(0.005)} & 
                      \small{(0.006)} & 
                      \small{(0.578)} & 
                      \small{(0.068)} &
                      \small{(0.203)} & 
                      \small{(0.039)} &
                      \small{(0.153)} & 
                      \small{(0.053)} \\
\cline{2-16}
{} & \multirow{2}{*}{Mini} & 7.882 & 2.318 & 0.242 & 0.107 & \textbf{0.273} & 0.103 & \textbf{3.810} & 1.663 & \textbf{1.769} & 1.104 & 6.888 & 3.018 & \multirow{2}{*}{\small{\textcolor{blue}{3}}} & \multirow{2}{*}{\small{0}}\\
                      {} & {} &
                      \small{(0.069)} & 
                      \small{(0.076)} & 
                      \small{(0.006)} & 
                      \small{(0.001)} & 
                      \small{(0.005)} & 
                      \small{(0.001)} & 
                      \small{(0.226)} & 
                      \small{(0.027)} &
                      \small{(0.108)} & 
                      \small{(0.008)} &
                      \small{(0.032)} & 
                      \small{(0.011)} \\
\cline{2-16}
{} & \multirow{2}{*}{Small} & 8.057 & 2.103 & 0.268 & 0.103 & 0.268 & 0.096 & 4.172 & 1.665 & 1.770 & 1.048 & 6.924 & 2.764 & \multirow{2}{*}{\small{0}} & \multirow{2}{*}{\small{0}}\\
                      {} & {} &
                      \small{(0.119)} & 
                      \small{(0.098)} & 
                      \small{(0.009)} & 
                      \small{(0.002)} & 
                      \small{(0.014)} & 
                      \small{(0.001)} & 
                      \small{(0.192)} & 
                      \small{(0.028)} &
                      \small{(0.197)} & 
                      \small{(0.016)} &
                      \small{(0.187)} & 
                      \small{(0.033)} \\
\cline{2-16}
{} & \multirow{2}{*}{Base} & 8.308 & \textbf{2.084} & 0.247 & \textbf{0.100} & 0.274 & \textbf{0.091} & 4.338 & 1.667 & 1.853 & \textbf{0.967} & 7.005 & \textbf{2.536} & \multirow{2}{*}{\small{0}} & \multirow{2}{*}{\small{\textcolor{blue}{5}}} \\
                      {} & {} &
                      \small{(0.228)} & 
                      \small{(0.072)} & 
                      \small{(0.014)} & 
                      \small{(0.006)} & 
                      \small{(0.007)} & 
                      \small{(0.001)} & 
                      \small{(0.123)} & 
                      \small{(0.065)} &
                      \small{(0.150)} & 
                      \small{(0.012)} &
                      \small{(0.255)} & 
                      \small{(0.061)} \\
\hline\hline
\multirow{4}{*}{\rotatebox[origin=c]{90}{\textbf{Attn. Type}}} & \multirow{2}{*}{Bidirectional Attn.} & \textbf{7.644} & \textbf{2.628} & \textbf{0.239} & \textbf{0.105} & 0.274 & 0.109 & \textbf{3.899} & 1.641 & 1.899 & \textbf{1.097} & \textbf{6.701} & 3.355 & \multirow{2}{*}{\small{\textcolor{blue}{4}}} & \multirow{2}{*}{\small{\textcolor{blue}{3}}}\\
                      {} & {} &
                      \small{(0.025)} & 
                      \small{(0.108)} & 
                      \small{(0.005)} & 
                      \small{(0.002)} & 
                      \small{(0.005)} & 
                      \small{(0.006)} & 
                      \small{(0.578)} & 
                      \small{(0.068)} &
                      \small{(0.203)} & 
                      \small{(0.039)} &
                      \small{(0.153)} & 
                      \small{(0.053)} \\
\cline{2-16}
{} & \multirow{2}{*}{Causal Attn.} & 7.978 & 2.828 & 0.248 & 0.106 & \textbf{0.267} & \textbf{0.105} & 4.307 & \textbf{1.589} & \textbf{1.820} & 1.170 & 6.891 & \textbf{3.337} & \multirow{2}{*}{\small{\textcolor{blue}{2}}} & \multirow{2}{*}{\small{\textcolor{blue}{3}}} \\
                      {} & {} &
                      \small{(0.02)} & 
                      \small{(0.233)} & 
                      \small{(0.005)} & 
                      \small{(0.004)} & 
                      \small{(0.010)} & 
                      \small{(0.003)} & 
                      \small{(0.144)} & 
                      \small{(0.042)} &
                      \small{(0.173)} & 
                      \small{(0.054)} &
                      \small{(0.133)} & 
                      \small{(0.044)} \\
\hline\hline
\multirow{4}{*}{\rotatebox[origin=c]{90}{\textbf{Proj./Head}}} & \multirow{2}{*}{Linear} & \textbf{7.644} & 2.628 & \textbf{0.239} & 0.105 & 0.274 & 0.109 & \textbf{3.899} & 1.641 & 1.899 & 1.097 & \textbf{6.701} & 3.355 & \multirow{2}{*}{\small{\textcolor{blue}{4}}} & \multirow{2}{*}{\small{0}} \\
                      {} & {} &
                      \small{(0.025)} & 
                      \small{(0.108)} & 
                      \small{(0.005)} & 
                      \small{(0.002)} & 
                      \small{(0.005)} & 
                      \small{(0.006)} & 
                      \small{(0.578)} & 
                      \small{(0.068)} &
                      \small{(0.203)} & 
                      \small{(0.039)} &
                      \small{(0.153)} & 
                      \small{(0.053)} \\
\cline{2-16}
{} & \multirow{2}{*}{Residual} & 8.537 & \textbf{2.617} & 0.311 & \textbf{0.102} & \textbf{0.256} & \textbf{0.085} & 4.880 & \textbf{1.595} & \textbf{1.871} & \textbf{0.924} & 7.931 & \textbf{2.524} & \multirow{2}{*}{\small{\textcolor{blue}{2}}} & \multirow{2}{*}{\small{\textcolor{blue}{6}}} \\
                      {} & {} &
                      \small{(0.184)} & 
                      \small{(0.043)} &
                      \small{(0.011)} & 
                      \small{(0.002)} & 
                      \small{(0.008)} & 
                      \small{(0.003)} & 
                      \small{(0.389)} &
                      \small{(0.039)} & 
                      \small{(0.296)} &
                      \small{(0.016)} &
                      \small{(0.655)} &
                      \small{(0.028)} \\
\hline\hline
\multirow{12}{*}{\rotatebox[origin=c]{90}{\textbf{Token (Patch) Length}}} & \multirow{2}{*}{8} & 9.327 & 3.183 & 0.346 & \textbf{0.103} & 0.300 & 0.111 & 7.721 & \textbf{1.578} & 2.581 & 1.532 & 8.048 & 3.573 & \multirow{2}{*}{\small{0}} & \multirow{2}{*}{\small{\textcolor{blue}{2}}} \\
                      {} & {} &
                      \small{(0.28)} & 
                      \small{(0.117)} & 
                      \small{(0.040)} & 
                      \small{(0.001)} & 
                      \small{(0.013)} & 
                      \small{(0.003)} & 
                      \small{(0.640)} & 
                      \small{(0.009)} &
                      \small{(0.335)} & 
                      \small{(0.002)} &
                      \small{(0.108)} & 
                      \small{(0.047)} \\
\cline{2-16}
{} & \multirow{2}{*}{16} & 9.007 & 3.573 & 0.418 & 0.105 & 0.283 & \textbf{0.103} & 9.139 & 1.673 & 2.731 & 1.362 & 8.499 & 3.510 & \multirow{2}{*}{\small{0}} & \multirow{2}{*}{\small{\textcolor{blue}{1}}} \\
                      {} & {} &
                      \small{(0.057)} & 
                      \small{(0.235)} & 
                      \small{(0.062)} & 
                      \small{(0.002)} & 
                      \small{(0.011)} & 
                      \small{(0.006)} & 
                      \small{(0.400)} & 
                      \small{(0.088)} &
                      \small{(0.939)} & 
                      \small{(0.25)} &
                      \small{(0.339)} & 
                      \small{(0.060)} \\
\cline{2-16}
{} & \multirow{2}{*}{32} & 10.11 & 3.719 & 0.244 & 0.108 & 0.289 & 0.109 & 5.256 & 1.652 & 2.059 & 1.318 & 7.014 & 3.463 & \multirow{2}{*}{\small{0}} & \multirow{2}{*}{\small{0}} \\
                      {} & {} &
                      \small{(0.07)} & 
                      \small{(0.190)} & 
                      \small{(0.011)} & 
                      \small{(0.004)} & 
                      \small{(0.007)} & 
                      \small{(0.002)} & 
                      \small{(0.591)} & 
                      \small{(0.071)} &
                      \small{(0.171)} & 
                      \small{(0.187)} &
                      \small{(0.167)} & 
                      \small{(0.015)} \\
\cline{2-16}
{} & \multirow{2}{*}{64} & 8.747 & 2.971 & \textbf{0.239 }& 0.106 & 0.285 & 0.107 & 4.201 & 1.651 & 1.819 & 1.265 & \textbf{6.589} & 3.408 & \multirow{2}{*}{\small{\textcolor{blue}{2}}} & \multirow{2}{*}{\small{0}} \\
                      {} & {} &
                      \small{(0.243)} & 
                      \small{(0.209)} & 
                      \small{(0.009)} & 
                      \small{(0.003)} & 
                      \small{(0.002)} & 
                      \small{(0.006)} & 
                      \small{(0.103)} & 
                      \small{(0.076)} &
                      \small{(0.032)} & 
                      \small{(0.198)} &
                      \small{(0.109)} & 
                      \small{(0.044)} \\
\cline{2-16}
{} & \multirow{2}{*}{96} & 7.644 & 2.628 & \textbf{0.239} & 0.105 & 0.274 & 0.109 & \textbf{3.899} & 1.641 & 1.899 & \textbf{1.097} & 6.701 & 3.355 & \multirow{2}{*}{\small{\textcolor{blue}{2}}} & \multirow{2}{*}{\small{\textcolor{blue}{1}}} \\
                      {} & {} &
                      \small{(0.025)} & 
                      \small{(0.108)} & 
                      \small{(0.005)} & 
                      \small{(0.002)} & 
                      \small{(0.005)} & 
                      \small{(0.006)} & 
                      \small{(0.578)} & 
                      \small{(0.068)} &
                      \small{(0.203)} & 
                      \small{(0.039)} &
                      \small{(0.153)} & 
                      \small{(0.053)} \\
\cline{2-16}
{} & \multirow{2}{*}{128} & \textbf{7.177} & \textbf{2.480} & 0.255 & 0.107 & \textbf{0.259} & 0.107 & 4.385 & 1.673 & \textbf{1.714} & 1.100 & 6.878 & \textbf{3.351} & \multirow{2}{*}{\small{\textcolor{blue}{3}}} & \multirow{2}{*}{\small{\textcolor{blue}{2}}} \\
                      {} & {} &
                      \small{(0.089)} & 
                      \small{(0.198)} & 
                      \small{(0.012)} & 
                      \small{(0.001)} & 
                      \small{(0.008)} & 
                      \small{(0.006)} & 
                      \small{(0.145)} & 
                      \small{(0.033)} &
                      \small{(0.040)} & 
                      \small{(0.069)} &
                      \small{(0.069)} & 
                      \small{(0.017)} \\
\hline\hline
\multirow{8}{*}{\rotatebox[origin=c]{90}{\textbf{Positional Encoding}}} & \multirow{2}{*}{Relative} & 7.751 & 2.750 & 0.284 & 0.105 & \textbf{0.268} & 0.111 & 4.373 & 1.640 & \textbf{1.826} & \textbf{1.093} & 6.883 & \textbf{3.337} & \multirow{2}{*}{\small{\textcolor{blue}{2}}} & \multirow{2}{*}{\small{\textcolor{blue}{2}}} \\
                      {} & {} &
                      \small{(0.163)} & 
                      \small{(0.262)} & 
                      \small{(0.059) } & 
                      \small{(0.002)} & 
                      \small{(0.015)} & 
                      \small{(0.002)} & 
                      \small{(0.254)} & 
                      \small{(0.033)} &
                      \small{(0.156)} & 
                      \small{(0.025)} &
                      \small{(0.162)} & 
                      \small{(0.020)} \\
\cline{2-16}
{} & \multirow{2}{*}{SinCos} & 7.881 & \textbf{2.456} & 0.247 & \textbf{0.104} & 0.270 & 0.112 & 4.357 & \textbf{1.624} & 1.839 & 1.464 & 6.818 & 3.366 & \multirow{2}{*}{\small{0}} & \multirow{2}{*}{\small{\textcolor{blue}{3}}} \\
                      {} & {} &
                      \small{(0.148)} & 
                      \small{(0.049)} & 
                      \small{(0.015)} & 
                      \small{(0.003)} & 
                      \small{(0.007)} & 
                      \small{(0.004)} & 
                      \small{(0.284)} & 
                      \small{(0.025)} &
                      \small{(0.195)} & 
                      \small{(0.051)} &
                      \small{(0.182} & 
                      \small{(0.055)} \\
\cline{2-16}
{} & \multirow{2}{*}{SinCos+Relative} & \textbf{7.644} & 2.628 & \textbf{0.239} & 0.105 & 0.274 & \textbf{0.109} & \textbf{3.899} & 1.641 & 1.899 & 1.097 & \textbf{6.701} & 3.355 & \multirow{2}{*}{\small{\textcolor{blue}{4}}} & \multirow{2}{*}{\small{\textcolor{blue}{1}}} \\
                      {} & {} &
                      \small{(0.025)} & 
                      \small{(0.108)} & 
                      \small{(0.005)} & 
                      \small{(0.002)} & 
                      \small{(0.005)} & 
                      \small{(0.006)} & 
                      \small{(0.578)} & 
                      \small{(0.068)} &
                      \small{(0.203)} & 
                      \small{(0.039)} &
                      \small{(0.153)} & 
                      \small{(0.053)} \\
\cline{2-16}
{} & \multirow{2}{*}{RoPE} & 7.921 & 2.701 & 0.255 & \textbf{0.104} & 0.274 & 0.112 & 4.608 & 1.647 & 1.872 & 1.106 & 6.721 & 3.378 & \multirow{2}{*}{\small{0}} & \multirow{2}{*}{\small{\textcolor{blue}{1}}} \\
                      {} & {} &
                      \small{(0.125)} & 
                      \small{(0.077)} & 
                      \small{(0.006)} & 
                      \small{(0.002)} & 
                      \small{(0.002)} & 
                      \small{(0.002)} & 
                      \small{(0.333)} & 
                      \small{(0.032)} &
                      \small{(0.125)} &
                      \small{(0.037)} & 
                      \small{(0.174)} &
                      \small{(0.033)} \\
\hline\hline
\multirow{8}{*}{\rotatebox[origin=c]{90}{\textbf{Loss Function}}} & \multirow{2}{*}{MAE} & \textbf{7.644} & 2.628 & 0.239 & 0.105 & 0.274 &\textbf{ 0.109} & 3.899 & \textbf{1.641} & 1.899 & \textbf{1.097} & 6.701 & 3.355 & \multirow{2}{*}{\small{\textcolor{blue}{1}}} & \multirow{2}{*}{\small{\textcolor{blue}{3}}} \\
                      {} & {} &
                      \small{(0.025)} & 
                      \small{(0.108)} & 
                      \small{(0.005)} & 
                      \small{(0.002)} & 
                      \small{(0.005)} & 
                      \small{(0.006)} & 
                      \small{(0.578)} & 
                      \small{(0.068)} &
                      \small{(0.203)} & 
                      \small{(0.039)} &
                      \small{(0.153)} & 
                      \small{(0.053)} \\
\cline{2-16}
{} & \multirow{2}{*}{MSE} & 7.694 & \textbf{2.618} & \textbf{0.229} & 0.109 & \textbf{0.262} & 0.116 & 3.633 & 1.752 & 1.891 & 1.411 & 6.613 & 3.302 & \multirow{2}{*}{\small{\textcolor{blue}{2}}} & \multirow{2}{*}{\small{\textcolor{blue}{1}}} \\
                      {} & {} &
                      \small{(0.178)} & 
                      \small{(0.088)} & 
                      \small{(0.006)} & 
                      \small{(0.004)} & 
                      \small{(0.007)} & 
                      \small{(0.010)} & 
                      \small{(0.466)} & 
                      \small{(0.030)} &
                      \small{(0.193)} & 
                      \small{(0.214)} &
                      \small{(0.118)} & 
                      \small{(0.045)} \\
\cline{2-16}
{} & \multirow{2}{*}{Huber} & 7.648 & 2.914 & 0.234 & 0.108 & 0.264 & 0.112 & \textbf{4.035} & 1.746 & 1.801 & 1.258 & \textbf{6.571} & \textbf{3.281} & \multirow{2}{*}{\small{\textcolor{blue}{2}}} & \multirow{2}{*}{\small{\textcolor{blue}{1}}} \\
                      {} & {} &
                      \small{(0.076)} & 
                      \small{(0.234)} & 
                      \small{(0.012)} & 
                      \small{(0.005)} & 
                      \small{(0.012)} & 
                      \small{(0.003)} & 
                      \small{(0.633)} & 
                      \small{(0.056)} &
                      \small{(0.084)} & 
                      \small{(0.194)} &
                      \small{(0.182)} & 
                      \small{(0.047)} \\
\cline{2-16}
{} & \multirow{2}{*}{StudentT} & 7.776 & 2.660 & 0.244 & \textbf{0.104} & 0.270 & 0.113 & 4.134 & 1.648 & \textbf{1.735} & 1.377 & 6.872 & 3.543 & \multirow{2}{*}{\small{\textcolor{blue}{1}}} & \multirow{2}{*}{\small{\textcolor{blue}{1}}} \\
                      {} & {} &
                      \small{(0.281)} & 
                      \small{(0.105)} & 
                      \small{(0.003)} & 
                      \small{(0.003)} & 
                      \small{(0.004)} & 
                      \small{(0.003)} & 
                      \small{(0.804)} & 
                      \small{(0.025)} &
                      \small{(0.037)} & 
                      \small{(0.024)} &
                      \small{(0.191)} & 
                      \small{(0.026)} \\
\hline\hline
\multirow{4}{*}{\rotatebox[origin=c]{90}{\textbf{Scaler}}} & \multirow{2}{*}{RevIN (Standard, non-learnable)} & \textbf{7.644} & \textbf{2.628} & \textbf{0.239} & 0.105 & 0.274 & 0.109 & \textbf{3.899} & \textbf{1.641} & 1.899 & \textbf{1.097} & \textbf{6.701} & \textbf{3.355} & \multirow{2}{*}{\small{\textcolor{blue}{4}}} & \multirow{2}{*}{\small{\textcolor{blue}{4}}} \\
                      {} & {} &
                      \small{(0.025)} & 
                      \small{(0.108)} & 
                      \small{(0.005)} & 
                      \small{(0.002)} & 
                      \small{(0.005)} & 
                      \small{(0.006)} & 
                      \small{(0.578)} & 
                      \small{(0.068)} &
                      \small{(0.203)} & 
                      \small{(0.039)} &
                      \small{(0.153)} & 
                      \small{(0.053)} \\
\cline{2-16}
% {} & \multirow{2}{*}{RevIN (Standard, learnable)} & 7.644 & 2.628 & 0.239 & 0.105 & 0.274 & 0.109 & 3.899 & 1.641 & 1.899 & 1.097 & 6.701 & 3.355 \\
%                       {} & {} &
%                       \small{(0.025)} & 
%                       \small{(0.108)} & 
%                       \small{(0.005)} & 
%                       \small{(0.002)} & 
%                       \small{(0.005)} & 
%                       \small{(0.006)} & 
%                       \small{(0.578)} & 
%                       \small{(0.068)} &
%                       \small{(0.203)} & 
%                       \small{(0.039)} &
%                       \small{(0.153)} & 
%                       \small{(0.053)} \\
% \cline{2-14}
{} & \multirow{2}{*}{Robust} & 7.931 & 2.725 & 0.326 & \textbf{0.103} & \textbf{0.270} & \textbf{0.106} & 8.170 & 1.734 & \textbf{1.736} & 1.232 & 6.982 & 3.412 & \multirow{2}{*}{\small{\textcolor{blue}{2}}} & \multirow{2}{*}{\small{\textcolor{blue}{2}}} \\
                      {} & {} &
                      \small{(0.026)} & 
                      \small{(0.234)} & 
                      \small{(0.006)} & 
                      \small{(0.003)} & 
                      \small{(0.007)} & 
                      \small{(0.007)} & 
                      \small{(0.389)} & 
                      \small{(0.093)} &
                      \small{(0.015)} & 
                      \small{(0.227)} &
                      \small{(0.271)} & 
                      \small{(0.043)} \\
\hline\hline
\multirow{4}{*}{\rotatebox[origin=c]{90}{\textbf{Context}}} & \multirow{2}{*}{256} & 7.644 & 2.628 & \textbf{0.239} & \textbf{0.105} & \textbf{0.274} & \textbf{0.109} & \textbf{3.899} & \textbf{1.641} & \textbf{1.899} & \textbf{1.097} & 6.701 & \textbf{3.355} & \multirow{2}{*}{\small{\textcolor{blue}{4}}} & \multirow{2}{*}{\small{\textcolor{blue}{5}}} \\
                      {} & {} &
                      \small{(0.025)} & 
                      \small{(0.108)} & 
                      \small{(0.005)} & 
                      \small{(0.002)} & 
                      \small{(0.005)} & 
                      \small{(0.006)} & 
                      \small{(0.578)} & 
                      \small{(0.068)} &
                      \small{(0.203)} & 
                      \small{(0.039)} &
                      \small{(0.153)} & 
                      \small{(0.053)} \\
\cline{2-16}
{} & \multirow{2}{*}{512} & \textbf{7.072} & \textbf{2.605} & 0.253 & 0.111 & 0.298 & 0.151 & 4.187 & 1.740 & 1.934 & 1.829 & \textbf{6.295} & 4.013 & \multirow{2}{*}{\small{\textcolor{blue}{2}}} & \multirow{2}{*}{\small{\textcolor{blue}{1}}} \\
                      {} & {} &
                      \small{(0.382)} & 
                      \small{(0.158)} & 
                      \small{(0.037)} & 
                      \small{(0.005)} & 
                      \small{(0.015)} & 
                      \small{(0.008)} & 
                      \small{(0.216)} & 
                      \small{(0.046)} &
                      \small{(0.083)} & 
                      \small{(0.094)} &
                      \small{(0.284)} & 
                      \small{(0.002)} \\
\hline\hline
\multirow{4}{*}{\rotatebox[origin=c]{90}{\textbf{Decomp.}}} & \multirow{2}{*}{None} & 7.644 & \textbf{2.628} & \textbf{0.239} & \textbf{0.105} & 0.274 & \textbf{0.109} & \textbf{3.899} & \textbf{1.641} & 1.899 & \textbf{1.097} & \textbf{6.701} & \textbf{3.355} & \multirow{2}{*}{\small{\textcolor{blue}{3}}} & \multirow{2}{*}{\small{\textcolor{blue}{6}}} \\
                      {} & {} &
                      \small{(0.025)} & 
                      \small{(0.108)} & 
                      \small{(0.005)} & 
                      \small{(0.002)} & 
                      \small{(0.005)} & 
                      \small{(0.006)} & 
                      \small{(0.578)} & 
                      \small{(0.068)} &
                      \small{(0.203)} & 
                      \small{(0.039)} &
                      \small{(0.153)} & 
                      \small{(0.053)} \\
\cline{2-16}
{} & \multirow{2}{*}{Moving Avg. Filter (DLinear, Autoformer)} & \textbf{6.726} & 2.860 & 0.265 & 0.106 & \textbf{0.264} & 0.112 & 4.018 & 1.663 & \textbf{1.769} & 1.252 & 6.942 & 3.470 & \multirow{2}{*}{\small{\textcolor{blue}{3}}} & \multirow{2}{*}{\small{0}} \\
                      {} & {} &
                      \small{(0.184)} & 
                      \small{(0.483)} & 
                      \small{(0.015)} & 
                      \small{(0.002)} & 
                      \small{(0.015)} & 
                      \small{(0.001)} & 
                      \small{(0.256)} & 
                      \small{(0.063)} &
                      \small{(0.132)} & 
                      \small{(0.243)} &
                      \small{(0.283)} & 
                      \small{(0.043)} \\
\bottomrule
\end{tabular}
}
\end{table}

%% file: arxiv/arxiv_version.bbl
\begin{thebibliography}{63}
\providecommand{\natexlab}[1]{#1}
\providecommand{\url}[1]{\texttt{#1}}
\expandafter\ifx\csname urlstyle\endcsname\relax
  \providecommand{\doi}[1]{doi: #1}\else
  \providecommand{\doi}{doi: \begingroup \urlstyle{rm}\Url}\fi

\bibitem[Aksu et~al.(2024)Aksu, Woo, Liu, Liu, Liu, Savarese, Xiong, and Sahoo]{aksu2024giftevalbenchmark}
Taha Aksu, Gerald Woo, Juncheng Liu, Xu~Liu, Chenghao Liu, Silvio Savarese, Caiming Xiong, and Doyen Sahoo.
\newblock Gift-eval: A benchmark for general time series forecasting model evaluation, 2024.
\newblock URL \url{https://arxiv.org/abs/2410.10393}.

\bibitem[Allen-Zhu and Li(2023)]{allenshu2023physics3.2}
Zeyuan Allen-Zhu and Yuanzhi Li.
\newblock Physics of language models: Part 3.2, knowledge manipulation, 2023.

\bibitem[Allen-Zhu and Li(2024)]{allenzhu2023physics3.1}
Zeyuan Allen-Zhu and Yuanzhi Li.
\newblock Physics of language models: Part 3.1, knowledge storage and extraction.
\newblock In \emph{Proceedings of the 41st International Conference on Machine Learning}, page 235, 2024.

\bibitem[Ansari et~al.(2024)Ansari, Stella, Turkmen, Zhang, Mercado, Shen, Shchur, Rangapuram, Arango, Kapoor, Zschiegner, Maddix, Wang, Mahoney, Torkkola, Wilson, Bohlke-Schneider, and Wang]{ansari2024chronos}
Abdul~Fatir Ansari, Lorenzo Stella, Caner Turkmen, Xiyuan Zhang, Pedro Mercado, Huibin Shen, Oleksandr Shchur, Syama~Sundar Rangapuram, Sebastian~Pineda Arango, Shubham Kapoor, Jasper Zschiegner, Danielle~C. Maddix, Hao Wang, Michael~W. Mahoney, Kari Torkkola, Andrew~Gordon Wilson, Michael Bohlke-Schneider, and Yuyang Wang.
\newblock Chronos: Learning the language of time series.
\newblock \emph{Transactions on Machine Learning Research}, 2024.

\bibitem[Bai et~al.(2018)Bai, Kolter, and Koltun]{bai2018_tcn}
Shaojie Bai, J.~Zico Kolter, and Vladlen Koltun.
\newblock An empirical evaluation of generic convolutional and recurrent networks for sequence modeling, 2018.

\bibitem[Brown(1956)]{brown_1956_ets}
Robert~G. Brown.
\newblock Exponential smoothing for predicting demand.
\newblock \emph{Philip Morris Records}, 1956.

\bibitem[Cai et~al.(2024)Cai, Choudhry, Goswami, and Dubrawski]{cai2024timeseriesexam}
Yifu Cai, Arjun Choudhry, Mononito Goswami, and Artur Dubrawski.
\newblock Time{S}eries{E}xam: A time series understanding exam.
\newblock \emph{NeurIPS 2024 Workshop on Time Series in the Age of Large Model}, 2024.

\bibitem[Challu et~al.(2022)Challu, Olivares, Oreshkin, Garza, Mergenthaler-Canseco, and Dubrawski]{challu_olivares2022_nhits}
Cristian Challu, Kin~G. Olivares, Boris~N. Oreshkin, Federico Garza, Max Mergenthaler-Canseco, and Artur Dubrawski.
\newblock {N-HiTS}: Neural hierarchical interpolation for time series forecasting.
\newblock In \emph{AAAI-23}, 2022.

\bibitem[Chen et~al.(2023)Chen, Li, Yoder, $\ddot{\text{O}}$. Arık, and Pfister]{chen2023tsmixer}
Si-An Chen, Chun-Liang Li, Nathanael~C. Yoder, Sercan $\ddot{\text{O}}$. Arık, and Tomas Pfister.
\newblock {TSMixer}: An all-{MLP} architecture for time series forecasting.
\newblock In \emph{Published in Transactions on Machine Learning Research}, 2023.

\bibitem[Chow et~al.(2024)Chow, Gardiner, Hallgrímsson, Xu, and Ren]{chow2024llmstimeseriesreasoning}
Winnie Chow, Lauren Gardiner, Haraldur~T. Hallgrímsson, Maxwell~A. Xu, and Shirley~You Ren.
\newblock Towards time-series reasoning with llms.
\newblock \emph{NeurIPS 2024 Workshop on Time Series in the Age of Large Model}, 2024.

\bibitem[Cui et~al.(2018)Cui, Ke, and Wang]{cui2018deep}
Zhiyong Cui, Ruimin Ke, and Yinhai Wang.
\newblock Deep bidirectional and unidirectional lstm recurrent neural network for network-wide traffic speed prediction.
\newblock \emph{arXiv preprint arXiv:1801.02143}, 2018.

\bibitem[Cui et~al.(2019)Cui, Henrickson, Ke, and Wang]{cui2019traffic}
Zhiyong Cui, Kristian Henrickson, Ruimin Ke, and Yinhai Wang.
\newblock Traffic graph convolutional recurrent neural network: A deep learning framework for network-scale traffic learning and forecasting.
\newblock \emph{IEEE Transactions on Intelligent Transportation Systems}, 2019.

\bibitem[Das et~al.(2024)Das, Kong, Sen, and Zhou]{das2024TimesFM}
Abhimanyu Das, Weihao Kong, Rajat Sen, and Yichen Zhou.
\newblock A decoder-only foundation model for time-series forecasting, 2024.

\bibitem[Dem\v{s}sar(2006)]{demsar2006cddiagrams}
Janez Dem\v{s}sar.
\newblock Statistical comparisons of classifiers over multiple data sets.
\newblock In \emph{Journal of Machine Learning Research}, pages 1--30, 2006.

\bibitem[Edwards et~al.(2024)Edwards, Alvey, Alsing, Nguyen, and Wandelt]{edwards2024scaling}
Thomas~DP Edwards, James Alvey, Justin Alsing, Nam~H Nguyen, and Benjamin~D Wandelt.
\newblock Scaling-laws for large time-series models.
\newblock \emph{arXiv preprint arXiv:2405.13867}, 2024.

\bibitem[Ekambaram et~al.(2024)Ekambaram, Jati, Nguyen, Dayama, Reddy, Gifford, and Kalagnanam]{ekambaram2024ttms}
Vijay Ekambaram, Arindam Jati, Nam~H Nguyen, Pankaj Dayama, Chandra Reddy, Wesley~M Gifford, and Jayant Kalagnanam.
\newblock {TTMs: Fast Multi-level Tiny Time Mixers for Improved Zero-shot and Few-shot Forecasting of Multivariate Time Series}.
\newblock \emph{arXiv preprint arXiv:2401.03955}, 2024.

\bibitem[Fukushima(1975)]{fukushima1975_mlp}
Kunihiko Fukushima.
\newblock Cognitron: A self-organizing multilayered neural network.
\newblock \emph{Biol. Cybernetics}, 20:\penalty0 121--–136, 1975.

\bibitem[Gao et~al.(2024)Gao, Koker, Queen, Hartvigsen, Tsiligkaridis, and Zitnik]{gao2024units}
Shanghua Gao, Teddy Koker, Owen Queen, Thomas Hartvigsen, Theodoros Tsiligkaridis, and Marinka Zitnik.
\newblock Units: A unified multi-task time series model.
\newblock In \emph{The Thirty-eighth Annual Conference on Neural Information Processing Systems}, 2024.

\bibitem[Garza and Mergenthaler-Canseco(2023)]{garza2023timegpt1}
Azul Garza and Max Mergenthaler-Canseco.
\newblock {TimeGPT-1}, 2023.

\bibitem[Goswami et~al.(2024{\natexlab{a}})Goswami, Sanil, Choudhry, Srinivasan, Udompanyawit, and Dubrawski]{goswami2024aqua}
Mononito Goswami, Vedant Sanil, Arjun Choudhry, Arvind Srinivasan, Chalisa Udompanyawit, and Artur Dubrawski.
\newblock Aqua: A benchmarking tool for label quality assessment.
\newblock \emph{Advances in Neural Information Processing Systems}, 36, 2024{\natexlab{a}}.

\bibitem[Goswami et~al.(2024{\natexlab{b}})Goswami, Szafer, Choudhry, Cai, Li, and Dubrawski]{goswami2024moment}
Mononito Goswami, Konrad Szafer, Arjun Choudhry, Yifu Cai, Shuo Li, and Artur Dubrawski.
\newblock {MOMENT}: A family of open time-series foundation models.
\newblock In \emph{41st International Conference on Machine Learning}, 2024{\natexlab{b}}.

\bibitem[Hanna et~al.(2023)Hanna, Liu, and Variengien]{hanna2023gptgreaterthan}
Michael Hanna, Ollie Liu, and Alexandre Variengien.
\newblock How does gpt-2 compute greater-than?: Interpreting mathematical abilities in a pre-trained language model.
\newblock In \emph{37th Conference on Neural Information Processing Systems}, 2023.

\bibitem[Hyndman and Khandakar(2008)]{hyndman2008_arima}
R.~J. Hyndman and Y.~Khandakar.
\newblock Automatic time series forecasting: The forecast package for r.
\newblock \emph{Journal of Statistical Software}, 27\penalty0 (3):\penalty0 1--22, 2008.

\bibitem[Hyndman et~al.(2024)Hyndman, Athanasopoulos, Garza, Challu, Mergenthaler, and Olivares]{hyndman2024forecasting}
Rob~J Hyndman, George Athanasopoulos, Azul Garza, Cristian Challu, Max Mergenthaler, and Kin~G. Olivares.
\newblock \emph{{Forecasting: Principles and Practice, the Pythonic Way}}.
\newblock {OTexts}, {Melbourne, Australia}, 2024.
\newblock available at https://otexts.com/fpppy/.

\bibitem[Ismail~Fawaz et~al.(2019)Ismail~Fawaz, Forestier, Weber, Idoumghar, and Muller]{IsmailFawaz2018deep}
Hassan Ismail~Fawaz, Germain Forestier, Jonathan Weber, Lhassane Idoumghar, and Pierre-Alain Muller.
\newblock Deep learning for time series classification: a review.
\newblock \emph{Data Mining and Knowledge Discovery}, 33\penalty0 (4):\penalty0 917--963, 2019.

\bibitem[Jiang et~al.(2024)Jiang, Han, Jiang, Zhao, and Wang]{jiang2024libcityunifiedlibraryefficient}
Jiawei Jiang, Chengkai Han, Wenjun Jiang, Wayne~Xin Zhao, and Jingyuan Wang.
\newblock Libcity: A unified library towards efficient and comprehensive urban spatial-temporal prediction, 2024.
\newblock URL \url{https://arxiv.org/abs/2304.14343}.

\bibitem[Lake and Baroni(2018)]{lake2018generalization}
Brenden~M. Lake and Marco Baroni.
\newblock Generalization without systematicity: On the compositional skills of sequence-to-sequence recurrent networks.
\newblock In \emph{Proceedings of the 35th International Conference on Machine Learning}, 2018.

\bibitem[Lim et~al.(2021)Lim, $\ddot{\text{O}}$. Arık, Loeff, and Pfister]{lim2021_tft}
Bryan Lim, Sercan $\ddot{\text{O}}$. Arık, Nicolas Loeff, and Tomas Pfister.
\newblock Temporal fusion transformers for interpretable multi-horizon time series forecasting.
\newblock \emph{International Journal of Forecasting}, 37\penalty0 (4):\penalty0 1748--1764, 2021.

\bibitem[Liu et~al.(2024{\natexlab{a}})Liu, Liu, Woo, Aksu, Liang, Zimmermann, Liu, Savarese, Xiong, and Sahoo]{moiraimoe}
Xu~Liu, Juncheng Liu, Gerald Woo, Taha Aksu, Yuxuan Liang, Roger Zimmermann, Chenghao Liu, Silvio Savarese, Caiming Xiong, and Doyen Sahoo.
\newblock Moirai-moe: Empowering time series foundation models with sparse mixture of experts.
\newblock \emph{arXiv preprint arXiv:2410.10469}, 2024{\natexlab{a}}.

\bibitem[Liu et~al.(2024{\natexlab{b}})Liu, Hu, Zhang, Wu, Wang, Ma, and Long]{liu2024itransformerinvertedtransformerseffective}
Yong Liu, Tengge Hu, Haoran Zhang, Haixu Wu, Shiyu Wang, Lintao Ma, and Mingsheng Long.
\newblock {iTransformer}: Inverted transformers are effective for time series forecasting, 2024{\natexlab{b}}.

\bibitem[Liu et~al.(2024{\natexlab{c}})Liu, Qin, Huang, Wang, and Long]{timerxl}
Yong Liu, Guo Qin, Xiangdong Huang, Jianmin Wang, and Mingsheng Long.
\newblock Timer-xl: Long-context transformers for unified time series forecasting.
\newblock \emph{arXiv preprint arXiv:2410.04803}, 2024{\natexlab{c}}.

\bibitem[Liu et~al.(2024{\natexlab{d}})Liu, Zhang, Li, Huang, Wang, and Long]{liutimer}
Yong Liu, Haoran Zhang, Chenyu Li, Xiangdong Huang, Jianmin Wang, and Mingsheng Long.
\newblock {Timer: Generative Pre-trained Transformers Are Large Time Series Models}.
\newblock In \emph{Forty-first International Conference on Machine Learning}, 2024{\natexlab{d}}.

\bibitem[Merrill et~al.(2024)Merrill, Tan, Gupta, Hartvigsen, and Althoff]{althoff2024llmtimeseriesreasoningstruggle}
Mike~A. Merrill, Mingtian Tan, Vinayak Gupta, Tom Hartvigsen, and Tim Althoff.
\newblock Language models still struggle to zero-shot reason about time series, 2024.

\bibitem[Mouatadid et~al.(2023)Mouatadid, Orenstein, Flaspohler, Oprescu, Cohen, Wang, Knight, Geogdzhayeva, Levang, Fraenkel, and Mackey]{mouatadid2024subseasonalsubseasonaldataset}
Soukayna Mouatadid, Paulo Orenstein, Genevieve Flaspohler, Miruna Oprescu, Judah Cohen, Franklyn Wang, Sean Knight, Maria Geogdzhayeva, Sam Levang, Ernest Fraenkel, and Lester Mackey.
\newblock Subseasonalclimateusa: A dataset for subseasonal forecasting and benchmarking.
\newblock In \emph{37th Conference on Neural Information Processing Systems (NeurIPS 2023) Track on Datasets and Benchmarks}, 2023.

\bibitem[Nair and Hinton(2010)]{nair2010_mlp}
Vinod Nair and Jeoffrey~E. Hinton.
\newblock Rectified linear units improve restricted boltzmann machines.
\newblock In \emph{ICML-23}, 2010.

\bibitem[Ni et~al.(2024)Ni, Yu, Liu, Li, and Lin]{ni2024basisformer}
Zelin Ni, Hang Yu, Shizhan Liu, Jianguo Li, and Weiyao Lin.
\newblock Basisformer: Attention-based time series forecasting with learnable and interpretable basis.
\newblock In \emph{37th Conference on Neural Information Processing Systems}, 2024.

\bibitem[Nie et~al.(2023)Nie, Nguyen, and an~Jayant~Kalagnanam2]{nie2023patchtst}
Yuqi Nie, Nam~H. Nguyen, and Phanwadee~Sinthong an~Jayant~Kalagnanam2.
\newblock A time series is worth 64 words: Long-term forecasting with transformers.
\newblock In \emph{Proceedings of the 11th International Conference on Learning Representations}, 2023.

\bibitem[Olivares et~al.(2022{\natexlab{a}})Olivares, Challu, Marcjasz, Weron, and Dubrawski]{OlivaresChallu2022_nbeats}
Kin~G. Olivares, Cristian Challu, Grzegorz Marcjasz, Rafal Weron, and Artur Dubrawski.
\newblock Neural basis expansion analysis with exogenous variables: Forecasting electricity prices with nbeatsx.
\newblock \emph{International Journal of Forecasting}, 39\penalty0 (2):\penalty0 884--900, 2022{\natexlab{a}}.

\bibitem[Olivares et~al.(2022{\natexlab{b}})Olivares, Challú, Garza, Canseco, and Dubrawski]{olivares2022library_neuralforecast}
Kin~G. Olivares, Cristian Challú, Federico Garza, Max~Mergenthaler Canseco, and Artur Dubrawski.
\newblock {NeuralForecast}: User friendly state-of-the-art neural forecasting models.
\newblock {PyCon} Salt Lake City, Utah, US 2022, 2022{\natexlab{b}}.
\newblock URL \url{https://github.com/Nixtla/neuralforecast}.

\bibitem[Oreshkin et~al.(2020)Oreshkin, Carpov, Chapados, and Bengio]{oreshkin2020nbeats}
Boris~N. Oreshkin, Dmitri Carpov, Nicolas Chapados, and Yoshua Bengio.
\newblock {N-BEATS:} neural basis expansion analysis for interpretable time series forecasting.
\newblock In \emph{8th International Conference on Learning Representations, {ICLR} 2020}, 2020.
\newblock URL \url{https://openreview.net/forum?id=r1ecqn4YwB}.

\bibitem[Rasul et~al.(2024)Rasul, Ashok, Williams, Ghonia, Bhagwatkar, Khorasani, Bayazi, Adamopoulos, Riachi, Hassen, Biloš, Garg, Schneider, Chapados, Drouin, Zantedeschi, Nevmyvaka, and Rish]{rasul2024lagllama}
Kashif Rasul, Arjun Ashok, Andrew~Robert Williams, Hena Ghonia, Rishika Bhagwatkar, Arian Khorasani, Mohammad Javad~Darvishi Bayazi, George Adamopoulos, Roland Riachi, Nadhir Hassen, Marin Biloš, Sahil Garg, Anderson Schneider, Nicolas Chapados, Alexandre Drouin, Valentina Zantedeschi, Yuriy Nevmyvaka, and Irina Rish.
\newblock {Lag-Llama}: Towards foundation models for probabilistic time series forecasting, 2024.

\bibitem[Rosenblatt(1958)]{rosenblatt1958_mlp}
Frank Rosenblatt.
\newblock The perceptron: A probabilistic model for information storage and organization in the brain.
\newblock \emph{Psychological Review}, 65\penalty0 (6):\penalty0 386–--408, 1958.

\bibitem[Sak et~al.(2014)Sak, Senior, and Beaufays]{sak2014_lstm}
Haşim Sak, Andrew Senior, and Françoise Beaufays.
\newblock Long short-term memory based recurrent neural network architectures for large vocabulary speech recognition, 2014.

\bibitem[Shi et~al.(2024)Shi, Wang, Nie, Li, Ye, Wen, and Jin]{timemoe}
Xiaoming Shi, Shiyu Wang, Yuqi Nie, Dianqi Li, Zhou Ye, Qingsong Wen, and Ming Jin.
\newblock Time-moe: Billion-scale time series foundation models with mixture of experts.
\newblock \emph{arXiv preprint arXiv:2409.16040}, 2024.

\bibitem[Tan et~al.(2024)Tan, Merrill, Gupta, Althoff, and Hartvigsen]{hartvigsen2024llmforecasting}
Mingtian Tan, Mike~A. Merrill, Vinayak Gupta, Tim Althoff, and Thomas Hartvigsen.
\newblock Are language models actually useful for time series forecasting?
\newblock In \emph{Proceedings of the 38th Conference on Neural Information Processing Systems}, 2024.

\bibitem[van~den Oord et~al.(2016)van~den Oord, Dieleman, Zen, Simonyan, Vinyals, Graves, Kalchbrenner, Senior, and Kavukcuoglu]{oord2016_tcn}
Aaron van~den Oord, Sander Dieleman, Heiga Zen, Karen Simonyan, Oriol Vinyals, Alex Graves, Nal Kalchbrenner, Andrew Senior, and Koray Kavukcuoglu.
\newblock {WaveNet}: A generative model for raw audio, 2016.

\bibitem[Vaswani et~al.(2017)Vaswani, Shazeer, Parmar, Uszkoreit, Jones, and Gomez]{vaswani_2021_attentionisallyouneed}
Ashish Vaswani, Noam Shazeer, Niki Parmar, Jakob Uszkoreit, Llion Jones, and Aidan~N. Gomez.
\newblock Attention is all you need, 2017.

\bibitem[Wang et~al.(2024)Wang, Yue, Su, and Sun]{wang2024grokked}
Boshi Wang, Xiang Yue, Yu~Su, and Huan Sun.
\newblock Grokked transformers are implicit reasoners: A mechanistic journey to the edge of generalization.
\newblock In \emph{38th Conference on Neural Information Processing Systems}, 2024.

\bibitem[Wili{\'n}ski et~al.(2025)Wili{\'n}ski, Goswami, Potosnak, {\.Z}ukowska, and Dubrawski]{wilinski2024exploring}
Micha{\l} Wili{\'n}ski, Mononito Goswami, Willa Potosnak, Nina {\.Z}ukowska, and Artur Dubrawski.
\newblock Exploring representations and interventions in time series foundation models.
\newblock In \emph{Proceedings of the Forty-Second International Conference on Machine Learning}, 2025.

\bibitem[Williams et~al.(2025)Williams, Ashok, Étienne Marcotte, Zantedeschi, Subramanian, Riachi, and et~al.]{Williams2025llmtimeseriestext}
Andrew~Robert Williams, Arjun Ashok, Étienne Marcotte, Valentina Zantedeschi, Jithendaraa Subramanian, Roland Riachi, and et~al.
\newblock Context is key: A benchmark for forecasting with essential textual information, 2025.

\bibitem[Wolf et~al.(2020)Wolf, Debut, Sanh, Chaumond, Delangue, Moi, Cistac, Rault, Louf, Funtowicz, Davison, Shleifer, von Platen, Ma, Jernite, Plu, Xu, Scao, Gugger, Drame, Lhoest, and Rush]{wolf_2020_hgtransformers}
Thomas Wolf, Lysandre Debut, Victor Sanh, Julien Chaumond, Clement Delangue, Anthony Moi, Pierric Cistac, Tim Rault, Rémi Louf, Morgan Funtowicz, Joe Davison, Sam Shleifer, Patrick von Platen, Clara Ma, Yacine Jernite, Julien Plu, Canwen Xu, Teven~Le Scao, Sylvain Gugger, Mariama Drame, Quentin Lhoest, and Alexander~M. Rush.
\newblock Transformers: State-of-the-art natural language processing.
\newblock In \emph{Proceedings of the 2020 Conference on Empirical Methods in Natural Language Processing: System Demonstrations}, pages 38--45, Online, October 2020. Association for Computational Linguistics.
\newblock URL \url{https://www.aclweb.org/anthology/2020.emnlp-demos.6}.

\bibitem[Wolff et~al.(2024)Wolff, Olivares, Oreshkin, Ruan, Yang, Katoch, Ramasubramanian, Zhang, Mahoney, Efimov, and Quenneville-Bélair]{wolff2024spade}
Malcolm Wolff, Kin~G. Olivares, Boris Oreshkin, Sunny Ruan, Sitan Yang, Abhinav Katoch, Shankar Ramasubramanian, Youxin Zhang, Michael~W. Mahoney, Dmitry Efimov, and Vincent Quenneville-Bélair.
\newblock $\spadesuit$ {SPADE} $\spadesuit$: Split peak attention decomposition.
\newblock In \emph{Thirty-Eighth Annual Conference on Neural Information Processing Systems {NeurIPS 2024}}, volume Time Series in the Age of Large Models Workshop, Vancouver, Canada, 2024. {NeurIPS 2024}.
\newblock URL \url{https://arxiv.org/abs/2411.05852}.

\bibitem[Woo et~al.(2024)Woo, Liu, Kumar, Xiong, Savarese, and Sahoo]{moirai2024}
Gerald Woo, Chenghao Liu, Akshat Kumar, Caiming Xiong, Silvio Savarese, and Doyen Sahoo.
\newblock Unified training of universal time series forecasting transformers.
\newblock In \emph{Proceedings of the 41st International Conference on Machine Learning}, 2024.

\bibitem[Wu et~al.(2021)Wu, Xu, Wang, and Long]{wu_2021_autoformer}
Haixu Wu, Jiehui Xu, Jianmin Wang, and Mingsheng Long.
\newblock Autoformer: Decomposition transformers with auto-correlation for long-term series forecasting, 2021.

\bibitem[Wu et~al.(2023)Wu, Hu, Liu, Zhou, Wang, and Long]{wu2023timesnettemporal2dvariationmodeling}
Haixu Wu, Tengge Hu, Yong Liu, Hang Zhou, Jianmin Wang, and Mingsheng Long.
\newblock {TimesNet}: Temporal 2d-variation modeling for general time series analysis.
\newblock In \emph{Proceedings of the 34th International Conference on Learning Representations}, 2023.

\bibitem[Yang et~al.(2024{\natexlab{a}})Yang, Cao, Li, and Yang]{yang2024rethinkingfourier}
Runze Yang, Longbing Cao, Jianxun Li, and Jie Yang.
\newblock Rethinking fourier transform from a basis functions perspective for long-term time series forecasting.
\newblock In \emph{Proceedings of the 38th Conference on Neural Information Processing Systems}, 2024{\natexlab{a}}.

\bibitem[Yang et~al.(2024{\natexlab{b}})Yang, Gribovskaya, Kassner, Geva, and Riedel]{soheeyang2024multihopreasoning}
Sohee Yang, Elena Gribovskaya, Nora Kassner, Mor Geva, and Sebastian Riedel.
\newblock Do large language models latently perform multi-hop reasoning?
\newblock \emph{arXiv preprint arXiv:2402.16837}, 2024{\natexlab{b}}.

\bibitem[Yao et~al.(2024)Yao, Yang, Jiang, Liang, Jin, and Pan]{yao2024scaling}
Qingren Yao, Chao-Han~Huck Yang, Renhe Jiang, Yuxuan Liang, Ming Jin, and Shirui Pan.
\newblock Towards neural scaling laws for time series foundation models.
\newblock \emph{arXiv preprint arXiv:2410.12360}, 2024.

\bibitem[Zeng et~al.(2023)Zeng, Chen, Zhang, and Xu]{zeng_2023_dlinear}
Ailing Zeng, Muxi Chen, Lei Zhang, and Qiang Xu.
\newblock Are transformers effective for time series forecasting?
\newblock In \emph{Proceedings of the AAAI Conference on Artificial Intelligence}, 2023.

\bibitem[Zhong et~al.(2023)Zhong, Wu, Manning, Potts, and Chen]{zhong2023mquake}
Zexuan Zhong, Zhengxuan Wu, Christopher~D Manning, Christopher Potts, and Danqi Chen.
\newblock {MQuAKE}: Assessing knowledge editing in language models via multi-hop questions.
\newblock \emph{arXiv preprint arXiv:2305.14795}, 2023.

\bibitem[Zhou et~al.(2021)Zhou, Zhang, Peng, Zhang, Li, Xiong, and Zhang]{zhou2021informerefficienttransformerlong}
Haoyi Zhou, Shanghang Zhang, Jieqi Peng, Shuai Zhang, Jianxin Li, Hui Xiong, and Wancai Zhang.
\newblock Informer: Beyond efficient transformer for long sequence time-series forecasting, 2021.

\bibitem[Zhou et~al.(2024)Zhou, Fu, Sharan, and Jia]{zhou2024llmfourieraddition}
Tianyi Zhou, Deqing Fu, Vatsal Sharan, and Robin Jia.
\newblock Pre-trained large language models use fourier features to compute addition, 2024.

\bibitem[{\.Z}ukowska et~al.(2024){\.Z}ukowska, Goswami, Wili{\'n}ski, Potosnak, and Dubrawski]{infinichannelmixer}
Nina {\.Z}ukowska, Mononito Goswami, Micha{\l} Wili{\'n}ski, Willa Potosnak, and Artur Dubrawski.
\newblock Towards long-context time series foundation models.
\newblock \emph{arXiv preprint arXiv:2409.13530}, 2024.

\end{thebibliography}
